\documentclass{article}
\newif\ifarxivstyle
\arxivstyletrue
\ifarxivstyle
  \usepackage{template/arxiv}
  \usepackage{natbib}
  \setcitestyle{authoryear,round,citesep={;},aysep={,},yysep={;}} %
  
  \usepackage{times}
  \providecommand{\iclrfinalcopy}{}
\fi

\usepackage{amsmath,amsfonts,bm}

\def\eqref#1{equation~\ref{#1}}

\def\1{\bm{1}}

\DeclareMathAlphabet{\mathsfit}{\encodingdefault}{\sfdefault}{m}{sl}
\SetMathAlphabet{\mathsfit}{bold}{\encodingdefault}{\sfdefault}{bx}{n}

\newcommand{\E}{\mathbb{E}}

\newcommand{\R}{\mathbb{R}}

\usepackage{hyperref}
\hypersetup{hypertexnames=false,colorlinks=true,linkcolor=[rgb]{0,0.08,0.45},citecolor=[rgb]{0,0.08,0.45},urlcolor=[rgb]{0,0.08,0.45}}
\usepackage{url}
\usepackage{amsmath,amssymb}
\usepackage{mathtools}
\usepackage{amsthm}
\usepackage{booktabs}
\usepackage{array} %
\usepackage{enumitem}
\usepackage{graphicx}
\usepackage{float}
\usepackage{wrapfig}
\usepackage{capt-of}
\usepackage{placeins}
\usepackage{pifont}
\usepackage[most]{tcolorbox}
\usepackage{enumitem}
\usepackage{graphicx}
\usepackage{oplotsymbl}

\newif\ifshowupdates
\showupdatesfalse

\definecolor{updatecolor}{RGB}{220,90,0}  %
\newcommand{\update}[1]{{\color{updatecolor}#1}}
\newcommand{\updatebox}[1]{{\setlength{\fboxsep}{3pt}\setlength{\fboxrule}{1.2pt}\fcolorbox{updatecolor}{white}{#1}}}

\newcommand{\hypbox}[2]{%
\begin{tcolorbox}[colback=white!98!black,colframe=white!30!black,boxsep=1.1pt,top=6.75pt]%
\vspace{1.75pt}%
\textbf{#1}\\[-0.575em]
\noindent\makebox[\linewidth]{\rule{\linewidth}{0.4pt}}\\[0.25em]
#2
\end{tcolorbox}}

\providecommand{\R}{\mathbb{R}}\renewcommand{\R}{\mathbb{R}}
\providecommand{\E}{\mathbb{E}}\renewcommand{\E}{\mathbb{E}}
\providecommand{\tr}{\operatorname{tr}}

\newcommand{\dW}{\Delta W}
\providecommand{\dh}{}\renewcommand{\dh}{\mathcal{F}}

\usepackage{caption}
\usepackage{aliascnt}
\usepackage[nameinlink,noabbrev]{cleveref}
\theoremstyle{plain}

\newaliascnt{proposition}{theorem}
\newtheorem{proposition}[proposition]{Proposition}
\aliascntresetthe{proposition}
\newaliascnt{lemma}{theorem}
\newtheorem{lemma}[lemma]{Lemma}
\aliascntresetthe{lemma}
\newaliascnt{corollary}{theorem}

\aliascntresetthe{corollary}
\theoremstyle{remark}
\newaliascnt{remark}{theorem}
\newtheorem{remark}[remark]{Remark}
\aliascntresetthe{remark}
\crefname{theorem}{Theorem}{Theorems}\Crefname{theorem}{Theorem}{Theorems}
\crefname{proposition}{Proposition}{Propositions}\Crefname{proposition}{Proposition}{Propositions}
\crefname{lemma}{Lemma}{Lemmas}\Crefname{lemma}{Lemma}{Lemmas}
\crefname{corollary}{Corollary}{Corollaries}\Crefname{corollary}{Corollary}{Corollaries}
\crefname{remark}{Remark}{Remarks}\Crefname{remark}{Remark}{Remarks}
\crefname{section}{Section}{Sections}\Crefname{section}{Section}{Sections}
\crefname{figure}{Figure}{Figures}\Crefname{figure}{Figure}{Figures}
\crefname{table}{Table}{Tables}\Crefname{table}{Table}{Tables}
\crefname{equation}{Equation}{Equations}\Crefname{equation}{Equation}{Equations}
\crefname{appendix}{Appendix}{Appendices}\Crefname{appendix}{Appendix}{Appendices}

\DeclareCaptionType[fileext=lrf,placement={tbp}]{rfigure}[Figure][List of rebuttal figures]
\DeclareCaptionType[fileext=lrt,placement={tbp}]{rtable}[Table][List of rebuttal tables]

\crefname{rfigure}{Figure}{Figures}\Crefname{rfigure}{Figure}{Figures}
\crefname{rtable}{Table}{Tables}\Crefname{rtable}{Table}{Tables}
\ifshowupdates
  \newcommand{\rfigtype}{rfigure}
\else
  \renewcommand{\update}[1]{#1}
  \renewcommand{\updatebox}[1]{#1}
  \renewenvironment{rfigure}[1][tbp]{\begin{figure}[#1]}{\end{figure}}
  \renewenvironment{rtable}[1][tbp]{\begin{table}[#1]}{\end{table}}
  \newcommand{\rfigtype}{figure}
\fi

\title{On Potentiality and Actuality Of \\Foundation Models}

\author{Fabian Gr\"oger, Shuo Wen, Maria Brbi\'c \\
EPFL\\
\texttt{\{fabian.groger,shuo.wen,mbrbic\}@epfl.ch}}

\iclrfinalcopy %
\ifarxivstyle
  \date{}
  \renewcommand{\shorttitle}{On Potentiality and Actuality Of Foundation Models} %
  \let\origwrapfigure\wrapfigure\let\endorigwrapfigure\endwrapfigure
  \RenewDocumentEnvironment{wrapfigure}{o m O{0pt} m}{\origwrapfigure{#2}[#3]{#4}}{\endorigwrapfigure}
\fi
\begin{document}
\addtocontents{toc}{\protect\setcounter{tocdepth}{-1}}
\frenchspacing
\maketitle
\renewcommand{\thefootnote}{\fnsymbol{footnote}}

\begin{abstract}
Foundation models are built through large-scale pretraining, often followed by post-training, and are then adapted to specific tasks and behaviors. But, does the adaptation add truly new capabilities, or the base model already holds it?
We decompose the update induced by different adaptations into a component shared across all inputs and a residual that varies with the input. \update{We find that the shared component suffices for most of the adaptation gain and can be learned directly:} a shift adapter that learns a constant vector per layer, identical for every input, reproduces the behavior of LoRA and full fine-tuning. We call this the \textit{potentiality principle} of foundation models: the base model already contains the capacity needed for task-specific expertise. To adapt to a new task, the model just needs to be \textit{actualized}, \textit{i.e.}, pushed in a certain direction via a constant, input-independent shift. We show the implications of this principle on parameter-efficient adaptation, steering, and convergence across models, and how it explains the neural thickets phenomenon.
\end{abstract}

\section{Introduction}\label{sec:intro}

\begin{wrapfigure}[26]{r}{0.45\textwidth}
   \centering
   \vspace{-2.8em}

   \hypbox{Potentiality and Actuality Principle}{%
       \small\raggedright
       Foundation models already contain the task-specific expertise in \textit{potentiality}. Adaptation only \textit{actualizes} this expertise via a constant, input-independent shift.
   }

   \vspace{0.25em}

    \includegraphics[
       width=\linewidth
    ]{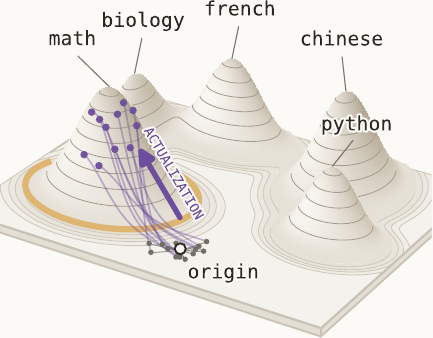}
    \vspace{-2em}
    \caption{\textbf{The Potentiality and Actuality Principle.}
       A foundation model holds task-specific expertise (hills) in \textit{potentiality}. For most tasks, a single shift, same for every input (arrow), is all it takes to move the inputs to one of these tasks. We call this process \textit{actualization}. 
    }
    \label{fig:potentiality}

    \vspace{-1.5\baselineskip}
\end{wrapfigure}

The rapid progress of foundation models \citep{bommasani2021foundation} has been driven by large-scale pre-training  on general-domain data, often followed by post-training and finally adaptation to a specific task, behavior, or domain. To adapt these models effectively, different adaptation strategies have been explored. The earliest form of adaptation is full fine-tuning, which updates all model parameters on a given task \citep{howard2018universal, devlin2019bert}. To reduce computational cost, parameter-efficient methods freeze the model and learn only a small update, such as the widely used low-rank update of LoRA \citep{hu2022lora,li2021prefix,lester2021power}. In-context learning showed that it is possible to adapt by eliminating parameter updates altogether and instead conditioning the model on demonstrations in the prompt \citep{brown2020language}. Similarly, activation steering leaves the model parameters unchanged, but adapts model behavior at inference time by directly modifying hidden activations \citep{turner2023actadd,rimsky2024caa,zou2023repe}. So far, the field has largely treated these adaptation strategies as different tools suited for different settings.

But what is the underlying mechanism of adaptation, and what changes inside a foundation model when it gains task-specific expertise? While these questions remain largely unanswered, a growing body of evidence suggests that task adaptation may rely on small changes to a pre-trained model. In particular, fine-tuning succeeds when the update is restricted to a small parameter space \citep{aghajanyan2021intrinsic,hu2022lora}, or even when it is restricted to biases \update{with a new task layer} \citep{benzaken2022bitfit} or to $0.01\%$ of the weights \citep{panigrahi2023skill}. Recent work on neural thickets even showed that random perturbations of pre-trained weights find dense sets of diverse task experts \citep{gan2026thickets}. Why such small interventions are sufficient to produce task-specific expertise remains poorly understood. Previous work has suggested that we are in the elicitation overhang: AI models have untapped capabilities, and we do not know how to elicit them from frontier models \citep{turner2023actadd}.

To understand the adaptation mechanism, we first decompose the update that LoRA learns into two components: \textit{(i)} a component shared across all inputs, and \textit{(ii)} an input-specific component, \textit{i.e.}, a residual that varies with the input. 
Across many diverse tasks, we find that \update{most of the adaptation gain can be reached without input-specific adaptation}: the update vectors point in nearly the same direction independent of the input. Quantitatively, on 26 benchmarks and 18 language models from 7 architecture families, the component shared across all the inputs preserves $85\%$ of the adapted behavior. %
Motivated by this finding, we adapt the model via \textit{shift adapter}, a simple adaptation mechanism that learns only the shared part of adaptation: a single vector per layer, constant for every input. \update{Since the update a shift adapter learns is added unchanged to every input, unlike the update of LoRA, bias-only training, or full fine-tuning, every difference between the answers to two inputs is computed by the frozen base model.} The shift adapter matches LoRA's performance across 17 tasks and 18 language models, with roughly two orders of magnitude fewer parameters. The same holds for adaptation without the rank constraint and for full fine-tuning, where the savings are even larger.

These findings suggest that the base model itself already has the capacity for diverse task-specific expertise and only a small push in the right direction is needed to put this capacity to work. We call this the \textit{potentiality principle} of foundation models, and the push towards a direction to gain task-specific expertise the \textit{actualization} of a foundation model. The terms reflect Aristotle's distinction between \emph{potentiality} (\emph{dunamis}), a capacity that a thing holds without exercising it, and \emph{actuality} (\emph{energeia}), that same capacity at work \citep{aristotle}. In this terminology, a pre-trained foundation model holds different task-specific expertise as potentiality, and that expertise can be actualized via a constant shift that moves the model into a state where it can be exercised (\cref{fig:potentiality}). 

The potentiality and actuality principle has implications on \textit{fine-tuning}, \textit{steering}, \textit{neural thickets}, and \textit{model convergence}. It provides an explanation for why fine-tuning can succeed in a tiny parameter space \citep{aghajanyan2021intrinsic, hu2022lora, benzaken2022bitfit, sinii2025biasonly}, identifies the trained shift as a reference direction for steering \citep{turner2023actadd, rimsky2024caa, zou2023repe}, relates the dense task experts found by random perturbations to states that actualize existing capacities \citep{gan2026thickets}, and shows that the states of different models grow alike as the models gain capability \citep{huh2024platonic,groeger2026aristotelian}. 
In each case the mechanism is the same: a state that actualizes a capacity the model already has.

\section{Shared and input-specific components of adaptation}\label{sec:field}

Imagine that a foundation model already holds the task-specific expertise in \textit{potentiality} and that adaptation only has to \textit{actualize} it by selecting the desired expertise.
In such a case, the selection would be the same for every input of the task, \textit{i.e.}, the update learned during adaptation would not vary across inputs, or would vary only slightly. 

To test this hypothesis, we split the change that adaptation does into two components: \textit{(i)} a component shared by all the inputs, and \textit{(ii)} an input-specific component. We measure how much of each component changes during adaptation, and evaluate the model using only one of the two components to determine which component the adaptation relies on.
We explore this for different kinds of adaptation: LoRA \citep{hu2022lora}, the same update trained without the rank constraint (the \emph{unconstrained update}), full fine-tuning \citep{howard2018universal,devlin2019bert}, and in-context learning \citep{brown2020language}.

\textbf{Adaptation field.} Every form of adaptation changes the hidden states that the model computes for an input, whether it updates the weights or places demonstrations in the prompt.
We thus describe adaptation through this change.
Let $h^{\mathrm{base}}_\ell(x)\in\R^{d}$ and $h^{\mathrm{adapted}}_\ell(x)\in\R^{d}$ be the hidden states after layer $\ell$ of the base and the adapted model for input $x$, where $d$ is the hidden size.
We define the \emph{adaptation field} at layer $\ell$ as their difference and decompose it into a component shared by all inputs, its mean, and an input-specific component,
\begin{equation}\label{eq:field}
\dh_\ell(x)=h^{\mathrm{adapted}}_\ell(x)-h^{\mathrm{base}}_\ell(x)
=\underbrace{\bar\dh_\ell}_{\text{shared component}}+\underbrace{\widetilde\dh_\ell(x)}_{\text{input-specific component}},
\qquad
\bar\dh_\ell=\E_x[\dh_\ell(x)].
\end{equation}
The shared component $\bar\dh_\ell$ is one vector per layer, the same for every input.
The input-specific component $\widetilde\dh_\ell(x)$ has zero mean over inputs, so the two components add in energy (\cref{prop:decomp}),
\begin{equation}\label{eq:energy}
\E_x\lVert\dh_\ell(x)\rVert^2=\lVert\bar\dh_\ell\rVert^2+\E_x\lVert\widetilde\dh_\ell(x)\rVert^2.
\end{equation}

\begin{wrapfigure}[12]{r}{0.36\linewidth}
\vspace{-1.6em}
\centering
\includegraphics[width=\linewidth]{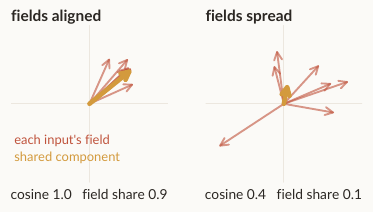}
\vspace{-2.0em}
\caption{\textbf{The field statistics.} Fields that point the same way (left) have a high cosine to their shared component and a high field share. Fields that spread (right) have neither.}
\label{fig:field-measure}
\vspace{-1.5\baselineskip}
\end{wrapfigure}

We quantify the decomposition with two statistics (\cref{fig:field-measure}, \cref{app:field-stats}): \textit{(i)} the share of the field's energy that the shared component holds defined as $\lVert\bar\dh_\ell\rVert^2/\E_x\lVert\dh_\ell(x)\rVert^2$ (the first term of \cref{eq:energy} over its left side), which we call the \emph{field share}, and \textit{(ii)} the average cosine between each input's field $\dh_\ell(x)$ and the shared component $\bar\dh_\ell$. The cosine similarity is close to 1 if adaptation moves every input in the same direction and close to 0 if it moves each input differently.

\textbf{Low-Rank Adaptation (LoRA).} In LoRA adaptation, a low-rank update $\dW=BA$ is added to each weight matrix $W\in\R^{d_{\mathrm{out}}\times d_{\mathrm{in}}}$, which maps a $d_{\mathrm{in}}$-dimensional input to a $d_{\mathrm{out}}$-dimensional output, with $B\in\R^{d_{\mathrm{out}}\times r}$, $A\in\R^{r\times d_{\mathrm{in}}}$ and rank $r\ll\min(d_{\mathrm{in}},d_{\mathrm{out}})$.
We adapt every linear projection in every layer, the common choice \citep{dettmers2023qlora,biderman2024lora}, and consider one adapted matrix $W$ at a time, since the derivation is the same for each.
Let $z_t(x)\in\R^{d_{\mathrm{in}}}$ be the hidden state that $W$ multiplies at token position $t$ for input $x$.
The base model computes $Wz_t(x)$ and the adapted model $(W+\dW)z_t(x)$, so what LoRA adds at this matrix is the vector $a_t(x)=\dW z_t(x)\in\R^{d_{\mathrm{out}}}$.
Unless stated otherwise, we take $z_t$ at the last token of the input, where the model has read the entire input and is about to answer, and drop the index $t$.
Because $\dW$ is linear, the added vector can be split into the shared and input-specific component:
\begin{equation}\label{eq:decomp}
a(x)=\dW z(x)=\underbrace{\dW\mu}_{\text{shared component}}+\underbrace{\dW\bigl(z(x)-\mu\bigr)}_{\text{input-specific component}}=\bar a+\widetilde a(x),
\end{equation}
where $\mu=\E_x[z(x)]\in\R^{d_{\mathrm{in}}}$ is the average hidden state estimated on held-out inputs the adapter has never been trained on.

\textbf{Experimental setup.} We apply the decomposition to trained LoRAs of rank $8$, on 18 models from 7 architecture families: 14 instruction-tuned, 3 base models, and 1 trained on instructions but not preference-tuned (\cref{app:models}).
Each model is adapted on 26 benchmarks from two categories: \textit{(i)} 17 reasoning and knowledge tasks with labeled answers which test what the model knows and how it reasons, and \textit{(ii)} 9 behaviors including 4 traits which ask for the answer in a given language, persona, or format, 3 transformations which ask the model to apply a specific rule to the input, and 2 maps, a letter cipher and a lookup table, which assign each input an answer by an arbitrary rule. Nothing in pretraining fixes that rule, so the maps are the one category the model can only learn from the adaptation data. We refer to the first category as \textit{tasks}, and the second as \textit{behaviors}. The tasks span commonsense reasoning, text classification, subject-specific question answering, and recent complex reasoning benchmarks (e.g., BoolQ, HellaSwag, MMLU-Pro, and SuperGPQA), following the protocol of \citet{hu2023llmadapters,liu2024dora} (details \cref{app:tasks}). Tasks are evaluated by accuracy on their evaluation splits, while behaviors are evaluated on $1{,}000$ held-out generations by adherence to the target behavior. For each intervention, we report the fraction of the adaptation gain over the base model that is retained. Unless stated otherwise, results are averaged over 3 seeds.

\begin{figure}[t]
\centering
\includegraphics[width=\textwidth]{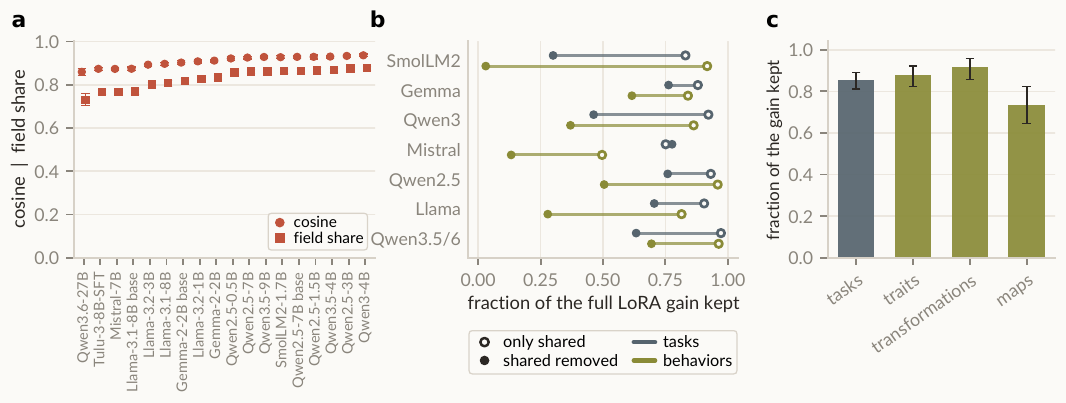}
\vspace{-2em}
\caption{
\textbf{LoRA adaptation mainly relies on the component shared across all inputs.}
\textbf{(a)}~Two decomposition statistics for LoRA: each input's field points along the shared component (cosine), and the shared component holds most of the field's energy (field share). Each point is the model's mean across 17 tasks, and the bars are $95\%$ bootstrap intervals over the runs, too small to be visible.
\textbf{(b)}~The fraction of the full LoRA's gain kept using only the component shared across all inputs (hollow) and with the shared component removed (filled), per model on the 17 tasks and on the 9 behaviors. Rows are the largest instruction-tuned model of each family: SmolLM2-1.7B, Gemma-2-2B, Qwen3-4B, Mistral-7B, Qwen2.5-7B, Llama-3.1-8B and Qwen3.6-27B.
\textbf{(c)}~The fraction of LoRA's gain kept using only the component shared across all inputs, for the 17 tasks, and 9 behaviors (4 traits, 3 transformations, and 2 maps, cipher and lookup) (\cref{app:interventions-bench}). Each bar pools the runs of 18 models, and the error bars are $95\%$ bootstrap intervals over models.
Results for all models and tasks are in \cref{app:interventions-bench} (\cref{fig:shared-all}).}
\label{fig:shared}
\vspace{-10pt}
\end{figure}

\textbf{Observation.} Across 17 tasks and 18 models, each input's field points along the shared component at a cosine of $0.91$ on average, and the shared component holds $83\%$ of the field's energy (\cref{fig:shared}a, all models in \cref{app:interventions-bench}). 
Across the 9 behaviors and 18 models, the cosine is $0.84$ and the shared component holds $71\%$ of the field's energy (\cref{app:interventions-behaviors}). The same decomposition applies to the adapter's own output $a(x)=\Delta W z(x)$, before it propagates through the model. There the shared component $\bar a$ holds $89\%$ of the energy on these behaviors \update{and $89\%$ on the tasks}, averaged over every adapted matrix (\cref{app:interventions-update}). The fields of different inputs thus point in nearly the same direction: LoRA moves every input mostly the same way, and the part of the change that depends on the input holds the smaller share of the energy on both tasks and behaviors.

\textbf{Interventions.} We next explored which component the LoRA gains come from using two interventions: \textit{(i) }keeping only the shared component, and \textit{(ii)} removing the shared component. In the first intervention, the same shared vector $\bar a$ replaces $a(x)$ in \cref{eq:decomp} at every adapted matrix and at every position so the adapter becomes one constant vector per matrix. In the second intervention, $\bar a$ is subtracted at every position, so the model relies only on the input-specific component $\widetilde a(x)$.
The former tests whether the shared component alone produces the adapted behavior on new inputs, while the latter tests whether the trained adapter depends on the shared component.

Over all 18 models, keeping only the shared component keeps on average $85\%$ of the adapter's gain on the tasks, and removing it keeps $61\%$ (\cref{fig:shared}b, \cref{fig:shared-all}). \update{Two other ways of removing it keep $84\%$ and $75\%$ (\cref{app:interventions-bench}), so on the tasks each component alone can produce most of the gain.}
All models keep more with the shared component alone than without it. The only exception is Mistral-7B, which \update{relies on both components about equally}.
The 3 models that never received instruction tuning keep $85\%$ of the gain with only the shared component and $59\%$ with it removed, against $87\%$ and $63\%$ on the instruction-tuned models, so the reliance on the shared component is not a product of instruction tuning (\cref{app:base-models}).
The same holds for behaviors, where the shared component alone keeps $86\%$ of the gain on average, and removing it keeps $33\%$ on the traits, $57\%$ on the transformations, and $26\%$ on the maps. The update matters while the model reads the input, not while it writes the answer: with the adapter switched on only for the input tokens, LoRA keeps $83\%$ of its gain, and only for the answer tokens $12\%$, on 18 models and 6 tasks (\cref{app:phase}). So most of what LoRA gains comes from one vector per matrix that is the same for every input, added while the model reads the input.

We next ask whether adaptation still relies on a shared component when producing the correct behavior requires input-specific computation. The 3 transformations, such as copying the input or returning its last word, apply a rule to the input that the base model can already follow, so the computation that varies with the input is the model's own.
Surprisingly, even there, keeping only the shared component keeps $92\%$ of the gain, as much as on the 4 traits, which ask for the same behavior on every input ($88\%$, \cref{fig:shared}c). 
The 2 maps apply a rule fixed by an arbitrary key, a substitution of the first letter, or a table of words, which the model can only learn during adaptation, and they keep the least, $74\%$ of adaptation gain. 
This indicates that gains of the shared component fall when the base model does not hold the rule. On the maps, what adaptation adds is the table, and what the base model provides is the ability to use one: a shift adapter trained on a map answers as well as LoRA does, and the table it stores produces nothing on its own (\cref{app:express}).

\textbf{Unconstrained update.} To investigate the influence of the low-rank update on the decomposition result, we train the same update without the low-rank factorization, so that $\dW$ is a full matrix at every adapted matrix, on the same 18 models, the 4 traits, and the 3 transformations.
After training, the update has high rank (median stable rank $47$\footnote{Stable rank is $\lVert W\rVert_F^2/\lVert W\rVert_2^2$, the squared Frobenius norm over the squared spectral norm.}) and raises adherence as much as LoRA does ($0.44$ against $0.43$ above the base model), yet its shared component holds $95\%$ of the energy of what it adds, against LoRA's $90\%$.
Without the shared component, it keeps $26\%$ of the adaptation gain on traits and $49\%$ on transformations, against LoRA's $39\%$ and $59\%$. With the shared component alone, it keeps $69\%$ against LoRA's $87\%$ on traits (\cref{app:rankfree}).
The shared component is therefore not a product of the low-rank constraint: with the constraint gone, it grows from $90\%$ of the energy to $95\%$, while the behavior relies on it slightly less.

\textbf{Fine-tuning and ICL.} 
We further examine whether full fine-tuning and in-context learning (ICL) rely on the shared component in the same way LoRA does. 
Full fine-tuning changes every weight, so its update cannot be decomposed and intervened on as LoRA's. However, we can examine the structure of its adaptation field: we find that adaptation via fine-tuning results in nearly the same shared structure as LoRA. Across the 17 tasks and 18 models, its cosine is $0.90$, and its field share is $0.81$ on average, compared to $0.91$ and $0.83$ for LoRA (\cref{app:parity-fullft}), suggesting that all inputs are moved in the same direction, just like we observe for LoRA. 
In contrast, ICL relies more on input-dependent changes: its shared component keeps less of the gain than LoRA's on traits, transformations and tasks, and its shared component remains nearly orthogonal to LoRA's even as their induced hidden-state changes become more aligned in later layers (\cref{app:icl}). This suggests that ICL reaches similar behaviors through a different, more input-dependent route. Additional results are in \cref{app:components}.

\section{Actualizing with a shift adapter}\label{sec:parameterizations}

\begin{figure}[t]
\centering
\includegraphics[width=\textwidth]{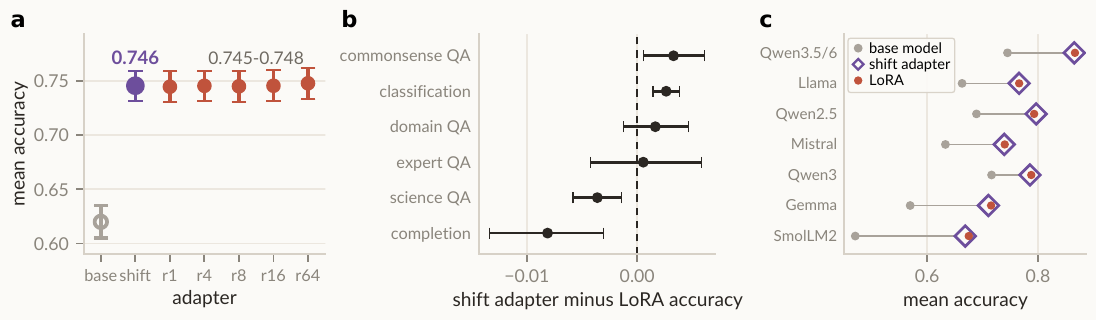}
\vspace{-2em}
\caption{\textbf{Shift adapter matches LoRA's performance.}
\textbf{(a)}~Mean accuracy of the base model, shift adapter, and LoRAs with different ranks. The mean is computed across 18 models and 17 tasks, with bootstrap $95\%$ intervals over the runs.
\textbf{(b)}~The paired difference between the shift and the rank-8 LoRA per task type, over 18 models and 17 tasks. Error bars are bootstrap $95\%$ intervals over the runs.
\textbf{(c)}~The largest instruction-tuned model in each family, ordered by scale, SmolLM2-1.7B to Qwen3.6-27B: mean accuracy of the base model, the shift, and the rank-8 LoRA, over 17 tasks. 
All models and tasks are shown in \cref{app:parity}.}
\label{fig:parity}
\vspace{-10pt}
\end{figure}

In the previous section, we showed that we can preserve most of the adaptation gain by keeping only the component shared across the inputs. If this shared component is indeed sufficient to produce the adapted behavior or solve a task, then we should be able to learn such a change directly.
We therefore define a \emph{shift adapter}: one learned vector $b_\ell\in\R^d$ per layer $\ell$, added to the hidden state of every input at every position.
Its only operation is a constant push.
The shift applies the same push to every input and moves the model into a state in which its expertise for one task is actualized. We call such a state an \emph{actualizing state}. %
Note that a constant parameter does not mean a constant effect: the layers above process the push together with an input-dependent hidden state, so the same push can move different inputs differently, and they can reach different actualizing states. However, the capacity to move inputs differently lies in the base model, and the shift only actualizes it via a constant push. \update{This sets the shift adapter apart from LoRA, bias training and full fine-tuning, which can change the output of a block differently for each input (\cref{app:pf-bias}).}
\update{This gives an operational distinction: a capacity is actualized when a constant vector suffices on inputs never seen in training, and memorized when the gain holds only on the items seen in training (\cref{app:few-items,app:conditions}).}

\textbf{Shift adapter matches LoRA.} We compare the shift adapter to LoRAs of rank $\{1,4,8,16,64\}$ on the 17 tasks across 18 models, with the training and evaluation protocol of \cref{app:parity}.
The shift reaches a mean accuracy of $0.746$, against $0.745$ for a LoRA of rank 8 and $0.620$ for the base model (\cref{fig:parity}a), recovering all of the LoRA's gain, with $L\times d$ trainable parameters, $0.04$M on Qwen2.5-1.5B compared to $9.2$M for LoRA with rank 8.
The same holds at every rank: from rank $1$ to rank $64$ the LoRA's accuracy moves by $0.003$ while its parameters grow from $1.2$M to $74$M on that model.
The shift stays within $0.012$ of the rank-8 LoRA on any task and within $0.011$ on any model, from $0.5$B to $26$B parameters (\cref{fig:parity}b,c).
Training all weights reaches $0.718$ under the protocol above, so the shift adapter is above full fine-tuning on every model.

\begin{wrapfigure}[18]{r}{0.50\linewidth}
\vspace{-1.5em}
\centering
\includegraphics[width=\linewidth]{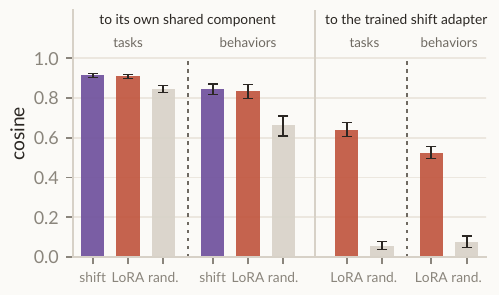}
\vspace{-2.0em}
\caption{
\textbf{Independent adaptations push the same way.}
Mean over runs with $95\%$ intervals over models. Left, the cosine of each input's field to its adapter's own shared component, for the shift, the LoRA and a LoRA with random weights, on the 17 tasks and on the 9 behaviors. Right, the cosine of the LoRA's shared component (LoRA and random LoRA) to the shift adapter, on the 17 tasks and on the 9 behaviors.
}
\label{fig:alignment}
\end{wrapfigure}

\textbf{The shift and LoRA agree on the actualizing direction.}
The shift adapter and LoRA are trained independently, so we next wondered whether they push the inputs along the same direction and finally reach the same actualizing state.
Across the 17 tasks and 18 models, the average cosine similarity of LoRA's shared component and the shift adapter is $0.64$ on average, while a LoRA with random weights gives $0.06$ (\cref{app:agreement}). Across the 9 behaviors and 18 models, the cosine similarity is $0.53$ on average, while a LoRA with random weights gives $0.07$ compared to the shift adapter (\cref{fig:alignment}). Every bar sits above its random counterpart: a LoRA with random weights reaches $0.84$ on the tasks and $0.66$ on the behaviors for the cosine to its own shared component, and only $0.06$ and $0.07$ for its cosine to the trained shift (\cref{fig:alignment}).
If, instead of weighting each layer equally as before, we weight each layer by how far the shift adapter moves it, cosine similarity rises to $0.69$, and the random LoRA remains at $0.08$ (\cref{app:field-stats}).
The comparable performance and similar direction of the shift adapter and LoRA indicate that LoRA relies on the direction the shift adapter learns, even though it can vary its update with every input. Additional results are in \cref{app:methods}. %

\section{The structure of actualization}\label{sec:structure}

\begin{wrapfigure}[18]{r}{0.40\linewidth}
\vspace{-4.5em}
\centering
\includegraphics[width=\linewidth]{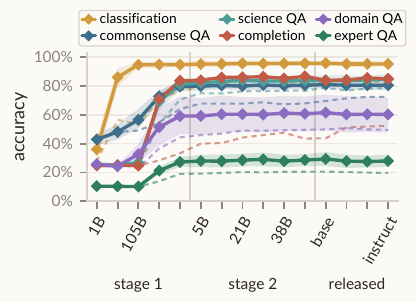}
\vspace{-2.0em}
\caption{\textbf{The task-specific expertise forms during pretraining, and a shift actualizes it whenever it has formed.}
OLMo-2-7B at $15$ checkpoints on the 17 tasks of \cref{fig:parity}.
Accuracy of the shift (solid) and of the base model with no adaptation (dashed) per task type: the 3 text classifications, the 4 commonsense question-answering tasks, the 3 science question-answering tasks, the 4 domain tasks, sentence completion, and the 2 expert question-answering tasks.
Bands are $95\%$ intervals. The per-task view is in \cref{app:development}.}
\label{fig:stages}
\vspace{-1.5\baselineskip}
\end{wrapfigure}

All these results indicate that the base model already holds task-specific expertise in potentiality, and adaptation can rely on a constant input-independent push to reach the state in which task-specific expertise can be actualized.
Together they form the two parts of the \textit{potentiality and actuality principle}: \textit{(i)} the base model holds the task-specific expertise in potentiality, and \textit{(ii)} a constant shift can actualize the task-specific expertise of the base model.
In this section, we aim to better understand this principle. In particular, we aim to understand when, during training, the potentiality forms in the base model, and what geometry the actualizing region has.

\textbf{When during training potentiality forms.} If the base model holds the task-specific expertise in potentiality and a shift actualizes it, when during training does the potentiality form?
We investigate this question in OLMo-2-7B \citep{olmo2024furious}, which has intermediate public checkpoints.
Specifically, we look at 15 checkpoints spanning its training: 5 from stage-one pretraining on Dolma \citep{soldaini2024dolma} between $1$B and $3{,}896$B tokens, 6 from the $50$B-token stage two, and 4 released base, supervised, preference-tuned, and instruct models.
At every checkpoint we repeat the comparison of \cref{sec:parameterizations} and the decomposition of \cref{sec:field}, on the 17 tasks (\cref{app:development}).

The analysis shows that the potentiality already forms during pretraining, at different points for different kinds of tasks (\cref{fig:stages}).
Text classification comes first, the shift adapter raises the base model's accuracy on the 3 classifications from $0.57$ to $0.86$ already at $9$B tokens.
Science question answering, the domain tasks, sentence completion, and expert question answering arrive far later, and until $105$B tokens the adapted model does no better than answering with the most frequent label.
So the potentiality forms in the base model during stage-one pretraining, at different points for different kinds of tasks, and once a kind of task-specific expertise has formed, the shift actualizes it at every later checkpoint, through stage two and post-training alike.

From $9$B tokens on, the shift and the rank-8 LoRA reach accuracy within $0.01$ of each other at every checkpoint, so the shift matches LoRA once the expertise has formed. Only at $1$B tokens does the rank-8 LoRA score slightly higher than the shift (\cref{app:development}). 
For behaviors, the potentiality principle forms later: at $9$B tokens a rank-8 LoRA produces \update{$40$ to $42$} of the $42$ writing systems \update{(non-Latin scripts such as Thai)}, where a shift produces \update{$7$ to $8$}, and the two meet only at $105$B tokens (\cref{app:ladder-behaviors}\update{,~\ref{app:existence}}). \update{The shift adapter fails at $9$B tokens because the capacity has not yet formed: at this checkpoint the trained LoRA itself does not rely on its shared component, which alone produces none of the writing systems, while at $105$B tokens the same shared component alone produces $37$ to $40$ of them and a shift adapter $38$ to $41$ (\cref{app:existence}). A shift adapter thus reaches a behavior from the point at which pretraining has formed the capacity, and LoRA starts to rely on its shared component from about the same point.}
Pythia-12B shows the same order across its pretraining checkpoints (\cref{app:ladder-further}). 
\update{On the 17 tasks, the LoRA} relies on its shared component at every checkpoint: with only that component it keeps between $70\%$ and $93\%$ of its gain at the 15 checkpoints, from $1$B tokens to the instruct model (\cref{tab:ladder}). So pretraining forms the task-specific expertise, and adaptation actualizes it through the component shared by all inputs, which remains the same from the first checkpoint on.

\begin{wrapfigure}[21]{r}{0.36\linewidth}
\vspace{-1.2em}
\centering
\includegraphics[width=\linewidth]{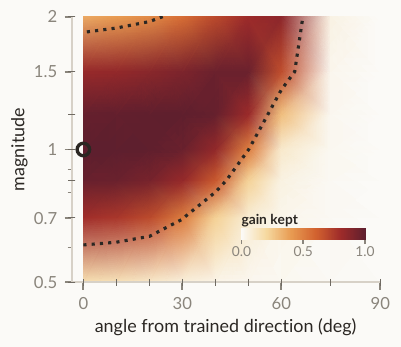}
\vspace{-2.0em}
\caption{
\textbf{The region around a trained state.}
The fraction of the trained shift's gain kept as the shift is scaled (vertical) and rotated away from the trained direction (horizontal), median over the 9 behaviors of the setup on 18 models. The circle is the trained shift, and inside the dotted line at least half of the gain is kept. Ticks mark the 7 magnitudes and 9 angles measured.
}
\label{fig:region-inline}
\vspace{-0.5\baselineskip}
\end{wrapfigure}

\textbf{What geometry does the actualizing region have.}
We characterize an actualizing region by three properties: \textit{(i)} width in magnitude, \textit{(ii)} width in direction, and \textit{(iii)} connectivity. Across the 9 behaviors of the setup on 18 models, we measure width in magnitude by scaling a trained shift and width in direction by rotating it away from its trained direction. The region is narrow in magnitude but wider in direction: at least $90\%$ of the gain is preserved between $0.80\times$ and $1.34\times$ the trained magnitude and up to a $34^\circ$ rotation, while half is preserved between $0.61\times$ and $1.86\times$ and up to $50^\circ$ (\cref{fig:region-inline}, \cref{app:basin,app:edge}). Directional width depends on what the shift is rotated toward: on the 4 traits, rotating toward another trait's shift reduces the $90\%$ and $50\%$ thresholds to $13^\circ$ and $26^\circ$, respectively (\cref{app:rotate}).
We measure \textit{connectivity} by interpolating between independently trained states for the same behavior.
Specifically, we train the same behavior at 3 seeds, take two of the resulting states, and score the shift at 5 equally spaced points on the straight line between them. 
Across the 9 behaviors and 18 models, the lowest point along the interpolation retains a median $99\%$ of the endpoint gain, indicating that the two states lie in a single connected region, with the gain remaining high along the entire interpolation path. %
Connectivity also holds when the two states are trained from different supervision sources: one from the model's system-prompted answers and the other from externally translated answers, across French, German, and Spanish. This shows that actualizing states of the same behavior lie in connected regions, not in separate pockets (\cref{app:region}). We next assessed whether the size of the region correlates with the performance. Indeed, larger measured regions show stronger behaviors: across the 9 behaviors and 18 models, region area correlates with adherence at the center at within-model Spearman $+0.64$ (\cref{app:region-size}).

\begin{figure}[t]
\centering
\updatebox{\includegraphics[width=\dimexpr\textwidth-8.4pt\relax]{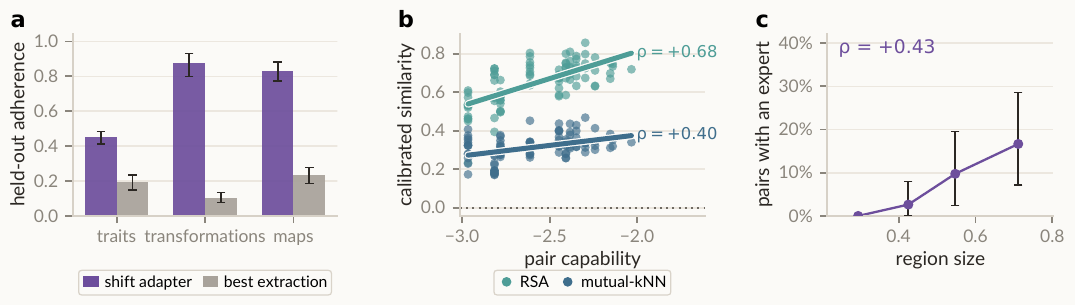}}
\vspace{-0.6em}
\caption{\textbf{Three implications of the potentiality and actuality principle.}
\textbf{(a)}~Steering: held-out adherence of the trained shift adapter (purple) and of the best of the three extracted vectors (grey), pooled over the behaviors of each class of the setup on 18 models. Bars are $95\%$ intervals over models. Every behavior on its own is in \cref{app:steerstress}.
\textbf{(b)}~Convergence across models: calibrated similarity of the sets of states of two models against the capability of the pair, for RSA and mutual-$k$NN, over all pairs of the 14 models on the 75 shared behaviors (Details in \cref{app:crossmodel}). The capability of a pair is the lower of the two models' WikiText log-likelihood per token, so a more capable pair sits further right.
\update{\textbf{(c)}~Neural thickets: the share of model--behavior pairs on which the random search finds an expert, by quartile of the size of the actualizing region measured in \cref{sec:structure} (the share of the magnitude-by-angle grid on which a perturbed shift keeps half the trained gain), over the 9 behaviors and 18 models, with $95\%$ bootstrap intervals. Every model is in \cref{app:thickets}.}}
\label{fig:implications}\label{fig:steering}\label{fig:scaling}
\vspace{-14pt}
\end{figure}

\section{Implications of the potentiality and actuality principle}\label{sec:implications}\label{sec:impact}

We examine implications the potentiality and actuality principle has on \textit{(i)} steering, \textit{(ii)} convergence, and \textit{(iii)} neural thickets. 

\textbf{Implications on steering.} 
Activation steering~\citep{subramani-etal-2022-extracting} provides a direct example of the potentiality and actuality principle: adding an appropriate direction in activation space can actualize a desired behavior without changing the model's weights.
However, the challenge for steering is to identify the right actualizing direction.
Existing steering methods extract such directions from model activations~\citep{rimsky2024caa,zou2023repe,beaglehole2026universal}, but the resulting interventions can be sensitive to how the direction is constructed and how strongly it is applied.
Instead of extracting the direction heuristically, the shift adapter learns the actualizing shift directly from behavioral examples.
We compare three extracted vectors, the mean difference of CAA \citep{rimsky2024caa}, the leading principal component of the same differences \citep{zou2023repe}, and the in-context task vector \citep{hendel2023taskvectors}, with the shift adapter on the 9 behaviors and 18 models. 
Following standard practice \citep{rimsky2024caa,zou2023repe}, we choose the strength on $200$ held-out instructions and evaluate all methods on $800$ further instructions. 
The shift adapter reaches an adherence of $0.68$ against $0.17$ for the best of the three extracted vectors, on average over 18 models with intervals over models, and it leads in every class: $0.45$ against $0.20$ on the traits, $0.87$ against $0.10$ on the transformations, and $0.83$ against $0.23$ on the maps (\cref{fig:implications}a). 
We further evaluate on AxBench \citep{wu2025axbench}, where the shift adapter scores $0.87$ compared with $0.21$ for CAA on Gemma-2-2B (\cref{app:axbench}).
In this view, activation steering is one way of actualizing what the model already holds. 
However, it often fails to find the right actualizing state. 
The shift adapter instead reliably actualizes the desired state.

\textbf{Implications on convergence.} 
The Platonic representation hypotheses \citep{huh2024platonic}
observes that the representations of different models grow alike as the models grow more capable. The potentiality and actuality principle raises the same question for the states a model can be brought to: do the actualizing states of different models converge, and does the agreement grow with capability? A state is named by the behavior it produces, so the state for answering in French in one model has a partner in every other model that answers in French, and we can compare how two models arrange their states just as we compare how they arrange their representations \citep{kriegeskorte2008rsa,kornblith2019cka}. We train one shift adapter for each of the $75$ behaviors that $14$ instruction-tuned models all reach, each behavior with a name and a system prompt (languages, output forms of the kind IFEval verifies \citep{zhou2023ifeval}, personas of the kind persona vectors control \citep{chen2025persona}, and eras, registers, affects and audiences, \cref{app:setup-depth}), and compare the arrangements of the $75$ states between all model pairs with RSA \citep{kriegeskorte2008rsa}, which compares the rank order of the angles between states, CKA \citep{kornblith2019cka}, which compares their inner products up to a rotation and a scale, and mutual-$k$NN \citep{huh2024platonic}, which compares which states are nearest neighbors. To know what agreement two unrelated arrangements show, we shuffle which behavior is which in one of the two models and score the shuffled pair the same way \citep{groeger2026aristotelian}. The states of different models show high agreement: RSA between models of different families is $0.64$, every pair lies above its shuffled counterpart ($p<0.01$), and the more capable the two models, the closer their arrangements under RSA (Spearman $\rho=0.68$, $p=0.003$) and mutual-$k$NN ($\rho=0.40$, $p=0.03$), while CKA does not rise with capability (\cref{fig:implications}b, \cref{app:crossmodel}). In this view, models converge not only in what they compute for a given input but in what they hold in potentiality, and they converge more the more capable they become.

\textbf{Implications on neural thickets.} 
\citet{gan2026thickets} recently showed that adding random perturbations to the weights of the large pretrained model can yield task-improving experts. 
In large models, the density of such experts is high: a substantial fraction of random weight perturbations improve performance on a task. 
However, why such task-improving experts become abundant around pretrained weights is an open question. 
The potentiality and actuality principle offers a way to explain this phenomenon: an expert is a perturbation that happens to land in the actualizing region of the task, \textit{i.e.}, the region in which the behavior can be actualized. 
A perturbation of fixed size lands in the actualizing region and finds experts by chance only when the actualizing regions are wide, and none when the regions are narrow. 
To test this, we first reproduce the random-search protocol of \citet{gan2026thickets} on GSM8K \citep{cobbe2021training} and then run it on the 9 behaviors of our setup, whose actualizing regions we measured in \cref{sec:structure}.
Following \citet{gan2026thickets}, we sample $300$ random weight perturbations at several noise scales, keep the $20$ that score highest on a selection set, and evaluate those on a disjoint held-out set. We call the share of the $20$ that still improve the behavior on the held-out set the expert fraction. On the GSM8K the search finds experts in \update{12} of the 18 models, with an expert fraction of \update{$0.20$} on average.
Across model-behavior pairs, expert density increases with actualizing-region size. In the 8 models where the search finds an expert, the behaviors with an expert have wider regions than those without: region size and expert density correlate at Spearman \update{$\rho=0.43$} over the 72 model--behavior pairs ($p<0.001$, \cref{app:thickets}), and the share of pairs with an expert rises from $0\%$ in the narrowest quarter of regions to \update{$17\%$} in the widest (\cref{fig:implications}c). The three behaviors with the narrowest regions have no expert in any model. 
Thus, neural thickets can be understood as random search entering actualizing regions by chance: the wider the region, the more likely it is to contain randomly discovered experts.

\section{Related work}\label{sec:related}
\vspace{-1mm}

The potentiality and actuality principle connects different adaptation strategies through the states they induce. More details are in \cref{app:related}.

\textbf{Parameter-efficient and representation fine-tuning.}
Parameter-efficient methods restrict adaptation to low-rank updates \citep{hu2022lora}, biases \update{with a new task layer on linguistic classification tasks} \citep{benzaken2022bitfit}, updates over frozen factors \citep{zhang2023lorafa}, or updates restricted to particular subspaces \citep{gao2024fourierft}. 
Representation-based methods instead modify hidden states directly, from input-dependent low-rank updates in ReFT \citep{wu2024reft} to fixed offsets at selected attention heads in LoFiT \citep{yin2024lofit}. \update{Each of these updates can change the output of a block differently for each input: the biases of \citet{benzaken2022bitfit} sit inside the block, so their change passes through attention and the MLP (\cref{app:pf-bias}), while the shift adapter adds its vector after the block. Moreover, 13/18 models we evaluate have no biases at all, making bias training not applicable. \citet{sinii2025biasonly} train one vector per layer with reinforcement learning for mathematical reasoning and see its input-independence as a limitation, while we find that most of what LoRA adds is the same for every input.} The potentiality and actuality principle shows that a constant, input-independent shift is sufficient to move the model into a state in which task-expertise can be actualized.

\textbf{In-context learning.}
In-context learning adapts through demonstrations \citep{brown2020language}. Prompt tuning, prefix tuning, and gisting learn that conditioning instead of supplying it \citep{lester2021power,li2021prefix,mu2023gisting}. Our comparison reveals a different route from LoRA: the demonstrations' shared component keeps less of the gain than LoRA's on traits, transformations and tasks, and is nearly orthogonal to LoRA's shared component.

\textbf{Activation steering.}
Activation steering modifies hidden states using directions extracted from activation contrasts \citep{subramani-etal-2022-extracting,rimsky2024caa} or their principal components \citep{zou2023repe}. We compare extracted directions with the shift adapter added at the same place in the model and find it reaches behaviors that steering misses.

\textbf{From weight updates to actualizing states.}
\citet{wang2025oocr} show that single-layer LoRA updates in out-of-context reasoning point in nearly the same direction across tokens, and that extracted or trained steering vectors can induce the learned behavior. \citet{adila2026activation} derive a local first-order relation between weight updates and activation shifts and build post-block activation adapters. 
Our paired intervention tests this correspondence in trained, multi-layer LoRAs: keeping only the shared component tests its sufficiency, while removing it tests the update's reliance on it.

\section{Conclusion}\label{sec:conclusion}
\vspace{-1mm}

In this paper, we explore what adaptation does to a foundation model. 
By disentangling the shared and input-dependent components of adaptation, we find the \textit{potentiality and actuality principle}: adaptation primarily actualizes expertise already present in the model.
\update{The shared component suffices for most of the adaptation gain and can be learned directly:} a shift adapter that trains a single input-independent vector per layer reproduces the behavior of LoRA and full fine-tuning. 
This principle provides a common perspective on parameter-efficient fine-tuning, activation steering, convergence across models, and neural thickets, and connects adaptation through weights, prompts, and hidden representations as different routes to actualizing model capabilities.

\ifarxivstyle %
\section*{Acknowledgements}
We gratefully acknowledge the support of the Swiss National Science Foundation (SNSF) starting grant TMSGI2\_226252/1, SNSF grant IC00I0\_231922, SNSF grant 10.004.411, Swiss Cancer League (KFS-6340-02-2025-R), Kamprad Foundation (Cancer Immunotherapy Networking Program), and the Swiss AI Initiative Large Call (ab006). Maria Brbi\'c is a CIFAR Fellow in the Multiscale Human Program.
\fi

\bibliographystyle{template/iclr2026_conference}
\bibliography{refs}

@book{aristotle,
  title        = {{Metaphysics}},
  author       = {Aristotle},
  year         = {ca. 350 B.C.E.},
  note         = {A revised text with introduction and commentary by W. D. Ross, Clarendon Press, Oxford, 1924}
}

@inproceedings{howard2018universal,
  title        = {Universal language model fine-tuning for text classification},
  author       = {Howard, Jeremy and Ruder, Sebastian},
  year         = 2018,
  booktitle    = {Annual Meeting of the Association for Computational Linguistics (ACL)}
}

@inproceedings{hu2022lora,
  title        = {{Lo{RA}: Low-Rank Adaptation of Large Language Models}},
  author       = {Hu, Edward J. and Shen, Yelong and Wallis, Phillip and Allen-Zhu, Zeyuan and Li, Yuanzhi and Wang, Shean and Wang, Lu and Chen, Weizhu},
  year         = 2022,
  booktitle    = {International Conference on Learning Representations (ICLR)}
}

@inproceedings{hu2023llmadapters,
  title        = {{{LLM-Adapters}: An Adapter Family for Parameter-Efficient Fine-Tuning of Large Language Models}},
  author       = {Hu, Zhiqiang and Wang, Lei and Lan, Yihuai and Xu, Wanyu and Lim, Ee-Peng and Bing, Lidong and Xu, Xing and Poria, Soujanya and Lee, Roy Ka-Wei},
  year         = 2023,
  booktitle    = {Conference on Empirical Methods in Natural Language Processing (EMNLP)}
}

@inproceedings{liu2024dora,
  title        = {{{DoRA}: Weight-Decomposed Low-Rank Adaptation}},
  author       = {Liu, Shih-Yang and Wang, Chien-Yi and Yin, Hongxu and Molchanov, Pavlo and Wang, Yu-Chiang Frank and Cheng, Kwang-Ting and Chen, Min-Hung},
  year         = 2024,
  booktitle    = {International Conference on Machine Learning (ICML)}
}

@inproceedings{dettmers2023qlora,
  title        = {{{QLoRA}: Efficient Finetuning of Quantized {LLMs}}},
  author       = {Dettmers, Tim and Pagnoni, Artidoro and Holtzman, Ari and Zettlemoyer, Luke},
  year         = 2023,
  booktitle    = {Advances in Neural Information Processing Systems (NeurIPS)}
}

@article{biderman2024lora,
  title        = {{{LoRA} Learns Less and Forgets Less}},
  author       = {Biderman, Dan and Portes, Jacob and Ortiz, Jose Javier Gonzalez and Paul, Mansheej and Greengard, Philip and Jennings, Connor and King, Daniel and Havens, Sam and Chiley, Vitaliy and Frankle, Jonathan and Blakeney, Cody and Cunningham, John P.},
  year         = 2024,
  journal      = {Transactions on Machine Learning Research (TMLR)}
}

@inproceedings{zhu2024asymmetry,
  title        = {{Asymmetry in Low-Rank Adapters of Foundation Models}},
  author       = {Zhu, Jiacheng and Greenewald, Kristjan and Nadjahi, Kimia and S{\'a}ez de Oc{\'a}riz Borde, Haitz and Br{\"u}el Gabrielsson, Rickard and Choshen, Leshem and Ghassemi, Marzyeh and Yurochkin, Mikhail and Solomon, Justin},
  year         = 2024,
  booktitle    = {International Conference on Machine Learning (ICML)}
}

@inproceedings{kopiczko2024vera,
  title        = {{Ve{RA}: Vector-based Random Matrix Adaptation}},
  author       = {Kopiczko, Dawid J. and Blankevoort, Tijmen and Asano, Yuki M.},
  year         = 2024,
  booktitle    = {International Conference on Learning Representations (ICLR)}
}

@article{zhang2023lorafa,
  title        = {{Lo{RA}-{FA}: Efficient and Effective Low Rank Representation Fine-tuning}},
  author       = {Zhang, Longteng and Zhang, Lin and Shi, Shaohuai and Chu, Xiaowen and Li, Bo},
  year         = 2023,
  journal      = {arXiv preprint}
}

@article{mandica2026gpart,
  title        = {{{GPart}: End-to-End Isometric Fine-Tuning via Global Parameter Partitioning}},
  author       = {Mandica, Paolo and Brzozowski, Micha{\l} and Dubanowska, Zuzanna and Chung, Neo Christopher},
  year         = 2026,
  journal      = {arXiv preprint}
}

@inproceedings{aghajanyan2021intrinsic,
  title        = {{Intrinsic Dimensionality Explains the Effectiveness of Language Model Fine-Tuning}},
  author       = {Aghajanyan, Armen and Gupta, Sonal and Zettlemoyer, Luke},
  year         = 2021,
  booktitle    = {Annual Meeting of the Association for Computational Linguistics (ACL)}
}

@inproceedings{benzaken2022bitfit,
  title        = {{Bit{F}it: Simple Parameter-efficient Fine-tuning for Transformer-based Masked Language-models}},
  author       = {Ben Zaken, Elad and Goldberg, Yoav and Ravfogel, Shauli},
  year         = 2022,
  booktitle    = {Annual Meeting of the Association for Computational Linguistics (ACL)}
}

@inproceedings{gao2024fourierft,
  title        = {{Parameter-Efficient Fine-Tuning with Discrete Fourier Transform}},
  author       = {Gao, Ziqi and Wang, Qichao and Chen, Aochuan and Liu, Zijing and Wu, Bingzhe and Chen, Liang and Li, Jia},
  year         = 2024,
  booktitle    = {International Conference on Machine Learning (ICML)}
}

@inproceedings{li2025unilora,
  title        = {{{Uni-LoRA}: One Vector is All You Need}},
  author       = {Li, Kaiyang and Han, Shaobo and Su, Qing and Li, Wei and Cai, Zhipeng and Ji, Shihao},
  year         = 2025,
  booktitle    = {Advances in Neural Information Processing Systems (NeurIPS)}
}

@inproceedings{zhao2024galore,
  title        = {{Ga{L}ore: Memory-Efficient {LLM} Training by Gradient Low-Rank Projection}},
  author       = {Zhao, Jiawei and Zhang, Zhenyu and Chen, Beidi and Wang, Zhangyang and Anandkumar, Anima and Tian, Yuandong},
  year         = 2024,
  booktitle    = {International Conference on Machine Learning (ICML)}
}

@inproceedings{groeger2026aristotelian,
  title        = {{Revisiting the Platonic Representation Hypothesis: An Aristotelian View}},
  author       = {Gr{\"o}ger, Fabian and Wen, Shuo and Brbi{\'c}, Maria},
  year         = 2026,
  booktitle    = {International Conference on Machine Learning (ICML)}
}

@inproceedings{huh2024platonic,
  title        = {{Position: The Platonic Representation Hypothesis}},
  author       = {Huh, Minyoung and Cheung, Brian and Wang, Tongzhou and Isola, Phillip},
  year         = 2024,
  booktitle    = {International Conference on Machine Learning (ICML)}
}

@article{kriegeskorte2008rsa,
  title        = {{Representational Similarity Analysis -- Connecting the Branches of Systems Neuroscience}},
  author       = {Kriegeskorte, Nikolaus and Mur, Marieke and Bandettini, Peter A.},
  year         = 2008,
  journal      = {Frontiers in Systems Neuroscience}
}

@inproceedings{kornblith2019cka,
  title        = {{Similarity of Neural Network Representations Revisited}},
  author       = {Kornblith, Simon and Norouzi, Mohammad and Lee, Honglak and Hinton, Geoffrey},
  year         = 2019,
  booktitle    = {International Conference on Machine Learning (ICML)}
}

@article{turner2023actadd,
  title        = {{Steering Language Models With Activation Engineering}},
  author       = {Turner, Alexander Matt and Thiergart, Lisa and Leech, Gavin and Udell, David and Vazquez, Juan J. and Mini, Ulisse and MacDiarmid, Monte},
  year         = 2023,
  journal      = {arXiv preprint}
}

@inproceedings{rimsky2024caa,
  title        = {{Steering Llama 2 via Contrastive Activation Addition}},
  author       = {Rimsky, Nina and Gabrieli, Nick and Schulz, Julian and Tong, Meg and Hubinger, Evan and Turner, Alexander Matt},
  year         = 2024,
  booktitle    = {Annual Meeting of the Association for Computational Linguistics (ACL)}
}

@article{zou2023repe,
  title        = {{Representation Engineering: A Top-Down Approach to AI Transparency}},
  author       = {Zou, Andy and Phan, Long and Chen, Sarah and Campbell, James and Guo, Phillip and Ren, Richard and Pan, Alexander and Yin, Xuwang and Mazeika, Mantas and Dombrowski, Ann-Kathrin and Goel, Shashwat and Li, Nathaniel and Byun, Michael J. and Wang, Zifan and Mallen, Alex and Basart, Steven and Koyejo, Sanmi and Song, Dawn and Fredrikson, Matt and Kolter, J. Zico and Hendrycks, Dan},
  year         = 2023,
  journal      = {arXiv preprint}
}

@inproceedings{todd2024functionvectors,
  title        = {{Function Vectors in Large Language Models}},
  author       = {Todd, Eric and Li, Millicent L. and Sharma, Arnab Sen and Mueller, Aaron and Wallace, Byron C. and Bau, David},
  year         = 2024,
  booktitle    = {International Conference on Learning Representations (ICLR)}
}

@inproceedings{hendel2023taskvectors,
  title        = {{In-Context Learning Creates Task Vectors}},
  author       = {Hendel, Roee and Geva, Mor and Globerson, Amir},
  year         = 2023,
  booktitle    = {Findings of the Conference on Empirical Methods in Natural Language Processing (EMNLP)}
}

@inproceedings{lester2021power,
  title        = {{The Power of Scale for Parameter-Efficient Prompt Tuning}},
  author       = {Lester, Brian and Al-Rfou, Rami and Constant, Noah},
  year         = 2021,
  booktitle    = {Conference on Empirical Methods in Natural Language Processing (EMNLP)}
}

@inproceedings{li2021prefix,
  title        = {{Prefix-Tuning: Optimizing Continuous Prompts for Generation}},
  author       = {Li, Xiang Lisa and Liang, Percy},
  year         = 2021,
  booktitle    = {Annual Meeting of the Association for Computational Linguistics (ACL)}
}

@inproceedings{mu2023gisting,
  title        = {{Learning to Compress Prompts with Gist Tokens}},
  author       = {Mu, Jesse and Li, Xiang Lisa and Goodman, Noah},
  year         = 2023,
  booktitle    = {Advances in Neural Information Processing Systems (NeurIPS)}
}

@misc{conover2023dolly,
  title        = {{Free Dolly: Introducing the World's First Truly Open Instruction-Tuned LLM}},
  author       = {Conover, Mike and Hayes, Matt and Mathur, Ankit and Xie, Jianwei and Wan, Jun and Shah, Sam and Ghodsi, Ali and Wendell, Patrick and Zaharia, Matei and Xin, Reynold},
  year         = 2023,
  note         = {databricks-dolly-15k dataset},
  howpublished = {Databricks blog},
  url          = {https://www.databricks.com/blog/2023/04/12/dolly-first-open-commercially-viable-instruction-tuned-llm}
}

@article{xia2025crosslora,
  title        = {{Cross-{LoRA}: A Data-Free {LoRA} Transfer Framework across Heterogeneous {LLMs}}},
  author       = {Xia, Feifan and Liao, Mingyang and Fang, Yuyang and Li, Defang and Xie, Yantong and Li, Weikang and Li, Yang and Xia, Deguo and Huang, Jizhou},
  year         = 2025,
  journal      = {arXiv preprint}
}

@inproceedings{gan2026thickets,
  title        = {{Neural Thickets: Diverse Task Experts Are Dense Around Pretrained Weights}},
  author       = {Gan, Yulu and Isola, Phillip},
  year         = 2026,
  booktitle    = {International Conference on Machine Learning (ICML)}
}

@article{wang2025oocr,
  title        = {{Simple Mechanistic Explanations for Out-Of-Context Reasoning}},
  author       = {Wang, Atticus and Engels, Joshua and Clive-Griffin, Oliver and Rajamanoharan, Senthooran and Nanda, Neel},
  year         = 2025,
  journal      = {arXiv preprint}
}

@inproceedings{sinii2025biasonly,
  title        = {{Steering {LLM} Reasoning Through Bias-Only Adaptation}},
  author       = {Sinii, Viacheslav and Gorbatovski, Alexey and Cherepanov, Artem and Shaposhnikov, Boris and Balagansky, Nikita and Gavrilov, Daniil},
  year         = 2025,
  booktitle    = {Conference on Empirical Methods in Natural Language Processing (EMNLP)}
}

@inproceedings{panigrahi2023skill,
  title        = {{Task-Specific Skill Localization in Fine-tuned Language Models}},
  author       = {Panigrahi, Abhishek and Saunshi, Nikunj and Zhao, Haoyu and Arora, Sanjeev},
  year         = 2023,
  booktitle    = {International Conference on Machine Learning (ICML)}
}

@inproceedings{adila2026activation,
  title        = {{Weight Updates as Activation Shifts: A Principled Framework for Steering}},
  author       = {Adila, Dyah and Cooper, John and Yun, Alexander and Trost, Avi and Sala, Frederic},
  year         = 2026,
  booktitle    = {International Conference on Machine Learning (ICML)}
}

@inproceedings{wu2024reft,
  title        = {{{ReFT}: Representation Finetuning for Language Models}},
  author       = {Wu, Zhengxuan and Arora, Aryaman and Wang, Zheng and Geiger, Atticus and Jurafsky, Dan and Manning, Christopher D. and Potts, Christopher},
  year         = 2024,
  booktitle    = {Advances in Neural Information Processing Systems (NeurIPS)}
}

@inproceedings{yin2024lofit,
  title        = {{{LoFiT}: Localized Fine-tuning on {LLM} Representations}},
  author       = {Yin, Fangcong and Ye, Xi and Durrett, Greg},
  year         = 2024,
  booktitle    = {Advances in Neural Information Processing Systems (NeurIPS)}
}

@inproceedings{brown2020language,
  title        = {{Language Models are Few-Shot Learners}},
  author       = {Brown, Tom B. and Mann, Benjamin and Ryder, Nick and Subbiah, Melanie and Kaplan, Jared and Dhariwal, Prafulla and Neelakantan, Arvind and Shyam, Pranav and Sastry, Girish and Askell, Amanda and others},
  year         = 2020,
  booktitle    = {Advances in Neural Information Processing Systems (NeurIPS)}
}

@inproceedings{devlin2019bert,
  title        = {{BERT: Pre-training of Deep Bidirectional Transformers for Language Understanding}},
  author       = {Devlin, Jacob and Chang, Ming-Wei and Lee, Kenton and Toutanova, Kristina},
  year         = 2019,
  booktitle    = {Conference of the North American Chapter of the Association for Computational Linguistics (NAACL)}
}

@article{bommasani2021foundation,
  title        = {{On the Opportunities and Risks of Foundation Models}},
  author       = {Bommasani, Rishi and Hudson, Drew A. and Adeli, Ehsan and Altman, Russ and Arora, Simran and von Arx, Sydney and Bernstein, Michael S. and Bohg, Jeannette and Bosselut, Antoine and Brunskill, Emma and others},
  year         = 2021,
  journal      = {arXiv preprint}
}

@inproceedings{wu2025axbench,
  title        = {{{AxBench}: Steering {LLMs}? Even Simple Baselines Outperform Sparse Autoencoders}},
  author       = {Wu, Zhengxuan and Arora, Aryaman and Geiger, Atticus and Wang, Zheng and Huang, Jing and Jurafsky, Dan and Manning, Christopher D. and Potts, Christopher},
  year         = 2025,
  booktitle    = {International Conference on Machine Learning (ICML)}
}

@inproceedings{liu2024infinigram,
  title        = {Infini-gram: Scaling Unbounded n-gram Language Models to a Trillion Tokens},
  author       = {Liu, Jiacheng and Min, Sewon and Zettlemoyer, Luke and Choi, Yejin and Hajishirzi, Hannaneh},
  year         = 2024,
  booktitle    = {Conference on Language Modeling (COLM)}
}

@inproceedings{soldaini2024dolma,
  title        = {Dolma: an Open Corpus of Three Trillion Tokens for Language Model Pretraining Research},
  author       = {Soldaini, Luca and Kinney, Rodney and Bhagia, Akshita and Schwenk, Dustin and Atkinson, David and Authur, Russell and others},
  year         = 2024,
  booktitle    = {Annual Meeting of the Association for Computational Linguistics (ACL)}
}

@article{zhou2023ifeval,
  title        = {Instruction-Following Evaluation for Large Language Models},
  author       = {Zhou, Jeffrey and Lu, Tianjian and Mishra, Swaroop and Brahma, Siddhartha and Basu, Sujoy and Luan, Yi and Zhou, Denny and Hou, Le},
  year         = 2023,
  journal      = {arXiv preprint}
}

@inproceedings{zheng2021casehold,
  title        = {When Does Pretraining Help? {A}ssessing Self-Supervised Learning for Law and the {CaseHOLD} Dataset},
  author       = {Zheng, Lucia and Guha, Neel and Anderson, Brandon R. and Henderson, Peter and Ho, Daniel E.},
  year         = 2021,
  booktitle    = {International Conference on Artificial Intelligence and Law (ICAIL)}
}

@inproceedings{pal2022medmcqa,
  title        = {{MedMCQA}: A Large-scale Multi-Subject Multi-Choice Dataset for Medical domain Question Answering},
  author       = {Pal, Ankit and Umapathi, Logesh Kumar and Sankarasubbu, Malaikannan},
  year         = 2022,
  booktitle    = {Conference on Health, Inference, and Learning (CHIL)}
}

@inproceedings{ling2017aqua,
  title        = {Program Induction by Rationale Generation: Learning to Solve and Explain Algebraic Word Problems},
  author       = {Ling, Wang and Yogatama, Dani and Dyer, Chris and Blunsom, Phil},
  year         = 2017,
  booktitle    = {Annual Meeting of the Association for Computational Linguistics (ACL)}
}

@inproceedings{welbl2017sciq,
  title        = {Crowdsourcing Multiple Choice Science Questions},
  author       = {Welbl, Johannes and Liu, Nelson F. and Gardner, Matt},
  year         = 2017,
  booktitle    = {EMNLP Workshop on Noisy User-generated Text (W-NUT)}
}

@inproceedings{lambert2024tulu,
  title        = {{T{\"u}lu} 3: Pushing Frontiers in Open Language Model Post-Training},
  author       = {Lambert, Nathan and Morrison, Jacob and Pyatkin, Valentina and Huang, Shengyi and Ivison, Hamish and Brahman, Faeze and Miranda, Lester James V. and Liu, Alisa and Dziri, Nouha and Lyu, Shane and others},
  year         = 2025,
  booktitle    = {Conference on Language Modeling (COLM)}
}

@inproceedings{olmo2024furious,
  title        = {2 {OLMo} 2 Furious},
  author       = {{Team OLMo} and Walsh, Pete and Soldaini, Luca and Groeneveld, Dirk and others},
  year         = 2025,
  booktitle    = {Conference on Language Modeling (COLM)}
}

@article{olsson2022induction,
  title        = {{In-context Learning and Induction Heads}},
  author       = {Olsson, Catherine and Elhage, Nelson and Nanda, Neel and Joseph, Nicholas and DasSarma, Nova and Henighan, Tom and Mann, Ben and Askell, Amanda and Bai, Yuntao and Chen, Anna and others},
  year         = 2022,
  journal      = {arXiv preprint}
}

@inproceedings{socher2013sst,
  title        = {Recursive Deep Models for Semantic Compositionality Over a Sentiment Treebank},
  author       = {Socher, Richard and Perelygin, Alex and Wu, Jean and Chuang, Jason and Manning, Christopher D. and Ng, Andrew and Potts, Christopher},
  year         = 2013,
  booktitle    = {Conference on Empirical Methods in Natural Language Processing (EMNLP)}
}

@inproceedings{zhang2015charcnn,
  title        = {Character-level Convolutional Networks for Text Classification},
  author       = {Zhang, Xiang and Zhao, Junbo and LeCun, Yann},
  year         = 2015,
  booktitle    = {Advances in Neural Information Processing Systems (NeurIPS)}
}

@inproceedings{arora2024zoology,
  title        = {Zoology: Measuring and Improving Recall in Efficient Language Models},
  author       = {Arora, Simran and Eyuboglu, Sabri and Timalsina, Aman and Johnson, Isys and Poli, Michael and Zou, James and Rudra, Atri and R{\'e}, Christopher},
  year         = 2024,
  booktitle    = {International Conference on Learning Representations (ICLR)}
}

@inproceedings{wang2024mmlupro,
  title        = {{MMLU-Pro}: A More Robust and Challenging Multi-Task Language Understanding Benchmark},
  author       = {Wang, Yubo and Ma, Xueguang and Zhang, Ge and Ni, Yuansheng and Chandra, Abhranil and Guo, Shiguang and Ren, Weiming and Arulraj, Aaran and He, Xuan and Jiang, Ziyan and Li, Tianle and Ku, Max and Wang, Kai and Zhuang, Alex and Fan, Rongqi and Yue, Xiang and Chen, Wenhu},
  year         = 2024,
  booktitle    = {Advances in Neural Information Processing Systems (NeurIPS)}
}

@inproceedings{map2025supergpqa,
  title        = {{SuperGPQA}: Scaling {LLM} Evaluation across 285 Graduate Disciplines},
  author       = {Du, Xeron and Yao, Yifan and Ma, Kaijing and Wang, Bingli and Zheng, Tianyu and others},
  year         = 2025,
  booktitle    = {Advances in Neural Information Processing Systems (NeurIPS)}
}

@inproceedings{chen2025persona,
  title        = {Persona Vectors: Monitoring and Controlling Character Traits in Language Models},
  author       = {Chen, Runjin and Arditi, Andy and Sleight, Henry and Evans, Owain and Lindsey, Jack},
  year         = 2026,
  booktitle    = {Advances in Neural Information Processing Systems (NeurIPS)}
}

@inproceedings{subramani-etal-2022-extracting,
  title        = {Extracting Latent Steering Vectors from Pretrained Language Models},
  author       = {Subramani, Nishant and Suresh, Nivedita and Peters, Matthew},
  year         = 2022,
  booktitle    = {Findings of the Annual Meeting of the Association for Computational Linguistics (ACL)}
}

@article{cobbe2021training,
  title        = {Training verifiers to solve math word problems},
  author       = {Cobbe, Karl and Kosaraju, Vineet and Bavarian, Mohammad and Chen, Mark and Jun, Heewoo and Kaiser, Lukasz and Plappert, Matthias and Tworek, Jerry and Hilton, Jacob and Nakano, Reiichiro and Hesse, Christopher and Schulman, John},
  year         = 2021,
  journal      = {arXiv preprint}
}

@article{beaglehole2026universal,
  title        = {Toward universal steering and monitoring of {AI} models},
  author       = {Beaglehole, Daniel and Radhakrishnan, Adityanarayanan and Boix-Adser{\`a}, Enric and Belkin, Mikhail},
  year         = 2026,
  journal      = {Science}
}

@inproceedings{vaswani2017attention,
  title     = {Attention Is All You Need},
  author    = {Vaswani, Ashish and Shazeer, Noam and Parmar, Niki and Uszkoreit, Jakob and Jones, Llion and Gomez, Aidan N. and Kaiser, {\L}ukasz and Polosukhin, Illia},
  booktitle = {Advances in Neural Information Processing Systems (NeurIPS)},
  year      = {2017}
}

@inproceedings{he2016deep,
  title     = {Deep Residual Learning for Image Recognition},
  author    = {He, Kaiming and Zhang, Xiangyu and Ren, Shaoqing and Sun, Jian},
  booktitle = {IEEE Conference on Computer Vision and Pattern Recognition (CVPR)},
  year      = {2016}
}

@inproceedings{zhang2019root,
  title     = {Root Mean Square Layer Normalization},
  author    = {Zhang, Biao and Sennrich, Rico},
  booktitle = {Advances in Neural Information Processing Systems (NeurIPS)},
  year      = {2019}
}

@inproceedings{xiong2020layer,
  title     = {On Layer Normalization in the Transformer Architecture},
  author    = {Xiong, Ruibin and Yang, Yunchang and He, Di and Zheng, Kai and Zheng, Shuxin and Xing, Chen and Zhang, Huishuai and Lan, Yanyan and Wang, Liwei and Liu, Tie-Yan},
  booktitle = {International Conference on Machine Learning (ICML)},
  year      = {2020}
}

@article{su2024roformer,
  title   = {{RoFormer}: Enhanced Transformer with Rotary Position Embedding},
  author  = {Su, Jianlin and Ahmed, Murtadha and Lu, Yu and Pan, Shengfeng and Bo, Wen and Liu, Yunfeng},
  journal = {Neurocomputing},
  year    = {2024}
}

@article{shazeer2020glu,
  title   = {{GLU} Variants Improve Transformer},
  author  = {Shazeer, Noam},
  journal = {arXiv preprint},
  year    = {2020}
}

\clearpage

\newpage
\appendix
\raggedbottom %
\makeatletter\setlength{\@fptop}{0pt}\makeatother %
\crefalias{section}{appendix}
\crefalias{subsection}{appendix}
\crefalias{subsubsection}{appendix}
\addtocontents{toc}{\protect\setcounter{tocdepth}{2}}
\renewcommand{\contentsname}{Appendix contents}
\tableofcontents

\newpage
\section{\update{Extended related work}}\label{app:related}

Adaptation strategies differ in what they change: parameter-efficient fine-tuning learns a small update to the weights, in-context learning conditions the model on demonstrations in the prompt, activation steering modifies the hidden activations at inference time, and representation fine-tuning learns a change to the representations themselves. The field has treated these as different tools. The potentiality and actuality principle reads each of them as reaching an actualizing state.

\textbf{Parameter-efficient fine-tuning.} Parameter-efficient fine-tuning freezes the model and trains a small update to its weights. The update can be a low-rank matrix on selected weights \citep{hu2022lora}, the existing biases \update{and a new task layer} \citep{benzaken2022bitfit}, per-feature scales over frozen random factors \citep{kopiczko2024vera}, one LoRA factor with the other left random \citep{zhu2024asymmetry,zhang2023lorafa}, or an update confined to a random, Fourier, shared or gradient subspace \citep{gao2024fourierft,li2025unilora,zhao2024galore,mandica2026gpart,xia2025crosslora}. Low intrinsic dimension is the broadest form of the observation that so little suffices \citep{aghajanyan2021intrinsic}. Grafting the fine-tuned values of $0.01\%$ of the weights, chosen by a learned sparse mask, onto the base model keeps over $95\%$ of the fine-tuned accuracy, and in a model tuned on several tasks at once the tasks use nearly disjoint sets of weights \citep{panigrahi2023skill}. That work finds which weights carry the gain and in which layers and modules they sit. Each such weight still multiplies the input, so its effect on the hidden states could differ from input to input. For LoRA and full fine-tuning we find the opposite: most of the gain lies in a component shared by all inputs (\cref{sec:field}). We show why it suffices: the update selects a state the base model holds in potentiality, and the shift adapter, one vector per layer, is the smallest member of this family that does so (\cref{sec:parameterizations}).

\textbf{In-context learning.} In-context learning adapts a model through demonstrations in the prompt, with no change to the weights \citep{brown2020language}, carried by attention heads that copy from the context \citep{olsson2022induction}. Mechanism work compresses the demonstrations into a single vector: a task vector read from the hidden state at the end of the prompt \citep{hendel2023taskvectors}, or a function vector read from a small set of attention heads, which reproduces the task when added to the forward pass \citep{todd2024functionvectors}. The prompt itself can also be learned: prompt tuning trains the embeddings of a few tokens placed before the input \citep{lester2021power}, prefix tuning trains such tokens at every layer \citep{li2021prefix}, and gisting trains the model to compress a long prompt into a few tokens that stand in for it \citep{mu2023gisting}. We place in-context learning beside adaptation on the same behaviors and find that it reaches them by a different route: its shared component keeps less of the gain than LoRA's on the traits, the transformations and the tasks, and it points nearly orthogonally to LoRA's shared component (\cref{sec:field}). The task vector is then one of the three extracted vectors we compare with the trained shift adapter in \cref{sec:impact}.

\textbf{Activation steering.} Activation steering adds a vector to the hidden states at inference time and leaves the weights untouched. The vector comes from the model's own activations: Activation Addition takes the difference between the activations of two prompts \citep{turner2023actadd}, Contrastive Activation Addition averages that difference over many prompt pairs with and without the behavior \citep{rimsky2024caa}, Representation Engineering reads the leading direction of such differences \citep{zou2023repe}, and persona vectors apply the same construction to character traits \citep{chen2025persona}. AxBench benchmarks these methods against fine-tuning on one fixed model, Gemma-2-2B, with its own judge \citep{wu2025axbench}. The trained shift adapter adds its vector at the same place, one per layer, but learns it by adaptation, and \cref{sec:impact} shows that training reaches behaviors that extraction misses.

\textbf{Representation fine-tuning.} Representation fine-tuning trains a change to the hidden representations and leaves the weights frozen. ReFT learns a low-rank transformation of the representations at chosen positions, which in general depends on the input \citep{wu2024reft}. LoFiT learns fixed offsets at a few selected attention heads \citep{yin2024lofit}. \update{Bias training trains the existing biases together with a new classification layer for each task, through which every answer leaves the model \citep{benzaken2022bitfit}}, and \citet{sinii2025biasonly} train a bias vector per layer with reinforcement learning to match fully tuned reasoning models on mathematical benchmarks, a bound on how few parameters reasoning needs. \update{They see the input-independence of this vector as a limitation that may cap performance, while our paper shows that most of what LoRA adds is the same for every input (\cref{sec:field}) and explains why such a vector suffices (\cref{app:existence,app:conditions}). The new layer of \citet{benzaken2022bitfit} adds an input-specific component trained on the task, so accuracy there cannot show that the expertise was already in the base model.} \update{The vector of a shift adapter} is the shared component of what any adaptation learns, on which a LoRA relies for most of its gain (\cref{sec:field}).

\textbf{From weight updates to actualizing states.}
\citet{wang2025oocr} study single-layer LoRAs on out-of-context reasoning tasks and find that the update's added vectors point in nearly the same direction across tokens, and that extracted or directly trained steering vectors can induce the learned behavior. \citet{adila2026activation} derive a local first-order relation between weight updates and activation shifts and use it to build post-block activation adapters. Both connect changes in weights to states in activation space. Our paired intervention turns this correspondence into a test of the trained update itself: keeping only the shared component and removing it establish when a flexible multi-layer LoRA relies on an actualizing state (\cref{sec:field}).

\update{\Cref{tab:positioning} summarizes what each of these works shows and the questions about adaptation that our paper adds.}

\begin{rtable}[h]
\centering
\caption{\update{\textbf{What related work on small updates shows, and the questions about adaptation that our paper adds.} For each work, its main finding, and whether it explains why a small update suffices, when the capacity forms during pretraining, how the states that actualize a behavior are structured, whether it predicts from the base model when a constant suffices, and which fields it connects. A dash marks a question the work does not address.}}
\label{tab:positioning}
\updatebox{\resizebox{\dimexpr\textwidth-8.4pt\relax}{!}{\scriptsize
\setlength{\tabcolsep}{2.5pt}\renewcommand{\arraystretch}{1.15}
\begin{tabular}{>{\raggedright\arraybackslash}p{4.4cm}>{\raggedright\arraybackslash}p{1.9cm}>{\raggedright\arraybackslash}p{1.9cm}>{\raggedright\arraybackslash}p{1.5cm}>{\raggedright\arraybackslash}p{1.6cm}>{\raggedright\arraybackslash}p{2.0cm}}
\toprule
Work and main finding & Why a small update suffices & When it forms & Structure & Prediction before adaptation & Fields connected \\
\midrule
\citet{benzaken2022bitfit}: training the biases and a new classification layer is competitive with fine-tuning on classification & hypothesis on linguistic knowledge & -- & -- & -- & -- \\
\citet{yin2024lofit}: offsets at a few selected attention heads match parameter-efficient fine-tuning & -- & -- & -- & -- & -- \\
\citet{wu2024reft}: low-rank edits of hidden representations match weight-based fine-tuning & -- & -- & -- & -- & -- \\
\citet{sinii2025biasonly}: one vector per layer, trained with reinforcement learning, matches tuned reasoning models & -- & -- & -- & -- & -- \\
\citet{wang2025oocr}: on Gemma-3-12B, a LoRA on one layer trained for out-of-context reasoning acts like a steering vector & -- & -- & -- & -- & -- \\
\midrule
\textbf{Ours}: adaptation actualizes a capacity that the pretrained model already holds & shared component of LoRA, full fine-tuning and in-context learning, 18 models, 26 benchmarks & OLMo-2-7B (15 checkpoints), OLMo-2-1B, Pythia-12B & connected region, its width, edge and size, and its growth over training & readout of the base model, 24 pairs of rule and model & fine-tuning, in-context learning, steering, convergence, neural thickets \\
\bottomrule
\end{tabular}}}
\end{rtable}

\section{Proofs}\label{app:proofs}

\subsection{The shared and input-specific decomposition}\label{app:pf-decomp}

\begin{proposition}[A linear update adds a shared component plus a low-rank input-specific component]
\label{prop:decomp}
Let $z$ be a random vector with mean $\mu$ and covariance $\Sigma$, and let the adapter add $a(z)=\dW z$ with $\operatorname{rank}(\dW)\le r$. Then
\begin{equation*}
  a(z)=\bar a+\widetilde a(z),\qquad
  \bar a=\dW\mu,\qquad
  \widetilde a(z)=\dW(z-\mu),
\end{equation*}
is the unique decomposition into a constant term and a zero-mean term. Moreover,
\begin{equation*}
  \operatorname{Cov}\!\left(\widetilde a(z)\right)=\dW\Sigma\dW^\top,
  \qquad
  \operatorname{rank}\!\left(\dW\Sigma\dW^\top\right)\le r,
\end{equation*}
and the energy identity of \cref{eq:energy} holds.
\end{proposition}

\begin{proof}
Linearity gives $\dW z=\dW\mu+\dW(z-\mu)$, and
$\E[\dW(z-\mu)]=\dW\E[z-\mu]=0$. If $\dW z=c+g(z)$ with $c$ constant and
$\E[g(z)]=0$, taking expectations forces $c=\dW\mu$, which proves uniqueness. The covariance of the input-specific component
is
\[
  \E\!\left[\dW(z-\mu)(z-\mu)^\top\dW^\top\right]
  =\dW\Sigma\dW^\top,
\]
whose rank is at most $\operatorname{rank}(\dW)\le r$. Expanding the squared norm gives
\[
\E\lVert\dW z\rVert^2
=\lVert\dW\mu\rVert^2
+2(\dW\mu)^\top\E[\dW(z-\mu)]
+\E\lVert\dW(z-\mu)\rVert^2.
\]
The middle term vanishes, and the last term equals
$\tr(\dW\Sigma\dW^\top)$.
\end{proof}

\begin{remark}[The decomposition is the intervention]\label{rem:decomp}
Keeping only $\bar a$ and keeping only $\widetilde a(z)$ are the two interventions of \cref{sec:field}. The proposition gives an exact split into the two components, and the interventions measure which one the adapted behavior relies on.
\end{remark}

\subsection{Why a constant parameter has an input-dependent effect}
\label{app:pf-fixed-state-effect}

\begin{lemma}[The frozen layers supply the input dependence]\label{lem:fixed-state-effect}
Let $G_\ell:\R^d\to\R^m$ be differentiable and let $b_\ell\in\R^d$ be fixed. For every input $x$,
the local effect of adding $b_\ell$ to $h_\ell(x)$ is
\begin{equation}\label{eq:fixed-state-jacobian}
\frac{d}{d\alpha}G_\ell\!\left(h_\ell(x)+\alpha b_\ell\right)\bigg|_{\alpha=0}
=J_{G_\ell}\!\left(h_\ell(x)\right)b_\ell.
\end{equation}
Two inputs $x_1,x_2$ therefore respond differently to first order whenever
$\bigl(J_{G_\ell}(h_\ell(x_1))-J_{G_\ell}(h_\ell(x_2))\bigr)b_\ell\ne0$.
\end{lemma}

\begin{proof}
Apply the chain rule to the path $g_x(\alpha)=G_\ell(h_\ell(x)+\alpha b_\ell)$:
$g_x'(0)=J_{G_\ell}(h_\ell(x))b_\ell$. Subtracting this identity at $x_1$ and $x_2$ gives the second statement.
\end{proof}

\subsection{\update{Bias training is not a shift adapter}}\label{app:pf-bias}

\update{Prior work on bias training adapts a model by training its existing bias terms \citep{benzaken2022bitfit}. The shift adapter instead adds one new vector at the output of each block, the same for every input and every position. Both train few parameters and reach the accuracy of larger updates, so to position the two works we ask whether they are the same parameterization. They are not. The biases of our models act inside attention, before the softmax and the MLP of the same block, so a change of them changes the output of the block differently for each input. The shift adapter adds the same vector to the output of the block for every input. Below we show which biases our models have and what a change of each bias can affect, and then formally show that no change of a block's attention parameters, its biases included, reproduces a shift adapter.}

\update{\textbf{Bias terms of our models.} Thirteen of our 18 models have no bias terms at all (\cref{tab:rebuttal-bias-census}), so bias training has no parameters to train on them. The other five are the Qwen2.5 models, which have biases only on the query, key, and value projections of attention. No model has a bias at the output of a block or at the output of its MLP, which is where the shift adapter is placed. On most of the models in this paper, bias training thus does not exist, while the shift adapter applies to every model, and where both exist, they differ in placement.}

\begin{rtable}[tb]
\centering
\caption{\update{\textbf{Bias terms of our models.} For each of the 18 models, the bias terms inside each block, the number of parameters bias training would train, and the number of parameters of the shift adapter.}}
\label{tab:rebuttal-bias-census}
\updatebox{\small
\begin{tabular}{lccc}
\toprule
Model & Biases in each block & Bias training & Shift adapter \\
\midrule
Qwen2.5-0.5B-Instruct & query, key, value & 27{,}648 & 21{,}504 \\
Qwen2.5-1.5B-Instruct & query, key, value & 57{,}344 & 43{,}008 \\
Qwen2.5-3B-Instruct & query, key, value & 92{,}160 & 73{,}728 \\
Qwen2.5-7B-Instruct & query, key, value & 129{,}024 & 100{,}352 \\
Qwen3-4B-Instruct-2507 & none & -- & 92{,}160 \\
Qwen3.5-4B & none & -- & 81{,}920 \\
Qwen3.5-9B & none & -- & 131{,}072 \\
Qwen3.6-27B & none & -- & 327{,}680 \\
Llama-3.2-1B-Instruct & none & -- & 32{,}768 \\
Llama-3.2-3B-Instruct & none & -- & 86{,}016 \\
Llama-3.1-8B-Instruct & none & -- & 131{,}072 \\
SmolLM2-1.7B-Instruct & none & -- & 49{,}152 \\
Mistral-7B-Instruct-v0.3 & none & -- & 131{,}072 \\
gemma-2-2b-it & none & -- & 59{,}904 \\
Llama-3.1-8B & none & -- & 131{,}072 \\
Qwen2.5-7B & query, key, value & 129{,}024 & 100{,}352 \\
gemma-2-2b & none & -- & 59{,}904 \\
Llama-3.1-Tulu-3-8B-SFT & none & -- & 131{,}072 \\
\bottomrule
\end{tabular}
}
\end{rtable}

\update{\textbf{What a change of each bias can affect.} Let $u_1,\dots,u_T$ be the normalized hidden states that enter attention at the $T$ positions of an input. For one head, attention computes at every position $j$ a query, a key, and a value,
\[
q_j=W_Q u_j+b_Q,\qquad k_j=W_K u_j+b_K,\qquad v_j=W_V u_j+b_V,
\]
and its output at position $i$ is a weighted average of the values,
\[
a_i=W_O\sum_{j\le i}\alpha_{ij}\,v_j,\qquad \alpha_{ij}=\operatorname{softmax}_j\!\Big(\tfrac{1}{\sqrt d}\,q_i^\top k_j\Big),
\]
where the weights $\alpha_{ij}$ of each position sum to one. Bias training changes $b_Q$, $b_K$, and $b_V$. We change one of them by a vector $\delta$, keep all other parameters fixed, and follow the change through the block (\cref{fig:rebuttal-bias-block}a).}

\update{\begin{itemize}[leftmargin=1.5em,itemsep=2pt,topsep=2pt]
\item \textit{Query bias, $b_Q\to b_Q+\delta$.} Every query becomes $q_i+\delta$, so the score between positions $i$ and $j$ changes by $(R_i\delta)^\top R_jk_j/\sqrt d$, with the rotary embeddings described for the key bias below. The key $k_j$ depends on the input at position $j$, so the attention weights $\alpha_{ij}$ change, and they change differently for different inputs.
\item \textit{Key bias, $b_K\to b_K+\delta$.} Every key becomes $k_j+\delta$, so the score changes by $q_i^\top\delta/\sqrt d$. Without rotary embeddings this is the same for every $j$, and the softmax removes it. Our models use rotary embeddings \citep{su2024roformer}, which rotate the query and the key by angles that depend on their positions, $q_i\mapsto R_iq_i$ and $k_j\mapsto R_jk_j$. The change of the score is then $(R_iq_i)^\top R_j\delta/\sqrt d$, which differs between positions $j$ and depends on the input through $q_i$.
\item \textit{Value bias, $b_V\to b_V+\delta$.} Every value becomes $v_j+\delta$. Since the weights $\alpha_{ij}$ sum to one, the output of attention changes by $W_O\delta$ at every position, the same vector for every input. This vector, however, is added to the hidden state before the MLP of the same block. The MLP then receives $h'_i+W_O\delta$ instead of the hidden state $h'_i$ after attention, and how much its output changes depends on $h'_i$, and so on the input.
\end{itemize}}
\update{In all three cases, the output of the block changes by an amount that depends on the input. The shift adapter instead adds its vector $b$ after the MLP, to the output $h''_i$ of the block, where nothing in the block acts on it. The block output then changes by the same vector $b$ for every input (\cref{fig:rebuttal-bias-block}b).}

\begin{rfigure}[tb]
\centering
\updatebox{\includegraphics[width=\dimexpr\textwidth-8.4pt\relax]{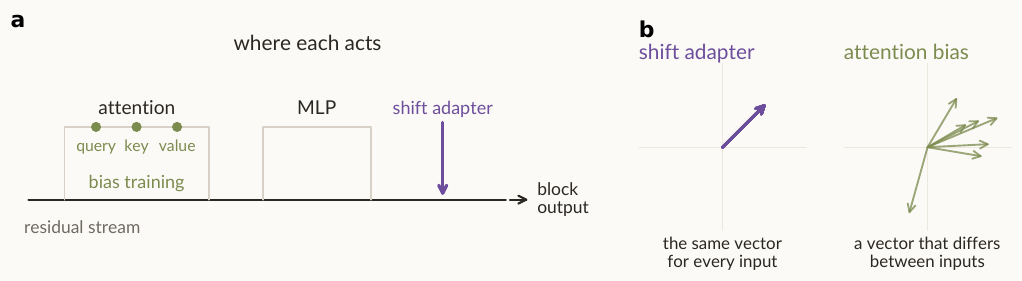}}
\caption{\update{\textbf{Where bias training and the shift adapter change a block.} \textbf{(a)}~One pre-norm block, read from left to right along the residual stream. Bias training changes the query, key, and value biases inside attention, before the softmax and before the MLP. The shift adapter adds its vector at the output of the block, after the MLP. \textbf{(b)}~A schematic of the change at the block output for several inputs: the shift adapter adds the same vector for every input, while a change of an attention bias adds a vector that differs between inputs.}}
\label{fig:rebuttal-bias-block}
\end{rfigure}

\update{\textbf{Notation.} We now show formally that this holds for every change of attention, not only for a change of its biases. We treat one pre-norm block \citep{xiong2020layer} as a function of its input $H=(h_1,\dots,h_T)$, the hidden states at the $T$ positions, each of dimension $d\ge2$. We consider every input whose hidden states are nonzero, not only the inputs that a prompt produces, because the proof scales inputs up. Let}

\update{\begin{itemize}[leftmargin=1.5em,itemsep=1pt,topsep=2pt]
\item $\mathrm{Norm}$ be the RMSNorm \citep{zhang2019root}, applied to each hidden state. It divides a vector by its root mean square, so scaling a vector does not change its normalized form: $\mathrm{Norm}(\lambda h)=\mathrm{Norm}(h)$ for every $\lambda>0$.
\item $U=\mathrm{Norm}(H)$ be the normalized input, and $\mathrm{Attn}_t(U)$ the output of attention at position $t$. Attention, including its rotary embeddings and its causal mask, sees the input only through $U$.
\item $\mathrm{MLP}$ be the MLP of the block. Our models use gated MLPs with the smooth activations SiLU or GELU \citep{shazeer2020glu}, so the MLP is differentiable.
\end{itemize}}
\update{We write $g(y)=\mathrm{MLP}(\mathrm{Norm}(y))$ for the output of the MLP on a hidden state $y\neq0$. It is differentiable, and like $\mathrm{Norm}$ it does not change when $y$ is scaled: $g(\lambda y)=g(y)$ for every $\lambda>0$. We assume that $g$ is not constant, since an MLP that gives the same output for every input would play no role.}

\update{\textbf{The block and its field.} Like every transformer block \citep{vaswani2017attention}, the block adds the output of attention and the output of the MLP to its input, through the residual connection \citep{he2016deep}:
\[
h'_t=h_t+\mathrm{Attn}_t(U),\qquad h''_t=h'_t+\mathrm{MLP}\big(\mathrm{Norm}(h'_t)\big).
\]
A change of the parameters of attention, whether of its biases, its weights, or a LoRA on them, turns $\mathrm{Attn}$ into $\mathrm{Attn}+\Delta\mathrm{Attn}$. We now follow this change through the block in three steps.}

\update{\begin{enumerate}[leftmargin=1.8em,itemsep=2pt,topsep=2pt]
\item After attention, the hidden state becomes $h'_t+\Delta\mathrm{Attn}_t(U)$.
\item The MLP receives this changed hidden state, so the block output becomes
\[
\tilde h''_t=h'_t+\Delta\mathrm{Attn}_t(U)+\mathrm{MLP}\big(\mathrm{Norm}(h'_t+\Delta\mathrm{Attn}_t(U))\big).
\]
\item The change of the block output, the field at position $t$, is the difference to the unchanged output $h''_t$:
\[
F_t(H)=\tilde h''_t-h''_t=\underbrace{\Delta\mathrm{Attn}_t(U)}_{\text{change of attention}}+\underbrace{\mathrm{MLP}\big(\mathrm{Norm}(h'_t+\Delta\mathrm{Attn}_t(U))\big)-\mathrm{MLP}\big(\mathrm{Norm}(h'_t)\big)}_{\text{change it causes in the MLP}}.
\]
\end{enumerate}
A shift adapter with vector $b$ at this block turns $h''_t$ into $h''_t+b$, so its field is $F_t(H)=b$ at every position and for every input. A change of attention reproduces a shift adapter exactly when its field is such a constant vector. The proposition shows that this never happens.}

\update{\begin{proposition}[No change of attention acts as a shift adapter]\label{prop:bias-shift}
In the setting above, suppose that a change of the attention parameters of a block gives the same field $F_t(H)=b$ at every position $t$ and for every input $H$. Then $b=0$, and the change of attention has no effect. Hence no change of the attention parameters of a block, its biases included, acts as a shift adapter at the output of that block.
\end{proposition}}

\update{\textbf{Proof idea.} Scaling the input up leaves attention unchanged and makes the two MLP terms cancel, so attention itself must add $b$. The MLP then has to give the same output whether or not $b$ is added before it. For a nonzero $b$, this means that the output of the MLP does not change when its input moves in the direction of $b$. Since scaling the input does not change it either, the MLP would give the same output everywhere, which it does not.}

\update{\begin{proof}
Suppose that $F_t(H)=b$ for every input $H$ and every position $t$. Fix $H$ and $t$, and write $a=\mathrm{Attn}_t(U)$ and $\Delta=\Delta\mathrm{Attn}_t(U)$. With $g$, the field reads
\begin{equation*}
F_t(H)=\Delta+g(h_t+a+\Delta)-g(h_t+a)=b. \tag{1}
\end{equation*}

\textit{Step 1: attention itself adds $b$.} Replace $H$ by $\lambda H$ with $\lambda>0$. Its hidden states are still nonzero, so (1) holds for $\lambda H$. The normalized input $U$ does not change, so $a$ and $\Delta$ do not change either, and $h_t$ becomes $\lambda h_t$. Since $g$ does not change when its argument is scaled, we can divide both arguments by $\lambda$, and (1) for $\lambda H$ reads
\begin{equation*}
\Delta+g\big(h_t+(a+\Delta)/\lambda\big)-g\big(h_t+a/\lambda\big)=b\qquad\text{for every }\lambda>0. \tag{2}
\end{equation*}
As $\lambda$ grows, both arguments of $g$ tend to $h_t$, which is nonzero, and $g$ is continuous there. The two $g$ terms therefore cancel in the limit, and $\Delta=b$. Since $H$ and $t$ were arbitrary, the change of attention adds the same vector $b$ for every normalized input and at every position.

\textit{Step 2: the MLP does not change along $b$.} Insert $\Delta=b$ into (2), cancel $b$ on both sides, and write $\varepsilon=1/\lambda$:
\begin{equation*}
g\big(h_t+\varepsilon(a+b)\big)=g\big(h_t+\varepsilon a\big)\qquad\text{for every }\varepsilon>0. \tag{3}
\end{equation*}
Subtract $g(h_t)$ on both sides and divide by $\varepsilon$. As $\varepsilon\to0$, the left side tends to the derivative of $g$ at $h_t$ in the direction $a+b$, and the right side to the derivative in the direction $a$. Because $g$ is differentiable, this derivative is linear in the direction, so the two sides differ by the derivative in the direction $b$, which must therefore be zero. Every nonzero vector $x$ can be the hidden state $h_t$ of some input (place $x$ at position $t$ and any nonzero vectors elsewhere), so
\begin{equation*}
\text{the derivative of $g$ at $x$ in the direction $b$ is zero, for every $x\neq0$.} \tag{4}
\end{equation*}

\textit{Step 3: a vector the MLP ignores is zero.} Suppose that $b\neq0$, and take any $x$ that is not on the line through $0$ and $b$. The points $x+sb$, for real $s$, then never reach $0$, and by (4) the function $s\mapsto g(x+sb)$ has derivative zero everywhere. It is therefore constant, so $g(x+sb)=g(x)$ for every $s>0$. Since $g$ does not change under scaling, $g(x+sb)=g(x/s+b)$, which tends to $g(b)$ as $s$ grows. Hence $g(x)=g(b)$ for every $x$ off the line of $b$. A nonzero $x$ on that line is the limit of the points $x+w/n$ off the line, for any $w\neq0$ orthogonal to $b$, so $g(x)=g(b)$ there too, by continuity. Then $g$ gives the same output for every nonzero hidden state, which contradicts our assumption. Hence $b=0$, by Step 1 the change of attention is zero, and the field is zero for every input.
\end{proof}}

\update{\textbf{What the proposition covers.}}
\update{\begin{itemize}[leftmargin=1.5em,itemsep=2pt,topsep=2pt]
\item \textit{Methods.} It covers every change of the attention parameters: its biases, as in bias training, its weights, a LoRA on its projections, and the offsets of LoFiT \citep{yin2024lofit}, which add a fixed vector to the output of selected heads before the output projection. None of them acts as a shift adapter.
\item \textit{Models.} It applies to the pre-norm blocks of 16 of our 18 models. The other 2, the Gemma-2 models, also normalize the output of attention and of the MLP before adding it to the residual stream. They have no bias, so bias training does not apply to them.
\item \textit{Earlier blocks.} A change of the biases of an earlier block reaches this block only through its input. The proposition applies to the block that holds the changed parameter.
\end{itemize}}
\update{The proof treats the RMSNorm as exactly scale invariant. In practice a constant of $10^{-6}$ is added to the mean square for numerical stability, which at the hidden-state norms of our models changes the normalization by less than one part in $10^{9}$. The proof also uses that the MLP is differentiable, which holds for the gated MLPs of all 18 models.}

\section{Experimental protocol}\label{app:setup}

\subsection{The setup}\label{app:the-setup}

Unless a subsection states otherwise, every experiment uses the setup of \cref{sec:field}: the 18 models of \cref{app:models}, the 26 benchmarks of \cref{tab:tasks} (17 tasks and 9 behaviors) and 3 training seeds. Every method trains at the learning rate selected for it (\cref{app:training}). A run is one model, benchmark, method and seed, so a benchmark contributes 54 runs per method. Every reported number is the mean over the seeds, and every interval a $95\%$ bootstrap interval over models. Tasks are scored on their full evaluation split and behaviors on $1{,}000$ held-out generations per run. A subsection that departs from this setup, in its models, benchmarks, seeds or evaluation size, states its own design in its first paragraph.

\subsection{Models}\label{app:models}

The text experiments use 18 models from 7 architecture families.
Fourteen are instruction-tuned: Qwen2.5-0.5B, 1.5B, 3B and 7B, Qwen3-4B, Qwen3.5-4B and 9B, Qwen3.6-27B,
Llama-3.2-1B and 3B, Llama-3.1-8B, SmolLM2-1.7B, Mistral-7B and Gemma-2-2B. Qwen3.5 and Qwen3.6 interleave
three gated linear-attention blocks with every full-attention block, so the models are not one
architecture repeated at different scales.
Four further models are read before preference tuning (\cref{app:base-models}): the base models behind Llama-3.1-8B, Qwen2.5-7B and Gemma-2-2B, and Tulu-3-8B-SFT \citep{lambert2024tulu}, which is trained on instructions but not preference tuned. The convergence analysis in \cref{sec:impact} uses the 14 instruction-tuned models.
\subsection{Benchmarks}\label{app:tasks}

\Cref{tab:tasks} lists the 26 benchmark datasets: 17 tasks and 9 behaviors, including 4 traits, 3 transformations and 2 maps. 

\begin{table}[ht]
\caption{\textbf{The benchmarks.} The setup's 26 benchmarks: 17 tasks, scored by accuracy on their evaluation split, and 9 behaviors, scored on $1{,}000$ held-out generations (\cref{app:evaluation}). The four commonsense tasks, the three science tasks and HellaSwag follow the protocol of \citet{hu2023llmadapters} and \citet{liu2024dora}. The traits are persona and style targets of the kind used for activation steering \citep{turner2023actadd,rimsky2024caa}, and French and bullet points are also instruction types of IFEval \citep{zhou2023ifeval}, response language and bullet lists. The transformations are the copying and recall tasks used to study in-context learning \citep{olsson2022induction,arora2024zoology}, and the maps follow the associative recall tasks of \citet{arora2024zoology}. MMLU-Pro and SuperGPQA ship one labeled split, and we hold out a fixed part of it for evaluation.}
\label{tab:tasks}
\centering\footnotesize
\begin{tabular}{@{}l p{0.26\textwidth} p{0.34\textwidth} p{0.17\textwidth}@{}}
\toprule
class & name & what the model must do & scored on \\
\midrule
task & BoolQ, PIQA, SIQA, WinoGrande & commonsense question answering & validation split \\
task & ARC-Easy, ARC-Challenge, OpenBookQA & science question answering & test split \\
task & HellaSwag & sentence completion & validation split \\
task & SST-2 \citep{socher2013sst} & sentiment of a sentence & validation split \\
task & DBpedia, AG News \citep{zhang2015charcnn} & topic of an article & test split \\
task & CaseHOLD \citep{zheng2021casehold} & pick a case holding (law) & test split \\
task & MedMCQA \citep{pal2022medmcqa} & medical entrance-exam questions & validation split \\
task & AQuA-RAT \citep{ling2017aqua} & algebraic word problems & test split \\
task & SciQ \citep{welbl2017sciq} & school science questions & test split \\
task & MMLU-Pro \citep{wang2024mmlupro} & ten-way questions over fourteen subjects & held-out part \\
task & SuperGPQA \citep{map2025supergpqa} & ten-way graduate questions over 285 disciplines & held-out part \\
\midrule
trait & French & answer in French & token-F1 adherence \\
trait & pirate & answer in a pirate persona & token-F1 adherence \\
trait & grumpy critic & answer as a grumpy critic & token-F1 adherence \\
trait & bullet points & answer as a bulleted list & rule-based verifier \\
transformation & copy & repeat the input & exact match \\
transformation & last word & return the input's last word & exact match \\
transformation & conditional echo & echo the input when a condition holds & exact match \\
map & cipher & replace the input's first letter through a fixed substitution & exact match \\
map & lookup & answer with the word a fixed $26$-entry table assigns to the input's first letter & exact match \\
\bottomrule
\end{tabular}
\end{table}

\subsection{Behavior sets for the convergence analysis}\label{app:setup-depth}

Each behavior of the convergence analysis is a system prompt. In each model a shift adapter is distilled for every behavior on the same $160$ instructions for $300$ steps. The analysis runs on the $75$ behaviors that every atlas holds (\cref{tab:shared92}).

\begin{table}[h]
\centering
\caption{\textbf{The $75$ behaviors every atlas holds.} Each is a system prompt. \emph{Sycophantic}, for instance, is ``answer with fawning, excessive flattery of the asker'' and \emph{epistolary} is ``write every answer as a personal letter''.}
\label{tab:shared92}
\small
\begin{tabular}{@{}lp{0.78\textwidth}@{}}
\toprule
category & behaviors \\
\midrule
language (18) & catalan, dutch, finnish, french, german, hungarian, indonesian, italian, norwegian, polish, portuguese, romanian, russian, spanish, swahili, swedish, tagalog, vietnamese \\
era (14) & 1980s, age of sail, ancient rome, belle epoque, cyberpunk future, dotcom 1999, enlightenment, industrial revolution, medieval, medieval bard, roaring twenties, space age 1960s, stone age, wild west \\
persona (13) & carnival barker, detective noir, excited child, fortune teller, grumpy critic, grumpy wizard, pirate, polite butler, sarcastic teen, shakespeare, valley girl, victorian gentleman, zen monk \\
format (9) & all caps, bullets, haiku, limerick, morse style, one word, rhymes, screenplay, telegram \\
humor (4) & absurdist, observational, punny, slapstick \\
rhetoric (4) & deadpan, metaphorical, ominous, pep talk \\
affect (3) & conspiratorial, passive aggressive, sycophantic \\
audience (3) & child genius, eli5, toddler \\
emotion (2) & joyful, melancholic \\
pov (2) & epistolary, omniscient \\
domain (1) & world \\
epistemic (1) & hedging \\
register (1) & hardboiled \\
\bottomrule
\end{tabular}
\end{table}

\subsection{Instructions, teachers and training}\label{app:training}

\begin{wrapfigure}{r}{0.42\linewidth}
\centering
\includegraphics[width=\linewidth]{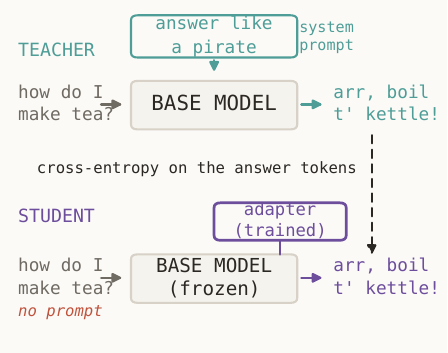}
\vspace{-1.6em}
\caption{\textbf{How a trait is trained.} The base model answers an instruction under the trait's system prompt (teacher). The same base model, frozen and without the prompt, is adapted to reproduce that answer, with cross-entropy on the answer tokens (student).}
\label{fig:method-train}
\end{wrapfigure}

\textbf{Instructions and teachers.} Every trait is trained and scored on instructions from the no-context part of dolly-15k \citep{conover2023dolly}, about $15$k instructions of $20$ to $160$ characters with duplicates removed. Each seed draws its own training instructions, and the held-out instructions a behavior is scored on never overlap the ones it trained on. The number of instructions is fixed within each experiment for every model and method it compares, and follows how many states the experiment trains. A behavior of the setup trains on $80$ instructions for $250$ steps, and a map on $160$ for $400$, so that every first letter occurs in training. An atlas, which trains hundreds of behaviors per model, uses $160$ for $300$ steps, as do the in-context comparison and the behavior counts along the training of OLMo-2-7B. A region grid, which scores hundreds of points around one state, uses $60$ for $300$, and the steering comparison uses $400$ for $200$. The teacher is the base model under a system prompt for a trait, and the rule itself for a transformation or a map.

\begin{figure}[h]
\centering
\includegraphics[width=\textwidth]{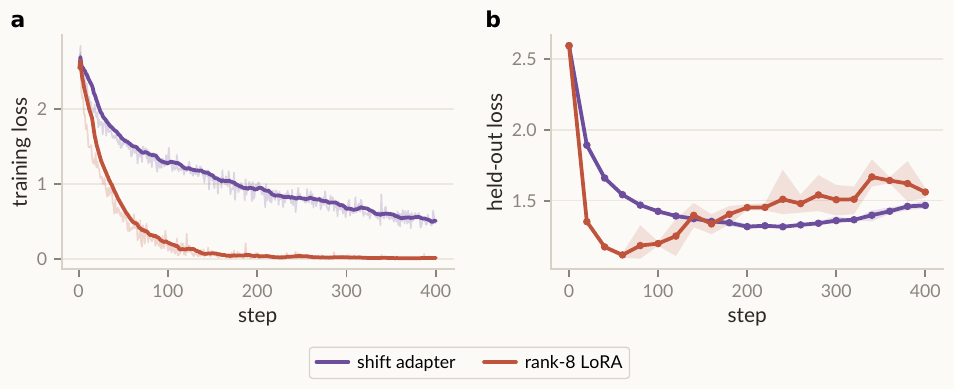}
\caption{\textbf{Loss and adherence rank the two adapters differently} (Qwen2.5-1.5B on pirate, the shift adapter and the rank-8 LoRA trained on the same teacher answers). Both losses are the next-token cross-entropy on the tokens of the teacher's answer, with the instruction masked. \textbf{(a)}~On the training answers, per step (faint: raw, bold: rolling mean). \textbf{(b)}~On held-out answers, every $20$ steps, with min--max bands over seeds. The model is given the teacher's answer and scored on predicting each next token of it, so nothing is generated. The LoRA reaches near-zero training loss and its held-out loss turns up, while the shift adapter's falls slowly and stays flat. Scored on generated held-out answers, the two adapters reach an adherence of $0.05$ (shift adapter) and $0.08$ (LoRA), a difference of $0.03$, which is why methods are compared on generated text and loss stays a diagnostic.}
\label{fig:curves}
\end{figure}

\textbf{Objective and defaults.}
The adapter (a shift adapter or a LoRA) is trained on the teacher's answers to bare prompts: the loss is next-token cross-entropy on the answer tokens only, with instruction and padding tokens masked, and AdamW updates the adapter parameters alone. The default rates are $10^{-3}$ for shift adapters and $2\times10^{-4}$ for LoRA, and on the tasks the rate chosen by the selection below, for $250$--$800$ steps per behavior and a batch of $4$ unless stated. \Cref{fig:curves} shows the training and held-out loss curves of one model on one trait. Loss is a diagnostic: every claim is scored on generated text (\cref{app:evaluation}).

\textbf{Learning-rate selection.}
For the 17 tasks of \cref{fig:parity} the learning rate is selected \emph{per model, adapter, and task} with the same procedure, so that no adapter is compared at a tuned rate against another at an untuned one. For each task we split the training data in two: a validation part and the rest. Every rate is trained on the rest and scored on the validation part, and the rate that scores highest is kept, ties within $0.005$ going to the smaller rate.
The adapter is then retrained on the full training data at that rate, for every seed. The evaluation split is never used for selection, and no rate carries from one task to another.
The rates are $\{1,5\}\times10^{-5}$, $\{1,5\}\times10^{-4}$, $\{1,2,5\}\times10^{-3}$ and $10^{-2}$ for the shift adapter, $\{1,2,5\}\times10^{-5}$ and $\{1,2,5\}\times10^{-4}$ for LoRA, and $\{3\times10^{-7}, 10^{-6}, 3\times10^{-6}, 10^{-5}, 3\times10^{-5}, 10^{-4}\}$ for full fine-tuning.

\textbf{Parameter counts.} Adapter sizes depend on the model's depth $L$, its width $d$ and the adapted matrices $m$. A shift adapter trains $L\times d$ parameters, $0.04$M on Qwen2.5-1.5B, and a rank-$r$ LoRA trains $r\sum_m(d^{(m)}_{\mathrm{in}}{+}d^{(m)}_{\mathrm{out}})$, $9.2$M for $r{=}8$ on the same model.

\subsection{Evaluation}\label{app:evaluation}

Every claim is scored on generated text for held-out instructions, never on teacher-forced loss. The format behaviors (bullet points, all caps, JSON, one word and socratic) are scored by a rule-based verifier, as the adapted rate minus the base model's rate. Every other trait is scored by the token-F1 similarity of the generation to the teacher's generation, minus the same similarity for the base model. We call this score \emph{adherence}. Transformations and maps are scored by exact match.

\textbf{Direct checks of the language and persona scores.}
We rescore the saved generations of the interventions of \cref{sec:field} with a direct measure, without new training or generation. A frozen $97$-language identifier finds the target language in under $1\%$ of the base model's answers and in $67\%$ to $100\%$ of the trained adapter's, on every one of the 18 models. Counting an answer as French when the identifier labels it French, across the 18 models, keeping only the shared component keeps $99\%$ of the gain and removing it keeps $16\%$ (medians over runs), close to the token-F1 values. A lexical marker gives the same ordering for the personas: with only the shared component, pirate keeps $99\%$. The result on the languages does not depend on token F1. Token F1 scores a generation by its overlap with the teacher's answer. A different wording of the same behavior scores low. The direct checks above read the behavior itself, and they give the same ordering.

\subsection{Field statistics}\label{app:field-stats}

For $n$ probe inputs, the cosine and the field share of \cref{sec:field} at layer $\ell$ and position $t$ are
\begin{equation*}
C_{\ell,t}=\frac{1}{n}\sum_{i=1}^{n}
\frac{\langle\dh_{\ell,t}(x_i),\bar\dh_{\ell,t}\rangle}
{\lVert\dh_{\ell,t}(x_i)\rVert\,\lVert\bar\dh_{\ell,t}\rVert},
\qquad
E_{\ell,t}=\frac{\lVert\bar\dh_{\ell,t}\rVert^2}
{n^{-1}\sum_i\lVert\dh_{\ell,t}(x_i)\rVert^2}.
\end{equation*}
A layer-averaged value weights every layer the same, and averages only the layers the adapter reaches: the field is zero at the embedding output, where $C_{\ell,t}$ is undefined, and including those layers would lower every method's cosine by about $0.07$.
The cosine between two shared components, $\cos(\bar\dh^{A}_{\ell,t},\bar\dh^{B}_{\ell,t})$ of \cref{fig:alignment}, is averaged the same way.
Weighting each layer by how far the adapter moves it, $\lVert\bar\dh^{A}_{\ell,t}\rVert\lVert\bar\dh^{B}_{\ell,t}\rVert$, changes no ranking and sharpens both comparisons: on the 9 behaviors and 18 models, the shift adapter, the LoRA and a random LoRA reach $0.79$, $0.78$ and $0.56$ against the unweighted $0.85$, $0.84$ and $0.66$, and the shift adapter against the LoRA's shared component reaches $0.69$ where a random LoRA stays at $0.08$.
A layer where either component is near zero has no direction to compare, which is what the weighting removes.

\subsection{Intervals}\label{app:uncertainty}

Unless a subsection states otherwise, brackets are bootstrap $95\%$ intervals that resample models (\cref{app:the-setup}), and a subsection that resamples runs, tasks or concepts says so. A plain range ($a$ to $b$) is a minimum and a maximum over the stated models. Where per-probe scores are available, we report the interval over runs beside a bootstrap over probes within runs (\cref{app:robust}).

The comparison of models (\cref{app:crossmodel}) uses $91$ pairs of models. Each model appears in 13 pairs, so the pairs are dependent, and we attach no $p$-value that assumes independence to their correlation with capability: that result uses a permutation of model labels that preserves the pairs, and $2{,}000$ permutations of behavior labels per pair calibrate each pair's similarity.

\FloatBarrier
\section{Shared and input-specific components}\label{app:components}

\subsection{The two interventions}\label{app:interventions}

The two interventions of \cref{sec:field} are reported as fractions of the adapter's gain over the base model. For an adaptation $A$ whose added vector can be isolated, and a held-out score $S$, accuracy or adherence, we define
\begin{equation}\label{eq:certificate}
R_{\mathrm{keep}}^A=\frac{S(A_{\mathrm{mean}})-S(\mathrm{base})}
{S(A)-S(\mathrm{base})},\qquad
N_{\mathrm{mean}}^A=\frac{S(A)-S(A_{\mathrm{remove\ mean}})}
{S(A)-S(\mathrm{base})}.
\end{equation}
The fraction of the gain kept with only the shared component is $R_{\mathrm{keep}}^A$, and the fraction kept with it removed is $1-N_{\mathrm{mean}}^A$. Over a set of runs, each fraction is a ratio of sums, the summed gain that remains over the summed gain, and it takes every run, so a run with a small gain carries little weight. A fraction for a single run divides by that run's own gain, so a median over runs (\cref{app:phase}) uses only the runs whose gain reaches $0.05$. \Cref{fig:shared} draws both fractions.

\subsection{\update{Interventions and field statistics on the tasks}}\label{app:interventions-bench}

\begin{figure}[ht]
\centering
\includegraphics[width=\textwidth]{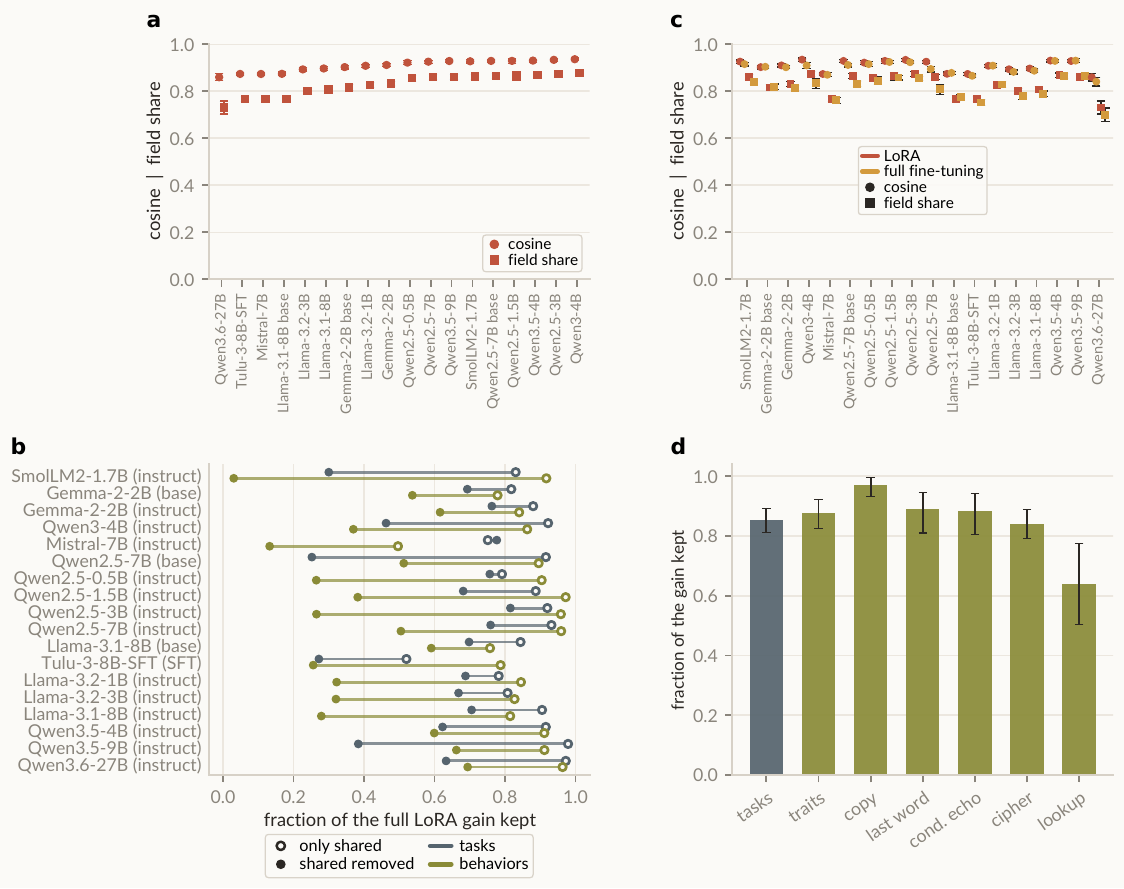}
\caption{\textbf{\Cref{fig:shared} with every model and every task.}
\textbf{(a)}~The two statistics of the decomposition for LoRA, per model, on the 17 tasks.
\textbf{(b)}~The fraction of the LoRA gain kept with only the shared component (hollow) and with it removed (filled), per model, on the 17 tasks and on the 9 behaviors, grouped by family and reading base, supervised-only, instruction-tuned.
\textbf{(c)}~The same two statistics for full fine-tuning against LoRA, on the 17 tasks and 18 models.
\textbf{(d)}~The fraction kept with only the shared component for the 17 tasks pooled and for each of the 9 behaviors, traits pooled.}
\label{fig:shared-all}
\end{figure}

We apply the two interventions to the rank-8 LoRAs trained on the 17 tasks of \cref{app:tasks}, on the 18 models. Each intervention is scored by accuracy on the task's full evaluation split, and the shared component is estimated from $300$ held-out probes. \Cref{fig:shared}b draws one model per family, and \cref{fig:shared-all} draws every model.

With only the shared component, the LoRA keeps $85\%$ [$81$, $89$] of its gain, and with it removed $61\%$ [$52$, $69$]. The LoRA gains less than $0.05$ in $325$ of the $918$ runs, most of them on MedMCQA, SciQ, the two ARC sets, SST-2 and SIQA, and these runs carry little weight in the ratio of sums. Per task, the fraction kept with only the shared component ranges from $66\%$ on HellaSwag to $99\%$ on DBpedia, and the fraction kept with it removed from $17\%$ on ARC-Easy to $88\%$ on AG News.
By task type, with intervals over models, the fractions kept with only the shared component and with it removed are $97\%$ [$96$, $98$] and $59\%$ [$46$, $71$] on the three text classifications, $87\%$ [$83$, $90$] and $58\%$ [$50$, $66$] on the four commonsense question-answering tasks, $78\%$ [$70$, $87$] and $49\%$ [$18$, $74$] on the three science question-answering tasks, $79\%$ [$73$, $85$] and $73\%$ [$64$, $82$] on the four domain tasks, $66\%$ [$45$, $79$] and $61\%$ [$31$, $77$] on HellaSwag, and $69\%$ [$60$, $76$] and $66\%$ [$58$, $75$] on the two expert question-answering tasks, MMLU-Pro and SuperGPQA. The shared component alone keeps at least $66\%$ of the gain on every task type.

Two checks concern how the fractions are measured. First, estimating the shared component as the average input of the matrix over all prompt positions, instead of at the last token of the input, keeps $70\%$ [$66$, $73$] of the gain with only that component and $93\%$ [$91$, $94$] with it removed. The gain relies on the component the adapter adds at the last token of the input, on the tasks as on the traits (\cref{app:robust}). Second, macro accuracy stays close to accuracy: within $0.037$ on those runs, and on the setup's runs, each scored under the trained adapter and the two interventions, within $0.04$ in all but 4 of the $2{,}754$ reports (at most $0.083$, under removal), with a mean gap of $0.00$. No fraction comes from predicting the majority class.

\textbf{Removing the shared component position by position.} The interventions estimate one shared component at the last token of the input and apply it at every position. To test what the earlier positions share, we estimate a second shared component over the positions before the last token, on the same held-out probes, and remove each component where it was estimated: \textit{(i)} the component estimated over the prompt from the prompt positions, and \textit{(ii)} the component estimated at the last token from that token. Across the 18 models and 17 tasks, removing both in this way keeps $84\%$ of LoRA's gain, against $61\%$ when the single component is removed at every position. Removing the direction of the shared component, by projecting the field at every position onto the subspace orthogonal to it so that nothing the adapter adds points along it, keeps $75\%$. Each of the 3 removals leaves a different part of the update, and each keeps more than half of the gain.

\update{\textbf{The two methods solve the same items.} Matched accuracy does not tell whether the shift adapter and LoRA solve the same held-out items. On the 17 tasks and 18 models, we take the items the base model gets wrong and compare the ones each method newly solves. Of the items one method newly solves, the other newly solves $82\%$ (median over the runs), against $51\%$ if the two methods picked their items at random from the base model's failures. In $916$ of the $918$ runs (one per model, task, and seed), the observed share is above this chance level. It is above chance on every one of the 17 tasks (\cref{fig:output-agreement}a), and most clearly on the harder tasks, such as MedMCQA ($68\%$ against $22\%$) and ARC-Challenge ($74\%$ against $30\%$). Chance is high only where both methods solve almost every item the base model gets wrong, as on DBpedia, since two nearly complete sets of items always overlap. Per model, it runs from $73\%$ on Mistral-7B to $91\%$ on the Llama-3.1-8B base model (\cref{fig:output-agreement}b). Two seeds of one method agree on the items to the same degree: of the items one seed of the shift adapter newly solves, another seed solves $85\%$, and for LoRA $82\%$. The shift adapter and LoRA thus solve the same items as closely as one method does across its own seeds.}

\begin{rfigure}[h]
\centering
\updatebox{\includegraphics[width=\dimexpr\textwidth-8.4pt\relax]{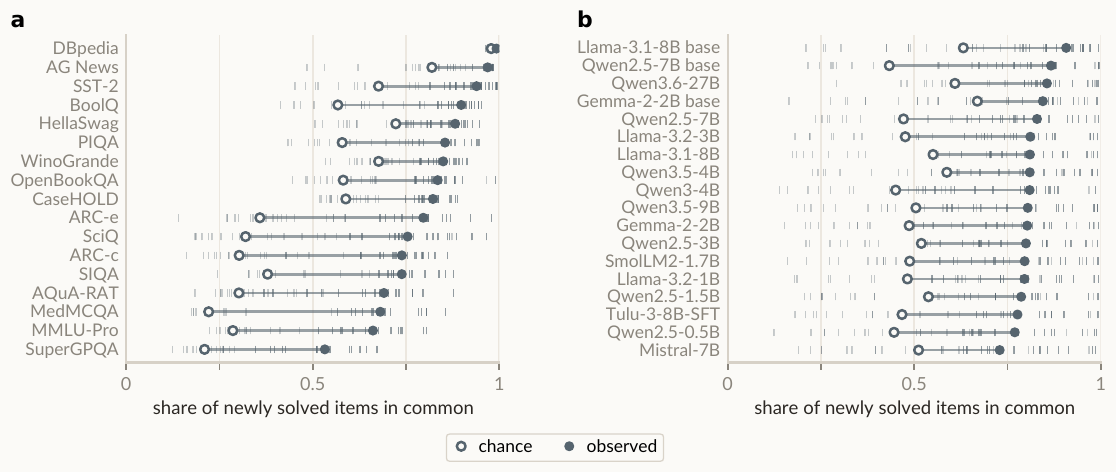}}
\caption{\update{\textbf{The shift adapter and LoRA solve the same items.} For each task \textbf{(a)} and each model \textbf{(b)}, of the held-out items one method newly solves (items the base model gets wrong), the share the other method newly solves too: observed (filled) against two random draws of the same sizes from the base model's failures (hollow). The seeds pool per model and task, each row is the median over the other axis, and the ticks are the values it summarizes.}}
\label{fig:output-agreement}
\end{rfigure}

\subsection{\update{The shared component inside the adapter}}\label{app:interventions-update}

The interventions act on what the adapter adds at each adapted matrix, $a(x)=\dW z(x)$, so its decomposition in \cref{eq:decomp} can be measured directly, without the hidden states. For each adapted matrix we compute the share of the energy of what the adapter adds that its shared component holds, $\lVert\bar a\rVert^2/\E_x\lVert a(x)\rVert^2$, the field share of \cref{sec:field} applied to $a$ in place of $\dh$. Over every adapted matrix of the runs on the 9 behaviors and the 18 models, the shared component holds $89\%$ of the energy on average (median $96\%$), from $66\%$ on the Gemma-2-2B base model to $99\%$ on SmolLM2-1.7B. For the update without the rank constraint the median is $98\%$ (\cref{app:rankfree}). At almost every adapted matrix, LoRA adds nearly the same vector for every input.

\begin{rfigure}[t]
\centering
\updatebox{\includegraphics[width=\dimexpr\textwidth-8.4pt\relax]{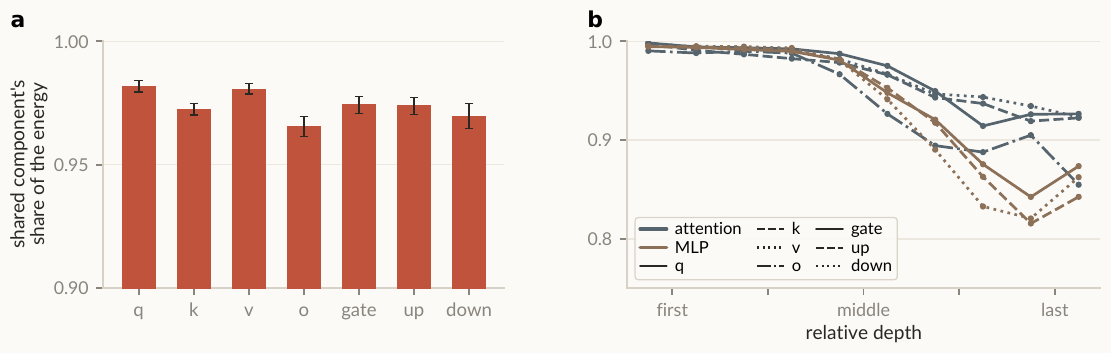}}
\caption{\update{\textbf{Share of the energy of what LoRA adds that is the same for every input, per adapted matrix, on the tasks.} Rank-8 LoRAs on the 17 tasks and 18 models. \textbf{(a)}~Median over every adapted matrix by matrix kind, bars are $95\%$ bootstrap intervals over runs. \textbf{(b)}~The same share against the matrix's relative depth, one line per kind, attention and MLP matrices apart.}}
\label{fig:site-energy}
\end{rfigure}

\update{We repeat this measurement on the 17 tasks and the 18 models, and the result is the same: at almost every adapted matrix, LoRA adds nearly the same vector for every input. The shared component holds $89\%$ of the energy on average (median $98\%$), from $82\%$ on Tulu-3-8B-SFT to $93\%$ on Qwen3.6-27B, and between $97\%$ and $98\%$ for every kind of matrix (\cref{fig:site-energy}a). The share stays above $99\%$ in the first tenth of the layers and falls only toward the last layers, to $92\%$ for the query, key, and value matrices and to $84$ to $87\%$ for the output and MLP matrices in the last tenth (\cref{fig:site-energy}b). The part of the update that differs between inputs is thus small, and it is concentrated in the last layers.}

\subsection{Models before instruction tuning}\label{app:base-models}

Instruction tuning could be what makes adaptation rely on a shared component. We test this on the 3 models that never received instruction tuning: the base models behind Llama-3.1-8B, Qwen2.5-7B and Gemma-2-2B. They run on the same 17 tasks.
With only the shared component, the LoRA keeps $85\%$ [$82$, $92$] of its gain, and with it removed $59\%$ [$25$, $70$], against $87\%$ [$84$, $91$] and $63\%$ [$53$, $72$] on the 14 instruction-tuned models. Per model, the fraction kept with only the shared component is $84\%$ for Llama-3.1-8B, $92\%$ for Qwen2.5-7B and $82\%$ for Gemma-2-2B.

On the 9 behaviors, the same 3 models keep $81\%$ [$76$, $90$] of the gain with only the shared component and $55\%$ [$51$, $59$] with it removed, against $87\%$ [$80$, $92$] and $39\%$ [$29$, $49$] on the 14 instruction-tuned models. Per model, the fraction kept with only the shared component is $76\%$ for Llama-3.1-8B, $90\%$ for Qwen2.5-7B and $78\%$ for Gemma-2-2B (\cref{fig:shared-all}b). Adaptation relies on the shared component before instruction tuning, on tasks and behaviors alike.

\subsection{Interventions on the traits and transformations}\label{app:interventions-behaviors}

\begin{wrapfigure}{r}{0.38\linewidth}
\vspace{-1.9em}
\centering
\includegraphics[width=0.92\linewidth]{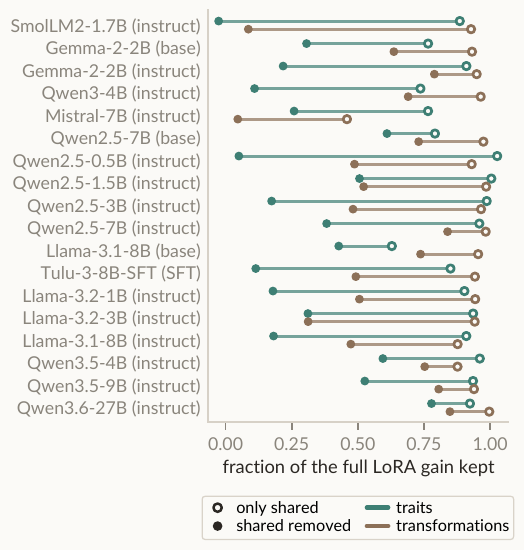}
\vspace{-0.8em}
\caption{\textbf{The two interventions per model on the traits and transformations.} Fraction of the LoRA gain kept with only the shared component (hollow) and with it removed (filled), 18 models, in the order of \cref{fig:shared-all}b.}
\label{fig:shared-traits}
\vspace{-2em}
\end{wrapfigure}
We apply the two interventions to LoRAs trained on the 4 traits and the 3 transformations. What the LoRA adds at an adapted matrix is $a(x)=\dW z(x)$. Keeping only the shared component replaces it by $\bar a=\dW\mu$ at every position, while the input is read and at every generated token, where $\mu$ is the mean input of the matrix at the last token of held-out probes: held-out instructions given as bare prompts, disjoint from training. Removing the shared component keeps $a(x)-\bar a=\dW(z(x)-\mu)$ at every position. The estimate of $\mu$ does not depend on which probes or how many are used (\cref{app:robust}).

Across the 18 models (14 instruction-tuned and the 4 of \cref{app:setup}), the LoRA keeps $88\%$ [$83$, $92$] of its gain on the traits and $92\%$ [$85$, $96$] on the transformations with only its shared component, and $33\%$ [$23$, $43$] and $57\%$ [$46$, $67$] with it removed. These intervals resample models. \Cref{fig:shared-traits} draws every model, with the traits and the transformations apart.

The adaptation field of these LoRAs points along its shared component at a cosine of $0.84$ [$0.80$, $0.87$], and the shared component holds $71\%$ [$65$, $76$] of the field's energy, over the 18 models on the 9 behaviors. Both statistics are averaged over the layers the adapter reaches, as on the tasks (\cref{app:field-stats}). On behaviors as on tasks, the shared component alone keeps most of the gain, and removing it keeps far less.

\subsection{Interventions on the maps}\label{app:controls}

The cipher and the lookup are the two behaviors whose rule the model can only learn during adaptation. Over the 18 models, the LoRA keeps $84\%$ [$79$, $89$] of its gain on the cipher and $64\%$ [$50$, $77$] on the lookup with only its shared component, and $33\%$ [$20$, $46$] and $19\%$ [$9$, $30$] with it removed. The lookup keeps the least of any benchmark, as expected for a table of new content (\cref{app:few-items}), so part of the table sits in the LoRA's input-specific component. A shift adapter trained directly on the lookup learns it (\cref{app:express}).

\subsection{Reading and writing}\label{app:phase}

\begin{wrapfigure}[18]{r}{0.36\linewidth}
\vspace{-1.6em}
\centering
\includegraphics[width=\linewidth]{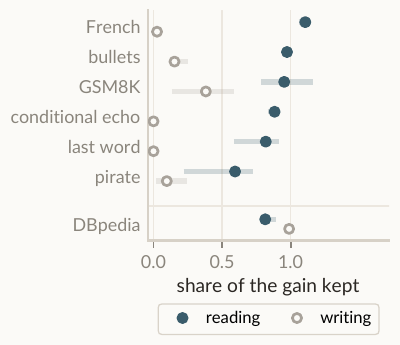}
\vspace{-2.0em}
\caption{\textbf{Adaptation acts while the model reads.} Fraction of the gain kept with the adapter active only while reading or only while writing, per task, median and quartiles over the runs of the 18 models whose gain reaches $0.05$. On DBpedia, a classification task scored on one label token, either phase alone keeps most of the gain.}
\label{fig:phase}
\vspace{-1.5\baselineskip}
\end{wrapfigure}

The interventions change what the adapter adds at every position. Here we ask at which positions the adapter has to be active at all. We keep the trained update unchanged in one phase and switch it off in the other. Active only while reading, the update is switched off at every generated position. Active only while writing, it is switched off while the input is read. Switching off acts on the same adapted matrices as the interventions. The first generated token comes from the forward pass over the input, so it belongs to the reading phase. We report each condition as a fraction of the run's gain (\cref{fig:phase}).

Across the 18 models, on the 6 generative behaviors and tasks (bullets, French, pirate, last word, conditional echo and GSM8K), over the $324$ runs, the update active only while reading keeps $83\%$ of the gain, and active only while writing $12\%$. In a quarter of the $303$ runs whose gain reaches $0.05$, reading alone keeps more than the full update, up to $165\%$, so switching the adapter off during generation can raise adherence.

Two cases differ. On label-token classification, either phase alone suffices: the update at the last input position alone keeps $92\%$ to $101\%$ of the gain, and the update everywhere except that position keeps $68\%$ to $100\%$ (medians over the seeds of each of the 18 models, and $82\%$ reading and $99\%$ writing on DBpedia pooled, \cref{fig:phase}). There is no generated text, and the update at the last position adds directly to the label logit. The order reverses on three behaviors of the base models: pirate on the base models behind Llama-3.1-8B and Gemma-2-2B, where writing keeps $76\%$ and $81\%$ of the gain against $-1\%$ and $4\%$ for reading, and bullets on the Gemma-2-2B base model, where writing keeps $113\%$ against $41\%$, every seed in the same order. In their saved generations the update active only while reading sets the first word and the model continues in its own register, and active only while writing the model starts in its own register and takes the persona or the list form within a sentence. Pirate on the Qwen2.5-7B base model leans the same way by a smaller margin ($62\%$ against $47\%$), with both phases producing the persona. On Gemma-2-2B with bullets the two phases tie ($61\%$ against $56\%$, the seeds split).

\textbf{The shared component alone, in each phase.} The test above switches the whole adapter on or off. We combine it with the first intervention to ask whether the shared component alone is enough within each phase. On 6 models (Llama-3.2-1B and 3B, Qwen2.5-1.5B and 3B, SmolLM2-1.7B and Gemma-2-2B), we apply only the shared component $\dW\mu$ during one phase and nothing during the other, and compare it with the full adapter active in the same phase.

While the input is read, the shared component alone keeps a median $79\%$ of the gain over the $90$ generative runs whose gain reaches $0.05$, against $92\%$ for the full adapter. The difference comes from one behavior on two models: pirate on Llama-3.2-3B and Llama-3.2-1B, where the shared component keeps $37\%$ and $50\%$ against $69\%$ and $76\%$ for the full adapter. It does not come from pirate needing the input-specific component: applied in both phases, the same shared component keeps $89\%$ to $97\%$ on these models. In the saved generations, a run takes the persona in its first words or not at all.

While the answer is written, the shared component alone keeps as much of the gain as the full adapter in all $30$ medians over family and behavior, including Gemma-2-2B with bullets, where the two phases carry the same share ($67\%$ against $66\%$). On the transformations both keep $0\%$. Wherever a phase carries part of the gain, the shared component alone carries the same part.

\subsection{The estimate of the shared component and the evaluation size}\label{app:robust}

The shared component is estimated from held-out probes. We check on 3 models from 3 families (Qwen2.5-1.5B, Llama-3.2-3B and SmolLM2-1.7B) that the number of probes does not move the fractions.
The estimate converges quickly: its mean cosine to the estimate from all $200$ probes is $0.988$ from $10$ probes, $0.995$ from $20$ and $0.999$ from $100$. The fraction kept with only the shared component does not change with the number of probes: $94\%$ [$90$, $97$] on the traits whether $\mu$ comes from $20$, $40$ or $100$ probes (largest change $0.001$), and $95\%$ on the transformations.
Most of the uncertainty comes from differences between runs: in adherence, the interval over runs for the drop on the traits, [$0.013$, $0.037$], is wider than the bootstrap over probes within runs, [$0.021$, $0.028$], so measurement noise within a run does not drive the pooled interval.
Estimating $\mu$ as the average input over every position of the prompt, instead of at its last token, keeps $55\%$ [$44$, $66$] of the gain on the traits against $87\%$ [$82$, $92$] at the last token, over the 18 models. The behavior relies on the component the adapter adds at the last token of the input.

\FloatBarrier
\subsection{\update{The update without the rank constraint}}\label{app:rankfree}

LoRA factorizes its update into two low-rank matrices, so the shared component could be a product of that factorization. We therefore apply the two interventions to an update that is not factorized. We train the matrices that a rank-8 LoRA adapts directly, with the ordinary optimizer, and keep every other parameter frozen, so weight decay pulls the update toward zero and not toward the base weights. The update is $W_{\mathrm{FT}}-W_0$, and we intervene on it exactly as on LoRA's, at every position of the input and the answer. The setup follows \cref{app:interventions-behaviors}, on the 18 models, the 4 traits and the 3 transformations. On the two maps, with only the shared component LoRA keeps $79\%$ [$70$, $87$] of its gain and the unconstrained update $54\%$ [$46$, $61$], and with it removed $26\%$ [$17$, $37$] and $19\%$ [$10$, $29$].

The update has high rank, and training without weight decay leaves its stable rank and the shared component's share of the energy unchanged on the 18 models ($47.8$ and $97.2\%$, against $46.6$ and $97.2\%$ with it), so weight decay does not produce the high rank. Its shared component holds $95\%$ of the energy of what it adds (median $98\%$), against $89\%$ for LoRA on the same runs (\cref{app:interventions-update}), so the concentration does not come from the parameterization. The two methods gain about the same, a mean trait gain of $0.442$ for LoRA and $0.437$ without the rank constraint, so their fractions compare directly. Removing the whole update returns the model to the base model within $0.020$ in every run, so the interventions reach all of the update. The table below gives both fractions per model for both methods, with the stable rank and the share of the energy of the unconstrained update.

\begin{center}\small\setlength{\tabcolsep}{4pt}
\begin{tabular}{lcccccc}
\toprule
& \multicolumn{2}{c}{LoRA} & \multicolumn{2}{c}{unconstrained update} & \multicolumn{2}{c}{update spectrum} \\
\cmidrule(lr){2-3}\cmidrule(lr){4-5}\cmidrule(lr){6-7}
model & shared only & without it & shared only & without it & stable rank & share of the energy \\
 & \multicolumn{4}{c}{\footnotesize fraction of the gain kept, both classes pooled} & & \\
\midrule
Llama-3.1-8B (base) & $0.840$ & $0.584$ & $0.722$ & $0.311$ & $50$ & $0.810$ \\
Llama-3.1-8B & $0.893$ & $0.466$ & $0.896$ & $0.182$ & $60$ & $0.976$ \\
Tulu-3-8B-SFT & $0.883$ & $0.543$ & $0.929$ & $0.342$ & $61$ & $0.975$ \\
Llama-3.2-1B & $0.921$ & $0.478$ & $0.725$ & $0.436$ & $60$ & $0.967$ \\
Llama-3.2-3B & $0.920$ & $0.378$ & $0.876$ & $0.196$ & $61$ & $0.970$ \\
Mistral-7B & $0.474$ & $0.183$ & $0.128$ & $-0.146$ & $34$ & $0.944$ \\
Qwen2.5-0.5B & $0.958$ & $0.402$ & $0.793$ & $0.428$ & $45$ & $0.965$ \\
Qwen2.5-1.5B & $0.975$ & $0.512$ & $0.785$ & $0.382$ & $64$ & $0.978$ \\
Qwen2.5-3B & $0.960$ & $0.369$ & $0.543$ & $0.346$ & $53$ & $0.983$ \\
Qwen2.5-7B (base) & $0.917$ & $0.742$ & $0.924$ & $0.593$ & $55$ & $0.982$ \\
Qwen2.5-7B & $0.952$ & $0.738$ & $0.687$ & $0.505$ & $51$ & $0.987$ \\
Qwen3-4B & $0.869$ & $0.455$ & $0.845$ & $0.036$ & $52$ & $0.979$ \\
Qwen3.5-4B & $0.950$ & $0.776$ & $0.914$ & $0.797$ & $47$ & $0.959$ \\
Qwen3.5-9B & $0.950$ & $0.673$ & $0.938$ & $0.677$ & $56$ & $0.979$ \\
Qwen3.6-27B & $0.977$ & $0.829$ & $0.555$ & $0.881$ & $62$ & $0.952$ \\
SmolLM2-1.7B & $0.929$ & $0.012$ & $0.021$ & $0.542$ & $113$ & $0.982$ \\
Gemma-2-2B (base) & $0.816$ & $0.527$ & $0.681$ & $0.262$ & $41$ & $0.823$ \\
Gemma-2-2B & $0.897$ & $0.504$ & $0.851$ & $0.245$ & $39$ & $0.962$ \\
\bottomrule
\end{tabular}
\end{center}
Pooled over runs as fractions of each method's own gain, with only the shared component LoRA keeps $87\%$ [$82$, $91$] on the traits and the unconstrained update $69\%$ [$57$, $80$], and on the transformations $91\%$ [$83$, $96$] against $74\%$ [$61$, $84$]. With the shared component removed, LoRA keeps $39\%$ [$28$, $51$] against $26\%$ [$13$, $39$] on the traits, and $59\%$ [$49$, $68$] against $49\%$ [$37$, $61$] on the transformations. These intervals resample models. Resampling runs, the shared component alone keeps $87\%$ [$83$, $90$] against $69\%$ [$63$, $75$] on the traits and $91\%$ [$88$, $94$] against $74\%$ [$69$, $78$] on the transformations, and the intervals of the two methods separate in both classes.
The gap varies by model: with only the shared component, the two methods are close on Llama-3.1-8B and Qwen3-4B and far apart on SmolLM2-1.7B.

\update{\textbf{Mistral-7B.} A shift adapter trained directly on Mistral-7B reaches $97\%$ of LoRA's gain on the traits and $92\%$ on the transformations (\cref{app:icl}), so a constant suffices on this model. Its LoRA keeps the least with either component alone, $47\%$ of LoRA's gain with only the shared component and $18\%$ with it removed, so neither component alone carries most of its gain. With only the shared component, it keeps $76\%$ on the traits, close to the other models, and $30\%$ on the transformations. We measure how strongly the full update changes each model by the perplexity on WikiText. The unconstrained update raises the perplexity of Mistral-7B by a median factor of $3.1$, against $1.0$ for Llama-3.1-8B and Qwen2.5-7B, so Mistral-7B reacts more strongly to an update, already before any component is removed, and its low fractions do not show a missing capacity. On the 17 tasks, Mistral-7B has two routes to its gain: its LoRA keeps $75\%$ of the gain with only the shared component and $78\%$ with it removed, while for the largest instruction-tuned model of each other family, removing the shared component keeps $30\%$ to $76\%$.}

Without the rank constraint, the shared component holds more of the energy of the update and the behavior relies on it somewhat less, so the reliance does not come from the low rank.

\FloatBarrier
\subsection{In-context learning}\label{app:icl}

In-context learning adapts without training, through demonstrations in the prompt. We apply the two interventions to the demonstrations and to a LoRA trained in the same runs, and call each of these two ways of adapting a \emph{route}. A shift adapter trained in the same runs is a third route.

\textbf{Experimental setup.} We use the 18 models and the 9 behaviors of the setup: the traits French, bullet points, pirate and grumpy critic, the transformations copy, last word and conditional echo, and the maps cipher and lookup.
In each run the demonstrations are four question--answer pairs before the query. For the traits, the answers are the model's own answers under the behavior's system prompt, each passing the behavior's rule-based verifier, and the prompt uses the chat template. For the transformations and maps, the answers are the task's targets, and the prompt is plain text behind a preamble that asks the model to continue the pattern, because an instruction-tuned model under its template answers the last question instead. The maps get their whole $26$-word table before the 4 examples, since 4 examples cover only 4 of its $26$ entries.
The field of the demonstrations is the change they make to the hidden state at the last token of the input, $h(\text{demonstrations}+x)-h(x)$, and its shared component is the mean of that change over $40$ probe queries. To keep only the shared component, we add it to the bare query, one vector per layer at every position of the input and the answer, scaled by $\alpha\in\{0.5,1,2\}$ chosen per family and class on seed $0$. To remove it, we keep the demonstrations and subtract it at every position. As a control, a random vector of the same norm per layer is added and subtracted in the same way.
A rank-8 LoRA and a shift adapter are trained in the same run on the same $160$ answers under the same prompt format, and the LoRA is run as trained, with only its shared component, and without it (\cref{app:interventions}). Every route is scored on the same $48$ held-out queries, by the behavior's rule-based verifier or, for pirate and grumpy critic, by token F1 against the model's own answers under the system prompt. Fractions pool every run, and intervals resample models. The demonstration, probe, training and evaluation queries are disjoint.

\begin{figure}[ht]
\centering
\includegraphics[width=\linewidth]{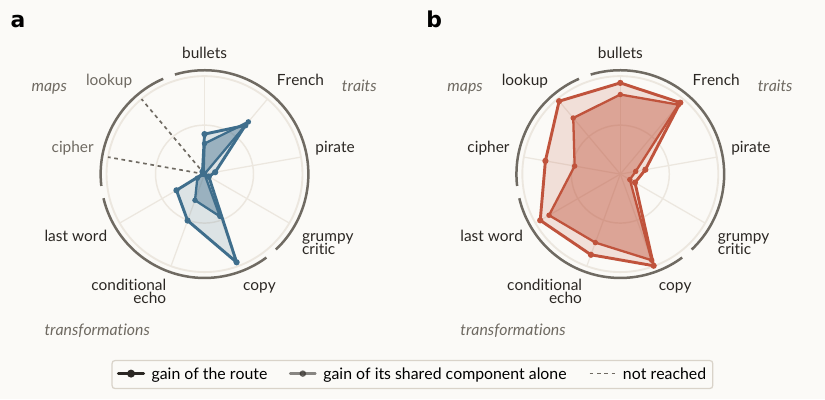}
\caption{\textbf{The two routes behavior by behavior.} Gain over the base model of \textbf{(a)}~the demonstrations and \textbf{(b)}~the LoRA trained in the same runs (outer line), and of each route's shared component alone (inner line), mean over the 18 models. A dashed spoke marks a behavior the route does not reach (gain below $0.05$).}
\label{fig:icl-radar}
\end{figure}
\textbf{Results.} \Cref{fig:icl-radar} compares the two routes behavior by behavior. The demonstrations gain $0.30$ on the traits, $0.60$ on the transformations and $0.02$ on the maps, where the LoRA gains $0.57$, $0.94$ and $0.87$, and the shift adapter $0.57$, $0.91$ and $0.68$.
With only their shared component, the demonstrations keep $64\%$ [$47$, $92$] of their gain on the traits and $46\%$ [$40$, $52$] on the transformations, where the LoRA keeps $87\%$ [$82$, $91$] and $89\%$ [$82$, $95$]. On the maps the LoRA keeps $70\%$ [$59$, $79$], and the demonstrations gain $0.02$, too little for their fraction to mean anything ($33\%$ [$3$, $84$]).
Removing the shared component of the demonstrations keeps none of their gain: $-25\%$ [$-43$, $-11$] on the traits and $-2\%$ [$-2$, $-1$] on the transformations. Removing a random vector of the same norm keeps $15\%$ [$-4$, $34$] on the traits and $14\%$ [$3$, $27$] on the transformations, so on both the removal is specific to the shared component. Removing the shared component of the LoRA keeps $43\%$ [$31$, $55$] on the traits and $62\%$ [$50$, $73$] on the transformations.

\Cref{tab:icl-models} gives the values per model. Where the demonstrations gain little on the traits ($0.00$ on Qwen2.5-0.5B, $0.04$ on Llama-3.2-1B and Qwen2.5-3B), the fraction is large and says little. Mistral-7B is the model whose LoRA keeps least of its gain on the transformations with only its shared component ($35\%$): in three of its 9 transformation runs, the shared component alone sends over $90\%$ of the answers into one repeated token.

\begin{table}[ht]
\centering\small
\caption{\textbf{The two routes per model.} Gain over the base model (mean over the model's runs) and the fraction of that gain its shared component alone keeps (ratio of sums over every run), for the demonstrations and for the LoRA trained in the same runs, on the 4 traits and the 3 transformations of the setup.}
\label{tab:icl-models}
\begin{tabular}{lcccccccc}
\toprule
& \multicolumn{4}{c}{traits} & \multicolumn{4}{c}{transformations} \\
\cmidrule(lr){2-5}\cmidrule(lr){6-9}
& \multicolumn{2}{c}{demonstrations} & \multicolumn{2}{c}{LoRA} & \multicolumn{2}{c}{demonstrations} & \multicolumn{2}{c}{LoRA} \\
model & gain & kept & gain & kept & gain & kept & gain & kept \\
\midrule
Llama-3.1-8B (base) & $0.56$ & $0.16$ & $0.66$ & $0.88$ & $0.62$ & $0.55$ & $0.95$ & $0.96$ \\
Llama-3.1-8B & $0.23$ & $1.02$ & $0.65$ & $0.83$ & $0.69$ & $0.43$ & $0.94$ & $0.91$ \\
Tulu-3-8B-SFT & $0.47$ & $0.67$ & $0.50$ & $0.77$ & $0.67$ & $0.44$ & $0.94$ & $0.68$ \\
Llama-3.2-1B & $0.04$ & $2.24$ & $0.61$ & $0.68$ & $0.52$ & $0.35$ & $0.93$ & $0.80$ \\
Llama-3.2-3B & $0.31$ & $0.30$ & $0.65$ & $0.88$ & $0.62$ & $0.38$ & $0.93$ & $0.92$ \\
Mistral-7B & $0.41$ & $0.60$ & $0.57$ & $0.73$ & $0.60$ & $0.59$ & $0.93$ & $0.35$ \\
Qwen2.5-0.5B & $0.00$ & $12.48$ & $0.34$ & $0.97$ & $0.44$ & $0.32$ & $0.92$ & $0.92$ \\
Qwen2.5-1.5B & $0.14$ & $1.15$ & $0.45$ & $1.01$ & $0.55$ & $0.31$ & $0.91$ & $0.97$ \\
Qwen2.5-3B & $0.04$ & $8.41$ & $0.54$ & $0.97$ & $0.58$ & $0.28$ & $0.93$ & $0.96$ \\
Qwen2.5-7B (base) & $0.16$ & $-0.39$ & $0.58$ & $0.87$ & $0.51$ & $0.44$ & $0.92$ & $0.97$ \\
Qwen2.5-7B & $0.11$ & $1.25$ & $0.56$ & $0.96$ & $0.61$ & $0.22$ & $0.92$ & $0.99$ \\
Qwen3-4B & $0.33$ & $0.41$ & $0.56$ & $0.77$ & $0.63$ & $0.37$ & $0.96$ & $0.94$ \\
Qwen3.5-4B & $0.57$ & $0.49$ & $0.63$ & $0.94$ & $0.63$ & $0.62$ & $0.97$ & $0.96$ \\
Qwen3.5-9B & $0.52$ & $0.65$ & $0.63$ & $0.96$ & $0.65$ & $0.74$ & $0.97$ & $0.94$ \\
Qwen3.6-27B & $0.56$ & $0.66$ & $0.61$ & $0.89$ & $0.70$ & $0.57$ & $0.96$ & $0.98$ \\
SmolLM2-1.7B & $0.27$ & $0.94$ & $0.55$ & $0.79$ & $0.53$ & $0.62$ & $0.93$ & $0.95$ \\
Gemma-2-2B (base) & $0.52$ & $0.36$ & $0.60$ & $0.91$ & $0.60$ & $0.48$ & $0.95$ & $0.94$ \\
Gemma-2-2B & $0.09$ & $1.89$ & $0.59$ & $0.82$ & $0.62$ & $0.51$ & $0.94$ & $0.95$ \\
\bottomrule
\end{tabular}
\end{table}

\textbf{Pirate and grumpy critic.} On these two traits the score and the generations disagree. The demonstrations gain $0.11$ on pirate and $0.05$ on grumpy critic, and with only their shared component they fall below the base model on both (gains of $-0.14$ and $-0.07$), which lowers the pooled trait numbers above. We read the saved generations of 10 instruction-tuned models. Under the demonstrations, the generations of Mistral-7B, SmolLM2-1.7B, Gemma-2-2B and Qwen3-4B carry a pirate marker as often as under the shift adapter ($87\%$ to $99\%$ of them, against $98\%$ to $100\%$), yet their F1 gain over the base model is $63\%$ to $106\%$ of the trained one (Mistral $0.42$ against $0.41$ for the shift adapter, SmolLM2 $0.39$ against $0.46$), because the trained routes reproduce the teacher's sentences and the demonstrations write their own. The three Qwen2.5 models carry the marker in $36\%$ to $41\%$ of their generations. Llama-3.2-1B refuses (``I can't help with that'', F1 $0.04$ against $0.20$ for the base model), and Llama-3.2-3B and Llama-3.1-8B continue the demonstration transcript (``I'll answer your questions one by one, matey. Q: \ldots'', markers in $49\%$ of their generations). The short pirate and grumpy critic spokes of \cref{fig:icl-radar} therefore measure how closely the demonstrations reproduce the teacher's answers, and the generations carry the pirate marker in $36\%$ to $99\%$ of cases in nine of these 10 models.

\begin{figure}[ht]
\centering
\includegraphics[width=\linewidth]{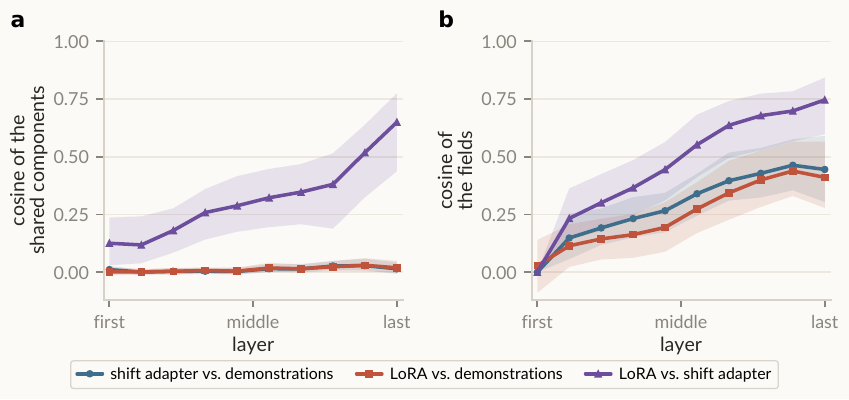}
\caption{\textbf{Shared components and their fields through depth.} Median cosine per layer decile over every run on the 9 behaviors of the setup, quartiles shaded, for 3 pairs of routes: the shift adapter and the demonstrations, the LoRA and the demonstrations, the LoRA and the shift adapter. \textbf{(a)}~Their shared components: the shift adapter's vector per layer, the vector per layer that keeping only the LoRA's shared component adds, and the demonstrations' shared component. \textbf{(b)}~Their fields.}
\label{fig:icl-depth}
\end{figure}
\textbf{Shared components and fields through depth.} \Cref{fig:icl-depth} compares the shared components of the three routes, and their fields, layer by layer. The shift adapter's vector and the shared component of the demonstrations are orthogonal at every layer: the median cosine per layer decile lies between $0.00$ and $0.03$ on the traits and transformations and is at most $0.08$ on the maps, and at most $0.31$ in any run and layer. The shared component of the LoRA, the vector per layer that keeping only it adds, is also orthogonal to that of the demonstrations (median cosine per layer decile at most $0.04$ on the traits and transformations and $0.09$ on the maps, at most $0.30$ in any run and layer). The two trained routes agree more with depth, from a median cosine of $0.15$, $0.17$ and $0.04$ in the first layer decile on the traits, transformations and maps to $0.52$, $0.68$ and $0.84$ in the last.
The fields are not orthogonal. By the ninth layer decile, the fields of the shift adapter and the demonstrations reach a cosine of $0.47$ on the traits and $0.51$ on the transformations, the fields of the shared component of the LoRA and of the demonstrations $0.43$ and $0.53$, and the fields of the two trained routes $0.77$ and $0.66$, reaching $0.84$ and $0.66$ in the last decile. The trained routes add vectors orthogonal to the shared component of the demonstrations, and their fields meet the field of the demonstrations in the late layers.

\textbf{On the tasks.} We run the same two interventions on 4 of the tasks of \cref{sec:parameterizations}, AG News, SST-2, OpenBookQA and HellaSwag, over the 18 models. Each run puts four label-balanced demonstrations before the query, takes their shared component as the mean change at the last token of the input over $300$ probe queries, and chooses the scale $\alpha\in\{0.05,0.1,0.25,0.5,1,2\}$ on a held-out slice of $300$ rows. Every route is then scored on the task's full evaluation split. A rank-8 LoRA trained in the same run on the task's training rows, never on the demonstrations, is scored the same way.

\begin{figure}[ht]
\centering
\includegraphics[width=\linewidth]{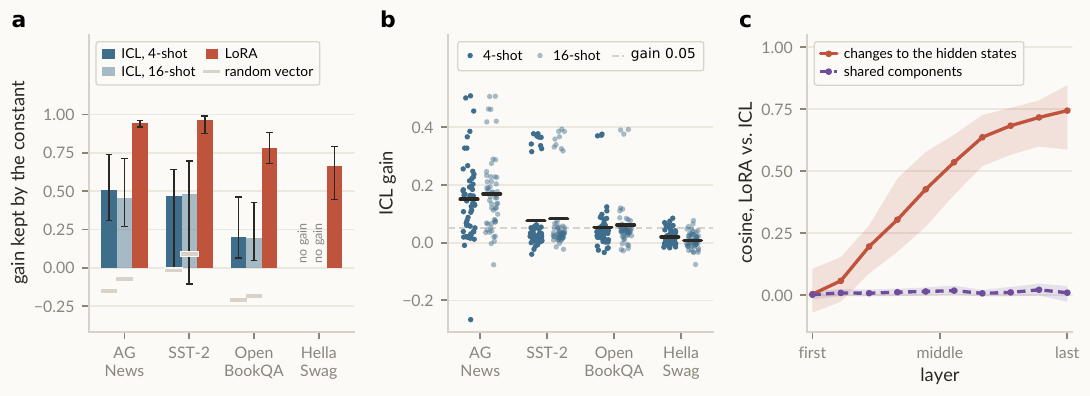}
\caption{\textbf{The two routes on standard tasks.} Color is the route, shade the number of demonstrations: solid for four, pale for sixteen. The LoRA is trained on the run's training rows, so it has one bar. \textbf{(a)}~The fraction of each route's own gain that its shared component alone keeps, per task, as a ratio of sums over every run, with whiskers resampling models, and the tick is a random vector of the same norm in place of the demonstrations' shared component. Where the demonstrations gain less than $0.05$ on average, there is no gain to divide (``no gain''). \textbf{(b)}~Every run's gain, per task, with the mean marked and a line at $0.05$. \textbf{(c)}~Through depth, the cosine between the two routes' shared components and between the changes they make to the hidden states, median per layer decile over all runs with quartile bands.}
\label{fig:icl-bench}
\end{figure}

Four demonstrations change these tasks little: the median gain over the runs is $0.033$ (quartiles $0.011$ to $0.078$), and one run (Llama-3.2-1B on AG News) ends $0.267$ below the bare prompt (\cref{fig:icl-bench}b). The demonstrations gain $0.05$ or more on average only on AG News, SST-2 and OpenBookQA.
With only their shared component, the demonstrations keep $42\%$ [$26$, $58$] of their gain over every run, where the LoRA trained in the same runs keeps $83\%$ [$75$, $89$] of its own, and a random vector of the same norm keeps $-14\%$ [$-36$, $-1$]. On AG News and SST-2 the demonstrations keep $50\%$ and $47\%$, where the LoRA keeps $94\%$ and $96\%$ on the same runs (\cref{fig:icl-bench}a). On OpenBookQA the demonstrations keep $20\%$ where the LoRA keeps $78\%$, and on HellaSwag their mean gain is below $0.05$, too little for a fraction, while the LoRA keeps $66\%$. The fractions of the LoRA on every task are in \cref{app:interventions-bench}.
With 16 demonstrations, the shared component alone keeps $41\%$ [$23$, $58$] and removing it keeps $49\%$ [$34$, $69$], against $42\%$ and $39\%$ with four.

Removing the shared component from the prompt with the demonstrations keeps $39\%$ [$17$, $57$], against $80\%$ [$71$, $87$] for removing a random vector of the same norm, so the direction matters. The scale is chosen for keeping only the shared component and reused for the removal, and in $35$ of the $216$ runs the removal sends the model to one class, which widens the interval. At full strength it does so in every run, as on the behaviors.
Through depth, the shared components of the two routes are orthogonal, with a median cosine per layer decile between $0.00$ and $0.02$ over all runs, while their fields converge, from $0.00$ in the first decile to $0.74$ in the last (quartiles [$0.59$, $0.85$], \cref{fig:icl-bench}c). \Cref{tab:icl-bench-models} gives the values per model.
On behaviors and tasks, the shared component of the demonstrations is orthogonal to LoRA's and keeps less of the gain than LoRA's on the traits, the transformations and the tasks, so in-context learning reaches similar behaviors by a different route from LoRA.

\begin{table}[ht]
\centering\small
\caption{\textbf{The two routes per model on the tasks.} Gain over the base model (mean over the model's runs) and the fraction of that gain its shared component alone keeps (ratio of sums over every run), for the demonstrations and for the LoRA trained in the same runs, over AG News, SST-2, OpenBookQA and HellaSwag. Where a route gains little, its fraction is large or negative and says little.}
\label{tab:icl-bench-models}
\begin{tabular}{l r c c c c}
\toprule
model & runs & \multicolumn{2}{c}{demonstrations} & \multicolumn{2}{c}{LoRA} \\
\cmidrule(lr){3-4}\cmidrule(lr){5-6}
 & & gain & kept & gain & kept \\
\midrule
Gemma-2-2B & $12$ & $+0.14$ & $+0.38$ & $+0.23$ & $+0.92$ \\
Gemma-2-2B (base) & $12$ & $+0.21$ & $+0.36$ & $+0.40$ & $+0.83$ \\
Llama-3.1-8B & $12$ & $+0.07$ & $+0.53$ & $+0.14$ & $+0.90$ \\
Llama-3.1-8B (base) & $12$ & $+0.06$ & $+0.15$ & $+0.18$ & $+0.90$ \\
Llama-3.2-1B & $12$ & $-0.01$ & $-0.69$ & $+0.19$ & $+0.74$ \\
Llama-3.2-3B & $12$ & $+0.06$ & $+0.00$ & $+0.15$ & $+0.78$ \\
Mistral-7B & $12$ & $+0.07$ & $+0.59$ & $+0.15$ & $+0.68$ \\
Qwen2.5-0.5B & $12$ & $+0.08$ & $+1.13$ & $+0.24$ & $+0.79$ \\
Qwen2.5-1.5B & $12$ & $+0.05$ & $+0.91$ & $+0.15$ & $+0.85$ \\
Qwen2.5-3B & $12$ & $+0.04$ & $+0.31$ & $+0.11$ & $+0.86$ \\
Qwen2.5-7B & $12$ & $+0.03$ & $+0.24$ & $+0.08$ & $+0.95$ \\
Qwen2.5-7B (base) & $12$ & $+0.03$ & $+0.37$ & $+0.10$ & $+0.94$ \\
Qwen3-4B & $12$ & $+0.01$ & $+0.22$ & $+0.05$ & $+0.89$ \\
Qwen3.5-4B & $12$ & $+0.15$ & $+0.37$ & $+0.21$ & $+0.97$ \\
Qwen3.5-9B & $12$ & $+0.05$ & $+0.52$ & $+0.09$ & $+0.96$ \\
Qwen3.6-27B & $12$ & $+0.11$ & $-0.25$ & $+0.13$ & $+1.00$ \\
SmolLM2-1.7B & $12$ & $+0.17$ & $+0.68$ & $+0.35$ & $+0.86$ \\
Tulu-3-8B-SFT & $12$ & $+0.03$ & $+0.41$ & $+0.10$ & $-0.02$ \\
\midrule
all & $216$ & $+0.07$ & $+0.42$ & $+0.17$ & $+0.83$ \\
\bottomrule
\end{tabular}
\end{table}

\FloatBarrier

\section{The shift adapter}\label{app:methods}

\subsection{The comparison on the tasks}\label{app:parity}\label{app:parameterization-tests}\label{app:fields}

\begin{wrapfigure}{r}{0.34\linewidth}
\centering
\includegraphics[width=0.8\linewidth]{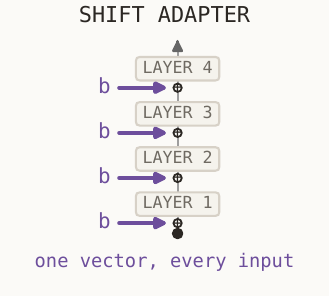}
\caption{\textbf{The shift adapter.} One trained vector $b_\ell$ per layer, added to the hidden state at every position, the same for every input.}
\label{fig:method-shift}
\end{wrapfigure}

We compare the shift adapter (\cref{fig:method-shift}) with LoRAs of rank $1$, $4$, $8$, $16$ and $64$. The shift adapter and the LoRAs are trained on the same teacher pairs, each at the learning rate selected for it (\cref{app:training}), and every task is scored by accuracy on its whole labeled evaluation split. These splits are unbalanced, so the majority-class rate is reported beside every result.

\begin{figure}[ht]
\centering
\includegraphics[width=\textwidth]{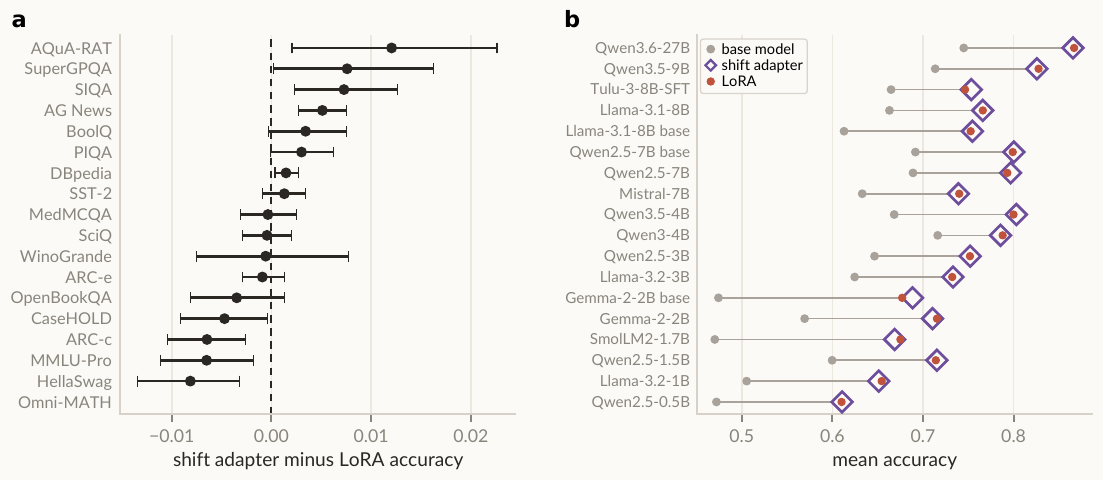}
\caption{\textbf{\Cref{fig:parity} with every task and every model.}
\textbf{(a)}~The paired shift-minus-rank-8 difference per task, with bootstrap $95\%$ intervals over the runs behind each.
\textbf{(b)}~Per model, ordered by scale: mean accuracy of the base model, of the shift adapter and of the rank-8 LoRA over the 17 tasks.}
\label{fig:parity-all}
\end{figure}

\Cref{tab:parity-full} gives every task and every method, each with a bootstrap interval over the combinations of model and seed behind it, and the paired difference between the shift adapter and the rank-8 LoRA with its own interval. Pooled over the 18 models, the shift adapter reaches $0.746$ and the rank-8 LoRA $0.745$, against $0.620$ for the base model, a paired difference of $+0.0006$ [$-0.0007$, $+0.0018$]. Against the other ranks the difference ranges from $+0.0010$ [$-0.0001$, $+0.0021$] for rank $1$ to $-0.0022$ [$-0.0034$, $-0.0011$] for rank $64$. Per model it ranges from $-0.006$ on SmolLM2-1.7B to $+0.011$ on the base Gemma-2-2B, so no single model carries the result. Per task, the shift adapter leads on AQuA-RAT ($+0.012$), SuperGPQA ($+0.008$), SIQA ($+0.007$) and AG News ($+0.005$), and the LoRA leads most on HellaSwag ($-0.008$), MMLU-Pro ($-0.007$) and ARC-Challenge ($-0.006$). \Cref{fig:parity-all} draws every task and every model.
The selected learning rates of the shift adapter span a factor of $1{,}000$ across models and tasks, from $10^{-5}$ to $10^{-2}$, against $50$ for the rank-8 LoRA, which is why each method trains at its own selected rate.

\begin{table}[ht]
\caption{\textbf{Every task and every method.} Generated-answer accuracy on the full labeled evaluation split of the 17 tasks, on the 18 models. Brackets are bootstrap $95\%$ intervals, and \emph{floor} is the majority-class rate. The last column is the paired shift-minus-rank-8 difference.}
\label{tab:parity-full}
\centering
\footnotesize
\setlength{\tabcolsep}{1.0pt}
\begin{tabular}{l r *{6}{c} c}
\toprule
benchmark & floor & shift & r1 & r4 & r8 & r16 & r64 & shift $-$ r8 \\
\midrule
\shortstack[l]{BoolQ\\[-1pt]{\tiny\strut}} & \shortstack{$0.622$\\[-1pt]{\tiny\strut}} & \shortstack{$0.863$\\[-1pt]{\tiny $[0.853, 0.873]$}} & \shortstack{$0.858$\\[-1pt]{\tiny $[0.845, 0.870]$}} & \shortstack{$0.857$\\[-1pt]{\tiny $[0.844, 0.869]$}} & \shortstack{$0.860$\\[-1pt]{\tiny $[0.847, 0.871]$}} & \shortstack{$0.860$\\[-1pt]{\tiny $[0.848, 0.871]$}} & \shortstack{$0.862$\\[-1pt]{\tiny $[0.852, 0.872]$}} & \shortstack{$0.003$\\[-1pt]{\tiny $[-0.000, 0.007]$}} \\
\shortstack[l]{PIQA\\[-1pt]{\tiny\strut}} & \shortstack{$0.505$\\[-1pt]{\tiny\strut}} & \shortstack{$0.828$\\[-1pt]{\tiny $[0.810, 0.845]$}} & \shortstack{$0.823$\\[-1pt]{\tiny $[0.804, 0.841]$}} & \shortstack{$0.823$\\[-1pt]{\tiny $[0.804, 0.841]$}} & \shortstack{$0.825$\\[-1pt]{\tiny $[0.807, 0.842]$}} & \shortstack{$0.826$\\[-1pt]{\tiny $[0.807, 0.844]$}} & \shortstack{$0.828$\\[-1pt]{\tiny $[0.809, 0.845]$}} & \shortstack{$0.003$\\[-1pt]{\tiny $[-0.000, 0.006]$}} \\
\shortstack[l]{SIQA\\[-1pt]{\tiny\strut}} & \shortstack{$0.336$\\[-1pt]{\tiny\strut}} & \shortstack{$0.747$\\[-1pt]{\tiny $[0.733, 0.761]$}} & \shortstack{$0.745$\\[-1pt]{\tiny $[0.729, 0.759]$}} & \shortstack{$0.743$\\[-1pt]{\tiny $[0.727, 0.757]$}} & \shortstack{$0.740$\\[-1pt]{\tiny $[0.726, 0.753]$}} & \shortstack{$0.746$\\[-1pt]{\tiny $[0.730, 0.759]$}} & \shortstack{$0.749$\\[-1pt]{\tiny $[0.734, 0.763]$}} & \shortstack{$0.007$\\[-1pt]{\tiny $[0.002, 0.013]$}} \\
\shortstack[l]{WinoGrande\\[-1pt]{\tiny\strut}} & \shortstack{$0.504$\\[-1pt]{\tiny\strut}} & \shortstack{$0.740$\\[-1pt]{\tiny $[0.716, 0.763]$}} & \shortstack{$0.739$\\[-1pt]{\tiny $[0.716, 0.762]$}} & \shortstack{$0.743$\\[-1pt]{\tiny $[0.718, 0.766]$}} & \shortstack{$0.741$\\[-1pt]{\tiny $[0.717, 0.764]$}} & \shortstack{$0.743$\\[-1pt]{\tiny $[0.719, 0.764]$}} & \shortstack{$0.747$\\[-1pt]{\tiny $[0.723, 0.770]$}} & \shortstack{$-0.001$\\[-1pt]{\tiny $[-0.008, 0.008]$}} \\
\shortstack[l]{ARC-e\\[-1pt]{\tiny\strut}} & \shortstack{$0.268$\\[-1pt]{\tiny\strut}} & \shortstack{$0.909$\\[-1pt]{\tiny $[0.892, 0.926]$}} & \shortstack{$0.910$\\[-1pt]{\tiny $[0.893, 0.927]$}} & \shortstack{$0.911$\\[-1pt]{\tiny $[0.894, 0.927]$}} & \shortstack{$0.910$\\[-1pt]{\tiny $[0.893, 0.928]$}} & \shortstack{$0.913$\\[-1pt]{\tiny $[0.897, 0.930]$}} & \shortstack{$0.913$\\[-1pt]{\tiny $[0.897, 0.929]$}} & \shortstack{$-0.001$\\[-1pt]{\tiny $[-0.003, 0.001]$}} \\
\shortstack[l]{ARC-c\\[-1pt]{\tiny\strut}} & \shortstack{$0.264$\\[-1pt]{\tiny\strut}} & \shortstack{$0.791$\\[-1pt]{\tiny $[0.759, 0.824]$}} & \shortstack{$0.794$\\[-1pt]{\tiny $[0.761, 0.826]$}} & \shortstack{$0.796$\\[-1pt]{\tiny $[0.765, 0.826]$}} & \shortstack{$0.798$\\[-1pt]{\tiny $[0.767, 0.828]$}} & \shortstack{$0.796$\\[-1pt]{\tiny $[0.765, 0.827]$}} & \shortstack{$0.800$\\[-1pt]{\tiny $[0.769, 0.829]$}} & \shortstack{$-0.006$\\[-1pt]{\tiny $[-0.010, -0.003]$}} \\
\shortstack[l]{HellaSwag\\[-1pt]{\tiny\strut}} & \shortstack{$0.257$\\[-1pt]{\tiny\strut}} & \shortstack{$0.833$\\[-1pt]{\tiny $[0.804, 0.860]$}} & \shortstack{$0.839$\\[-1pt]{\tiny $[0.812, 0.864]$}} & \shortstack{$0.844$\\[-1pt]{\tiny $[0.817, 0.869]$}} & \shortstack{$0.841$\\[-1pt]{\tiny $[0.814, 0.866]$}} & \shortstack{$0.842$\\[-1pt]{\tiny $[0.815, 0.867]$}} & \shortstack{$0.837$\\[-1pt]{\tiny $[0.809, 0.864]$}} & \shortstack{$-0.008$\\[-1pt]{\tiny $[-0.013, -0.003]$}} \\
\shortstack[l]{OpenBookQA\\[-1pt]{\tiny\strut}} & \shortstack{$0.276$\\[-1pt]{\tiny\strut}} & \shortstack{$0.847$\\[-1pt]{\tiny $[0.825, 0.869]$}} & \shortstack{$0.854$\\[-1pt]{\tiny $[0.832, 0.874]$}} & \shortstack{$0.854$\\[-1pt]{\tiny $[0.832, 0.875]$}} & \shortstack{$0.851$\\[-1pt]{\tiny $[0.829, 0.872]$}} & \shortstack{$0.841$\\[-1pt]{\tiny $[0.809, 0.869]$}} & \shortstack{$0.855$\\[-1pt]{\tiny $[0.832, 0.877]$}} & \shortstack{$-0.003$\\[-1pt]{\tiny $[-0.008, 0.001]$}} \\
\shortstack[l]{SST-2\\[-1pt]{\tiny\strut}} & \shortstack{$0.509$\\[-1pt]{\tiny\strut}} & \shortstack{$0.947$\\[-1pt]{\tiny $[0.943, 0.950]$}} & \shortstack{$0.944$\\[-1pt]{\tiny $[0.939, 0.949]$}} & \shortstack{$0.943$\\[-1pt]{\tiny $[0.939, 0.947]$}} & \shortstack{$0.946$\\[-1pt]{\tiny $[0.942, 0.949]$}} & \shortstack{$0.947$\\[-1pt]{\tiny $[0.943, 0.951]$}} & \shortstack{$0.950$\\[-1pt]{\tiny $[0.947, 0.953]$}} & \shortstack{$0.001$\\[-1pt]{\tiny $[-0.001, 0.003]$}} \\
\shortstack[l]{DBpedia\\[-1pt]{\tiny\strut}} & \shortstack{$0.071$\\[-1pt]{\tiny\strut}} & \shortstack{$0.986$\\[-1pt]{\tiny $[0.985, 0.987]$}} & \shortstack{$0.985$\\[-1pt]{\tiny $[0.984, 0.985]$}} & \shortstack{$0.984$\\[-1pt]{\tiny $[0.983, 0.985]$}} & \shortstack{$0.984$\\[-1pt]{\tiny $[0.983, 0.985]$}} & \shortstack{$0.984$\\[-1pt]{\tiny $[0.983, 0.985]$}} & \shortstack{$0.984$\\[-1pt]{\tiny $[0.982, 0.985]$}} & \shortstack{$0.001$\\[-1pt]{\tiny $[0.000, 0.003]$}} \\
\shortstack[l]{AG News\\[-1pt]{\tiny\strut}} & \shortstack{$0.250$\\[-1pt]{\tiny\strut}} & \shortstack{$0.901$\\[-1pt]{\tiny $[0.899, 0.902]$}} & \shortstack{$0.897$\\[-1pt]{\tiny $[0.895, 0.898]$}} & \shortstack{$0.894$\\[-1pt]{\tiny $[0.889, 0.898]$}} & \shortstack{$0.896$\\[-1pt]{\tiny $[0.893, 0.898]$}} & \shortstack{$0.898$\\[-1pt]{\tiny $[0.895, 0.900]$}} & \shortstack{$0.899$\\[-1pt]{\tiny $[0.896, 0.902]$}} & \shortstack{$0.005$\\[-1pt]{\tiny $[0.003, 0.008]$}} \\
\shortstack[l]{CaseHOLD\\[-1pt]{\tiny\strut}} & \shortstack{$0.202$\\[-1pt]{\tiny\strut}} & \shortstack{$0.726$\\[-1pt]{\tiny $[0.713, 0.739]$}} & \shortstack{$0.727$\\[-1pt]{\tiny $[0.716, 0.738]$}} & \shortstack{$0.729$\\[-1pt]{\tiny $[0.717, 0.740]$}} & \shortstack{$0.731$\\[-1pt]{\tiny $[0.719, 0.742]$}} & \shortstack{$0.727$\\[-1pt]{\tiny $[0.715, 0.737]$}} & \shortstack{$0.724$\\[-1pt]{\tiny $[0.711, 0.736]$}} & \shortstack{$-0.005$\\[-1pt]{\tiny $[-0.009, -0.000]$}} \\
\shortstack[l]{MedMCQA\\[-1pt]{\tiny\strut}} & \shortstack{$0.322$\\[-1pt]{\tiny\strut}} & \shortstack{$0.542$\\[-1pt]{\tiny $[0.518, 0.567]$}} & \shortstack{$0.543$\\[-1pt]{\tiny $[0.519, 0.568]$}} & \shortstack{$0.543$\\[-1pt]{\tiny $[0.518, 0.568]$}} & \shortstack{$0.543$\\[-1pt]{\tiny $[0.517, 0.568]$}} & \shortstack{$0.544$\\[-1pt]{\tiny $[0.519, 0.568]$}} & \shortstack{$0.542$\\[-1pt]{\tiny $[0.518, 0.567]$}} & \shortstack{$-0.000$\\[-1pt]{\tiny $[-0.003, 0.002]$}} \\
\shortstack[l]{AQuA-RAT\\[-1pt]{\tiny\strut}} & \shortstack{$0.248$\\[-1pt]{\tiny\strut}} & \shortstack{$0.394$\\[-1pt]{\tiny $[0.362, 0.427]$}} & \shortstack{$0.385$\\[-1pt]{\tiny $[0.350, 0.421]$}} & \shortstack{$0.386$\\[-1pt]{\tiny $[0.352, 0.421]$}} & \shortstack{$0.382$\\[-1pt]{\tiny $[0.347, 0.418]$}} & \shortstack{$0.387$\\[-1pt]{\tiny $[0.353, 0.423]$}} & \shortstack{$0.391$\\[-1pt]{\tiny $[0.357, 0.426]$}} & \shortstack{$0.012$\\[-1pt]{\tiny $[0.002, 0.022]$}} \\
\shortstack[l]{SciQ\\[-1pt]{\tiny\strut}} & \shortstack{$0.269$\\[-1pt]{\tiny\strut}} & \shortstack{$0.915$\\[-1pt]{\tiny $[0.904, 0.927]$}} & \shortstack{$0.915$\\[-1pt]{\tiny $[0.904, 0.926]$}} & \shortstack{$0.916$\\[-1pt]{\tiny $[0.905, 0.927]$}} & \shortstack{$0.916$\\[-1pt]{\tiny $[0.904, 0.927]$}} & \shortstack{$0.916$\\[-1pt]{\tiny $[0.905, 0.928]$}} & \shortstack{$0.915$\\[-1pt]{\tiny $[0.903, 0.927]$}} & \shortstack{$-0.000$\\[-1pt]{\tiny $[-0.003, 0.002]$}} \\
\shortstack[l]{MMLU-Pro\\[-1pt]{\tiny\strut}} & \shortstack{$0.112$\\[-1pt]{\tiny\strut}} & \shortstack{$0.425$\\[-1pt]{\tiny $[0.394, 0.455]$}} & \shortstack{$0.426$\\[-1pt]{\tiny $[0.393, 0.459]$}} & \shortstack{$0.431$\\[-1pt]{\tiny $[0.399, 0.463]$}} & \shortstack{$0.431$\\[-1pt]{\tiny $[0.399, 0.463]$}} & \shortstack{$0.426$\\[-1pt]{\tiny $[0.394, 0.458]$}} & \shortstack{$0.433$\\[-1pt]{\tiny $[0.402, 0.464]$}} & \shortstack{$-0.007$\\[-1pt]{\tiny $[-0.011, -0.002]$}} \\
\shortstack[l]{SuperGPQA\\[-1pt]{\tiny\strut}} & \shortstack{$0.105$\\[-1pt]{\tiny\strut}} & \shortstack{$0.278$\\[-1pt]{\tiny $[0.257, 0.299]$}} & \shortstack{$0.274$\\[-1pt]{\tiny $[0.251, 0.297]$}} & \shortstack{$0.277$\\[-1pt]{\tiny $[0.253, 0.301]$}} & \shortstack{$0.270$\\[-1pt]{\tiny $[0.247, 0.294]$}} & \shortstack{$0.277$\\[-1pt]{\tiny $[0.254, 0.301]$}} & \shortstack{$0.283$\\[-1pt]{\tiny $[0.260, 0.306]$}} & \shortstack{$0.008$\\[-1pt]{\tiny $[0.000, 0.016]$}} \\
\bottomrule
\end{tabular}
\end{table}

\subsection{Full fine-tuning on the same tasks}\label{app:parity-fullft}

Full fine-tuning updates every weight. We train it with AdamW in bfloat16 for $500$ steps, the budget of the LoRA, at a learning rate selected per model and task from the 6 rates of \cref{app:training}, on the 17 tasks and 18 models, scored on the same evaluation splits. The shift adapter is retrained in the same run at its selected rate, so every difference is paired within a run. The retrained shift adapter reproduces the shift adapter of \cref{tab:parity-full} to a mean absolute difference of $0.004$.

Pooled over the 18 models, full fine-tuning reaches $0.718$, the shift adapter and the rank-8 LoRA $0.745$, against $0.620$ for the base model, a paired difference between the shift adapter and full fine-tuning of $+0.028$ [$+0.025$, $+0.030$] (\cref{tab:parity-fullft}). The shift adapter is at or above full fine-tuning on every task except ARC-Challenge ($-0.002$), and leads most on WinoGrande ($+0.066$), CaseHOLD ($+0.060$) and HellaSwag ($+0.057$). Per model its lead ranges from $+0.011$ on Qwen3.6-27B to $+0.069$ on the base Gemma-2-2B.
Over $900$ of these runs, the field of full fine-tuning has a cosine of $0.90$ to its shared component and a field share of $0.81$, against $0.91$ and $0.83$ for the rank-8 LoRA, paired differences of $-0.009$ and $-0.015$ (\cref{fig:shared-all}c). Full fine-tuning moves every input in nearly the same direction, as LoRA does.

\begin{table}[ht]
\caption{\textbf{Full fine-tuning against the shift adapter on every task.} Accuracy on the evaluation split of the 17 tasks, on the 18 models. The last column is the paired shift-minus-full-fine-tuning difference with its bootstrap $95\%$ interval.}
\label{tab:parity-fullft}
\centering
\footnotesize
\begin{tabular}{l r r c c c c}
\toprule
benchmark & \shortstack{majority-\\class rate} & base & \shortstack{shift\\adapter} & \shortstack{rank-8\\LoRA} & full FT & shift $-$ full FT \\
\midrule
BoolQ & $0.622$ & $0.766$ & $0.862$ & $0.860$ & $0.838$ & $+0.024$ {\tiny $[+0.020, +0.028]$} \\
PIQA & $0.505$ & $0.746$ & $0.829$ & $0.825$ & $0.809$ & $+0.020$ {\tiny $[+0.015, +0.025]$} \\
SIQA & $0.336$ & $0.686$ & $0.748$ & $0.740$ & $0.722$ & $+0.026$ {\tiny $[+0.019, +0.034]$} \\
WinoGrande & $0.504$ & $0.594$ & $0.739$ & $0.741$ & $0.673$ & $+0.066$ {\tiny $[+0.052, +0.081]$} \\
ARC-e & $0.268$ & $0.888$ & $0.910$ & $0.910$ & $0.904$ & $+0.006$ {\tiny $[+0.003, +0.009]$} \\
ARC-c & $0.264$ & $0.770$ & $0.792$ & $0.798$ & $0.794$ & $-0.002$ {\tiny $[-0.006, +0.002]$} \\
HellaSwag & $0.257$ & $0.632$ & $0.832$ & $0.841$ & $0.775$ & $+0.057$ {\tiny $[+0.045, +0.071]$} \\
OpenBookQA & $0.276$ & $0.732$ & $0.847$ & $0.851$ & $0.817$ & $+0.030$ {\tiny $[+0.022, +0.038]$} \\
SST-2 & $0.509$ & $0.857$ & $0.946$ & $0.946$ & $0.937$ & $+0.009$ {\tiny $[+0.006, +0.012]$} \\
DBpedia & $0.071$ & $0.511$ & $0.985$ & $0.984$ & $0.984$ & $+0.002$ {\tiny $[+0.000, +0.003]$} \\
AG News & $0.250$ & $0.638$ & $0.900$ & $0.896$ & $0.888$ & $+0.013$ {\tiny $[+0.009, +0.016]$} \\
CaseHOLD & $0.202$ & $0.506$ & $0.726$ & $0.731$ & $0.665$ & $+0.060$ {\tiny $[+0.054, +0.067]$} \\
MedMCQA & $0.322$ & $0.508$ & $0.543$ & $0.543$ & $0.529$ & $+0.014$ {\tiny $[+0.011, +0.018]$} \\
AQuA-RAT & $0.248$ & $0.297$ & $0.392$ & $0.382$ & $0.348$ & $+0.044$ {\tiny $[+0.034, +0.054]$} \\
SciQ & $0.269$ & $0.893$ & $0.915$ & $0.916$ & $0.904$ & $+0.011$ {\tiny $[+0.007, +0.015]$} \\
MMLU-Pro & $0.112$ & $0.308$ & $0.423$ & $0.431$ & $0.376$ & $+0.047$ {\tiny $[+0.038, +0.057]$} \\
SuperGPQA & $0.105$ & $0.205$ & $0.277$ & $0.270$ & $0.236$ & $+0.041$ {\tiny $[+0.035, +0.047]$} \\
\midrule
all & & $0.620$ & $0.745$ & $0.745$ & $0.718$ & $+0.028$ {\tiny $[+0.025, +0.030]$} \\
\bottomrule
\end{tabular}
\end{table}

\subsection{The comparison before instruction tuning}\label{app:parity-base}

We repeat the comparison on the 3 models that never received instruction tuning (\cref{app:base-models}), on the 17 tasks. The shift adapter reaches $0.748$ and the rank-8 LoRA $0.743$, against $0.592$ for the base model, a paired difference of $+0.005$ [$+0.002$, $+0.008$], and the shift adapter is at or above every rank from $1$ to $64$. Full fine-tuning on the same models and tasks reaches $0.707$, below the shift adapter by $+0.041$ [$+0.034$, $+0.048$]. The parity with LoRA holds before instruction tuning, as the reliance on the shared component does (\cref{app:base-models}).

\subsection{The shift adapter and the LoRA push in the same direction}\label{app:agreement}

The shift adapter and the LoRA are trained independently on the same teacher pairs, so we ask whether they push the inputs in the same direction. On the 17 tasks and 18 models, with both trained in the same run, the shared component of the LoRA aligns with the shift adapter at a cosine of $0.64$ [$0.60$, $0.68$], against $0.06$ [$0.04$, $0.08$] for a rank-8 LoRA with random weights and no training, measured in the same runs. On the 9 behaviors and 18 models, the two align at $0.53$ [$0.50$, $0.55$], against $0.07$ [$0.05$, $0.11$] for the random LoRA (\cref{fig:alignment}). There, each input's field points along its adapter's shared component at a mean cosine of $0.85$ [$0.82$, $0.87$] for the shift adapter and $0.84$ [$0.80$, $0.87$] for the LoRA, against $0.66$ [$0.61$, $0.71$] for the random LoRA. Weighting each layer by how far the two adapters move it raises the agreement between the shift adapter and the LoRA to $0.69$, while the random LoRA stays at $0.08$ (\cref{app:field-stats}). A LoRA that can vary what it adds with every input pushes along the direction the shift adapter learns.

\FloatBarrier
\subsection{\update{What a shift adapter can store}}\label{app:express}

With only its shared component, LoRA keeps the least of its gain on the maps, whose rule the model can only learn during adaptation (\cref{app:controls}). A shift adapter trained directly on a map stores that content: it learns both maps, holds about $200$ entries of a table, and splits into a shared component that performs the lookup and \update{stored content} that holds the entries. \update{A map separates capacity from content. The base model holds the capacity to look an entry up and answer with it, but the entries of the table are new content, which the shift adapter has to store. Trained on few items, a shift adapter therefore answers a map on the entries it saw and not on the others, as stored content predicts, while on the behaviors it reaches most of its gain on instructions never seen in training (\cref{app:few-items}).}

\begin{rfigure}[h]
\centering
\updatebox{\includegraphics[width=\dimexpr\textwidth-8.4pt\relax]{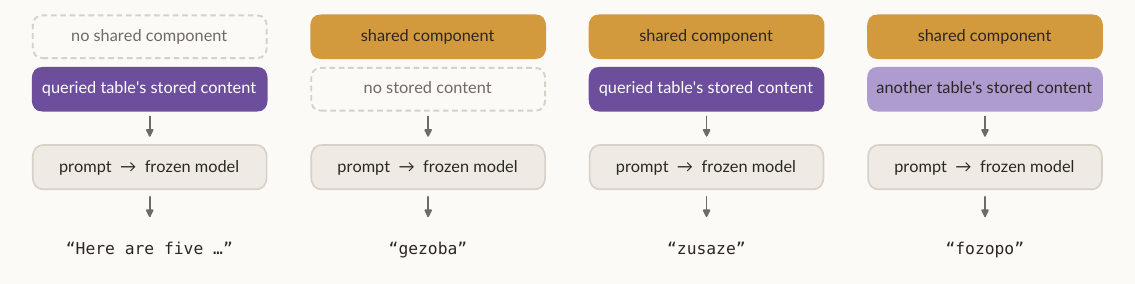}}
\caption{\update{\textbf{The four combinations of shared component and stored content, and the answer each produces.} The shared component (the training estimate) and \update{the stored content} are each added to the frozen model or left out, and the model answers a held-out prompt. With the stored content alone it writes ordinary text, with the shared component alone it invents an entry, and with both it returns an entry of a table. Each quoted answer is the first generated answer of Llama-3.2-3B (seed 0) under that combination.}}
\label{fig:invoke-schematic}
\end{rfigure}

\textbf{A shift adapter learns the maps.} We train a shift adapter directly on the cipher and on a lookup table whose entries are assigned at random, on 5 models from 4 families, and score its held-out generations by exact match. Each map has seven possible answers, so we read the score against the majority-class rate, the score of always giving the most common answer. The shift adapter exceeds that rate on both maps in every family, by $0.08$ to $0.38$, with its answers spread over three to five of the seven, at the level of the LoRA. As a control, a third map, the number of words of the input modulo three, stays at or below the majority-class rate in every family. A shift adapter can therefore store content that pretraining never showed the model. The lower fraction on the maps in \cref{app:controls} comes from the trained LoRA, which puts part of the map in its input-specific component, and a shift adapter trained directly on the map does without that component.

\begin{figure}[t]
\centering
\includegraphics[width=\linewidth]{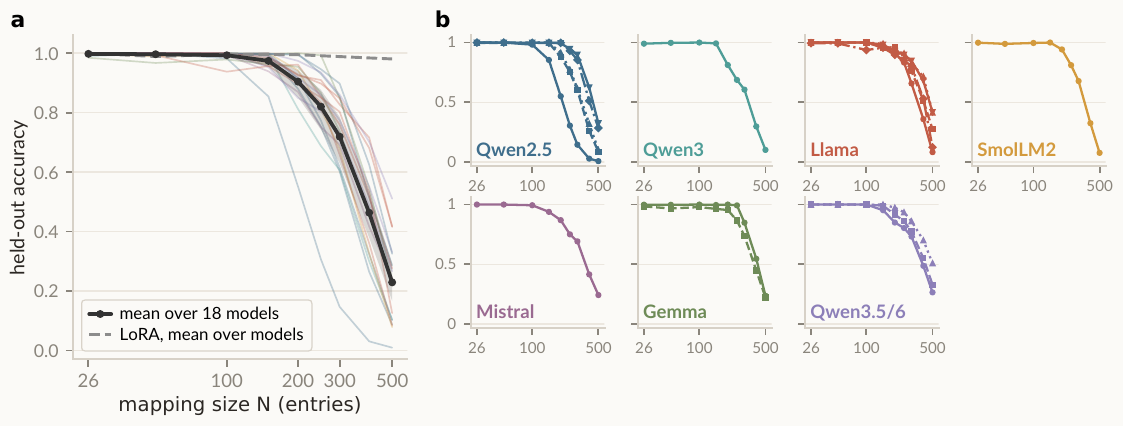}
\caption{\textbf{An actualizing state carries \update{a finite amount of stored content}.} \textbf{(a)}~Held-out lookup accuracy against the size of the table: the mean over the 18 models with a $95\%$ interval over models, each model faint behind it in its family's color, and the rank-8 LoRA mean dashed. \textbf{(b)}~The same per family, one line per model.}
\label{fig:payload-main}
\end{figure}
\textbf{How much fits.} We train a shift adapter on key--value tables of $100$ to $500$ entries and count the entries it answers on held-out queries. The count rises with the table, $99$, $146$, $181$, then $205$ at $250$ entries and $216$ at $300$, and then falls, to $185$ at $400$ entries and $115$ at $500$ (\cref{fig:payload-main}). At $200$ entries the models answer $0.55$ to $1.00$ of the table. A shift adapter therefore holds about $200$ entries, and a larger table lowers the count. The field does not change at the limit: on the 5 families measured, from $100$ to $500$ entries, its magnitude ($728$, $824$, $729$), the share of its energy in the shared component ($0.43$, $0.46$, $0.45$) and the effective rank of its input-specific component ($59$, $54$, $55$) stay the same.
Rank does not move the limit: a rank-1 LoRA holds 500 entries as well as a rank-64 one ($0.950$ against $0.949$). With only its shared component, a trained LoRA keeps $14\%$ of its gain at 500 entries, against $76\%$ at $100$, so past the shift adapter's limit the content sits in the LoRA's input-specific component.
Documents exceed the limit as well, and there the shift adapter stays close to the LoRA. On SQuAD a shift adapter recovers $56\%$ [$53$, $59$] of the gain of reading the document in context, against $61\%$ [$59$, $64$] for the LoRA, and on CoQA $13\%$ [$10$, $16$] against $17\%$ [$14$, $21$] ($30$ held-out documents each, 5 families).

\begin{figure}[t]
\centering
\updatebox{\includegraphics[width=\dimexpr\linewidth-8.4pt\relax]{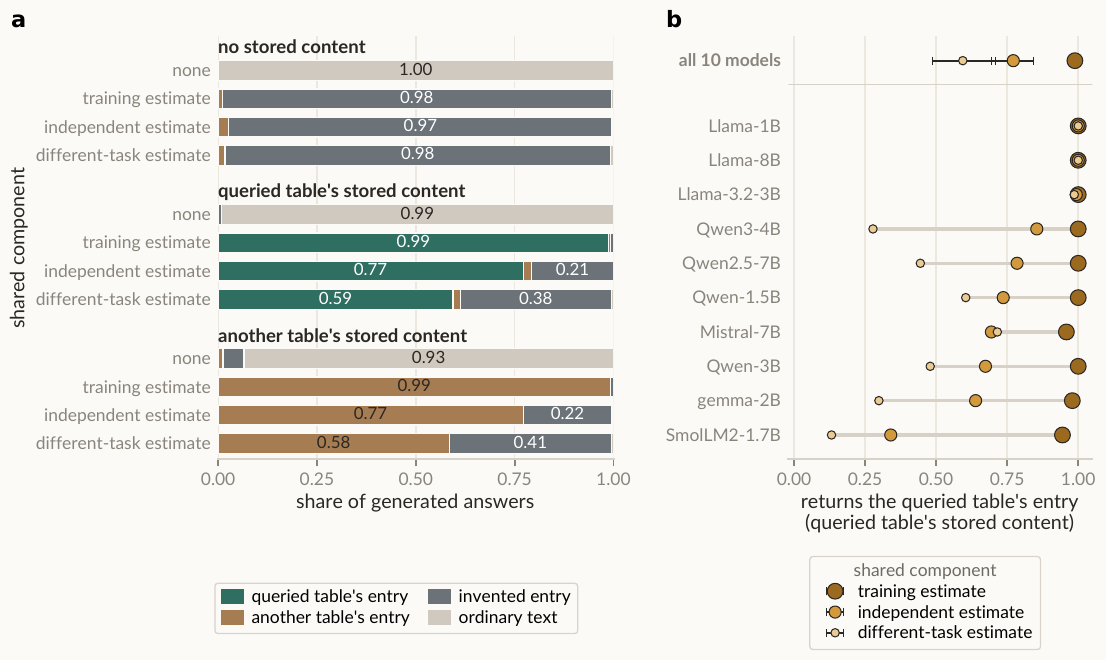}}
\caption{\update{\textbf{Each bar splits the generated answers into four kinds, pooled over 10 models with 3 seeds each and 4 tables of stored content.} The four kinds are the queried table's entry, another table's entry (any other table's real entry), an invented entry, and ordinary text. \textbf{(a)}~Bars are grouped by the stored content (none, the queried table's, another table's), and within a group by the source of the shared component: none, the training estimate (the mean of the $24$ shift adapters under which the stored content was trained), an independent estimate from $24$ other tables, or an estimate from a different task. \textbf{(b)}~The rate of returning the queried table's entry with the queried table's stored content, per model (mean over 3 seeds), under each source of the shared component. The top row pools the 10 models, with bootstrap $95\%$ intervals over the 30 model-seed runs.}}
\label{fig:invoke}
\end{figure}
\textbf{The shared component invokes the lookup, and \update{the stored content} supplies the entries.} On the lookup, one trained shift adapter holds two things: the operation of looking an entry up, and the entries of one table. We separate them by construction. The shared component is estimated as the mean of $24$ shift adapters, each trained on a different table over one set of $50$ keys. With that estimate frozen in the model, a new per-layer vector, the \emph{\update{stored content}}, is trained on the entries of a table outside the set. We then combine shared components and \update{stored content} from different sources and sort each generated answer into four kinds: the queried table's entry, another table's entry, an invented entry, and ordinary text (10 models, 4 \update{tables of stored content}, $48$ held-out prompts each). \update{One generated answer for each combination is shown in \cref{fig:invoke-schematic}.}
The \update{stored content} alone leaves the model's behavior unchanged. The shared component alone makes the model look an entry up with nothing to read, so it invents one ($0.98$ [$0.97$, $0.99$] of answers). The shared component with another table's \update{stored content} returns that table's entries ($0.99$ [$0.99$, $1.00$]). A shared component estimated from $24$ other tables still reads the \update{stored content} at $0.77$ [$0.70$, $0.84$], from $0.99$ or more on every Llama model to $0.34$ on SmolLM2-1.7B (\cref{fig:invoke}). Under a shared component trained for the reverse lookup, from value to key, the forward \update{stored content} returns $0.000$ [$0.000$, $0.001$] of its entries, while a random direction at the same distance returns $0.76$ [$0.72$, $0.81$], so the \update{stored content} holds the table in the direction the forward lookup reads it. The shared component carries the operation, and the \update{stored content} carries the content it reads.

\FloatBarrier
\subsection{\update{Training on few examples}}\label{app:few-items}

\update{In the experiments above, we saw that a shift adapter can store about $200$ table entries (\cref{app:express}). One could therefore ask whether its success on the behaviors comes from simply storing its training examples, \textit{i.e.}, performing parameter-efficient memorization. To test this, we vary only the number of training examples. Across 4 architecture families (Qwen2.5-1.5B, Llama-3.2-3B, SmolLM2-1.7B and Gemma-2-2B) and 3 seeds, we train shift adapters and rank-8 LoRAs on $8$, $32$ or $80$ training examples ($8$, $32$ or $160$ for the maps). For each seed, the training examples are drawn at random from the examples of the behavior. Every adapter is then evaluated on one set of $200$ held-out examples, none of which is ever used for training, so all scores are on examples the adapter has never seen. We report the fraction of the gain at the largest training size that is kept at each smaller size.}

\update{\textbf{Behaviors.} With $32$ training examples, a shift adapter keeps on average $83\%$ of its gain at $80$ training examples on the traits and $92\%$ on the transformations (\cref{fig:rebuttal-few-gain}). With only $8$ training examples, it keeps $59\%$ on the traits and $68\%$ on the transformations. A rank-8 LoRA trained on the same examples keeps $71\%$ and $66\%$. The values per family are given in \cref{tab:rebuttal-few}. All of these scores are on examples never seen in training, so the gain cannot come from storing the training examples. A few training examples thus suffice to actualize most of a behavior that the base model already holds.}

\begin{rfigure}[htbp]
\centering
\updatebox{\includegraphics[width=\dimexpr\textwidth-8.4pt\relax]{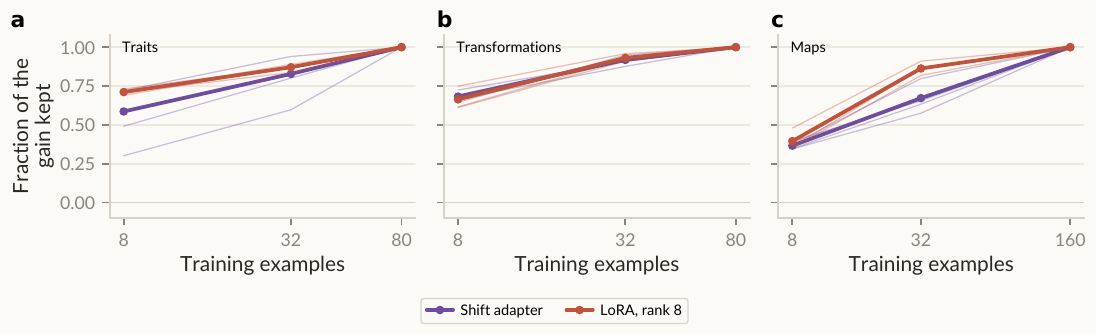}}
\caption{\update{\textbf{Fraction of the gain kept when training on fewer training examples.} The fraction of the gain at the largest training size that a shift adapter (purple) and a rank-8 LoRA (red) keep when trained on fewer training examples, scored on $200$ held-out examples. Thick lines pool the 4 families and 3 seeds as a ratio of sums, thin lines show each family. \textbf{(a)}~Traits. \textbf{(b)}~Transformations. \textbf{(c)}~Maps.}}
\label{fig:rebuttal-few-gain}
\end{rfigure}

\update{\textbf{Lookup tables.} The maps are the opposite case. The base model has the capacity to look an entry up and answer with it, but the entries of a table are new content that it does not hold. The lookup maps the first letter of the input to one of $26$ unrelated words, so a correct answer can only come from a key seen in training. Trained on $8$ training examples, which cover $8$ of the $26$ keys, a shift adapter answers $28\%$ of the evaluation prompts whose key it saw, and $0.1\%$ of the prompts with one of the $18$ keys it never saw (\cref{fig:rebuttal-few-lookup}). A rank-8 LoRA behaves the same way ($48\%$ and $0\%$). The cipher, which substitutes letters, behaves the same way too: it answers $69\%$ on letters seen in training and none on the others, and its fraction in \cref{fig:rebuttal-few-gain}c is high only because the $8$ training examples start with the letters of most evaluation prompts. A shift adapter thus stores exactly the entries it was trained on, while on the behaviors it reaches most of its gain on examples it has never seen. This separates storing content from actualizing a capacity that the base model already holds.}

\begin{rtable}[htbp]
\centering
\caption{\update{\textbf{The shift adapter on few training examples, per family.} The fraction of the gain at $80$ training examples kept with $8$ and $32$ training examples on the traits and the transformations, and the held-out lookup accuracy with $8$ training keys, on the keys seen in training and on the $18$ unseen keys. Pooled over 3 seeds, the last row over the 4 families.}}
\label{tab:rebuttal-few}
\updatebox{\small\begin{tabular}{lcccccc}
\toprule
 & \multicolumn{2}{c}{Traits} & \multicolumn{2}{c}{Transformations} & \multicolumn{2}{c}{Lookup, 8 keys} \\
\cmidrule(lr){2-3}\cmidrule(lr){4-5}\cmidrule(lr){6-7}
Family & 8 & 32 & 8 & 32 & Seen & Unseen \\
\midrule
Qwen2.5-1.5B & $0.30$ & $0.60$ & $0.72$ & $0.91$ & $0.12$ & $0.00$ \\
Llama-3.2-3B & $0.70$ & $0.88$ & $0.65$ & $0.95$ & $0.49$ & $0.00$ \\
SmolLM2-1.7B & $0.49$ & $0.80$ & $0.69$ & $0.93$ & $0.43$ & $0.00$ \\
Gemma-2-2B & $0.72$ & $0.94$ & $0.67$ & $0.88$ & $0.06$ & $0.00$ \\
\midrule
All four & $0.59$ & $0.83$ & $0.68$ & $0.92$ & $0.28$ & $0.00$ \\
\bottomrule
\end{tabular}
}
\end{rtable}

\begin{rfigure}[htbp]
\centering
\updatebox{\includegraphics[width=\dimexpr\textwidth-8.4pt\relax]{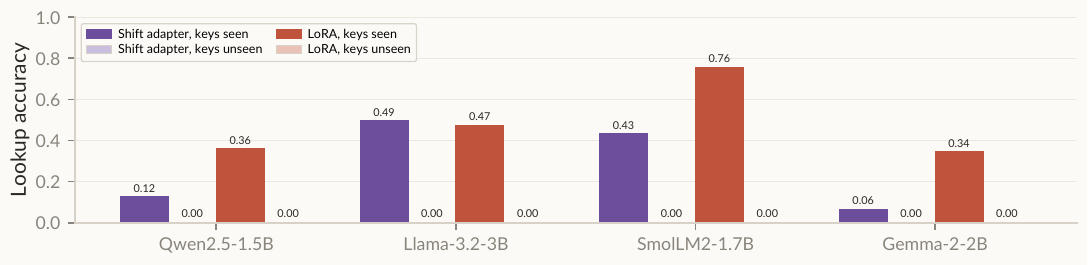}}
\caption{\update{\textbf{Lookup accuracy on keys seen and not seen in training.} Held-out lookup accuracy of a shift adapter and a rank-8 LoRA trained on $8$ of the $26$ keys, on the keys seen in training (dark) and on the $18$ unseen keys (light), per family, pooled over 3 seeds. The base model answers none of the keys.}}
\label{fig:rebuttal-few-lookup}
\end{rfigure}

\update{\textbf{Averaging over example sets.} If a shift adapter stored its training examples, shift adapters trained on different examples would hold different content, and averaging them would cancel it. The three seeds of one behavior and family are trained on three disjoint sets of $8$ training examples. We average their three shift adapters and evaluate the average on examples that none of the three was trained on, together with each single shift adapter in the same run. On the traits, the average keeps $59\%$ of the gain of the full-data shift adapter, compared to $47\%$ for a single one, and it is higher in each of the 4 families. A single shift adapter scaled to the norm of the average keeps $50\%$, so the improvement comes from the direction the three share, not from a change of norm. Averaging shift adapters of different behaviors instead lowers the score below the base model. On the transformations, the average keeps as much as a single shift adapter ($66\%$ and $65\%$). On the lookup, where each seed saw different keys, it keeps less ($7\%$ against $10\%$).}

\update{Shift adapters trained on disjoint examples therefore find the same direction, and averaging them removes what is specific to each set of examples and strengthens the behavior. What a shift adapter learns from a few training examples is the direction that actualizes the behavior, not the examples themselves.}

\subsection{\update{The shift adapter on the other tasks}}\label{app:off-task}

\update{The comparisons above evaluate each shift adapter on the task it is trained on. If a shift adapter improved the model in general, it would also raise the accuracy on the other tasks. We therefore evaluate every shift adapter on all 17 tasks, on the 18 models, and compare it to a random vector with the same norm at every layer. On its own task, the shift adapter raises accuracy by $0.124$ [$0.108$, $0.142$] on average. On the 16 other tasks for which the shift adapter was not trained, accuracy changes by $-0.004$ [$-0.011$, $+0.002$] with the shift adapter and by $-0.004$ [$-0.006$, $-0.001$] with the random vector. The gain of a shift adapter is thus specific to the task it is trained on, and on average performance on other tasks remains at the base model's accuracy.}

\section{The structure of actualization}\label{app:structure}

\subsection{The principle at each checkpoint of OLMo-2-7B}\label{app:development}

\begin{table}[ht]
\caption{\textbf{The principle at each checkpoint of OLMo-2-7B.} Mean accuracy over the 17 tasks: the majority class, the base model with no adaptation, the shift adapter and the rank-8 LoRA, and the paired shift-minus-rank-8 difference. \emph{Not formed} counts the runs of $51$ that neither method lifts more than $2\%$ above the majority class, since the potentiality for the task has not yet formed at that checkpoint. The last two columns are the fraction of the rank-8 LoRA's gain kept with only its shared component and with that component removed, pooled over every run. At $1$B and $9$B tokens the LoRA gains little on most tasks, so these two fractions are wide. Brackets are $95\%$ intervals clustered by task.}
\label{tab:ladder}
\centering
\scriptsize
\setlength{\tabcolsep}{3.5pt}
\begin{tabular}{l r r r r r r r r}
\toprule
checkpoint & majority & base & shift & rank-8 & shift $-$ rank-8 & not formed & kept, shared only & kept, shared removed \\
\midrule
\multicolumn{9}{l}{\emph{stage 1}} \\
1B & $0.30$ & $0.28$ & $0.30$ & $0.32$ & $-0.022$ {\tiny $[-0.062, +0.003]$} & 44/51 & $0.91$ {\tiny $[0.46, 2.20]$} & $0.04$ {\tiny $[-0.23, 0.20]$} \\
9B & $0.30$ & $0.34$ & $0.39$ & $0.40$ & $-0.004$ {\tiny $[-0.015, +0.005]$} & 42/51 & $0.70$ {\tiny $[-1.82, 2.50]$} & $0.31$ {\tiny $[-0.31, 0.70]$} \\
105B & $0.30$ & $0.33$ & $0.45$ & $0.45$ & $+0.002$ {\tiny $[-0.019, +0.020]$} & 28/51 & $0.70$ {\tiny $[0.20, 0.94]$} & $0.25$ {\tiny $[0.06, 0.43]$} \\
449B & $0.30$ & $0.47$ & $0.65$ & $0.65$ & $+0.000$ {\tiny $[-0.015, +0.018]$} & 3/51 & $0.81$ {\tiny $[0.70, 0.90]$} & $0.41$ {\tiny $[0.28, 0.53]$} \\
3,896B & $0.30$ & $0.56$ & $0.72$ & $0.72$ & $-0.007$ {\tiny $[-0.015, +0.001]$} & 0/51 & $0.80$ {\tiny $[0.72, 0.87]$} & $0.47$ {\tiny $[0.20, 0.71]$} \\
\addlinespace
\multicolumn{9}{l}{\emph{stage 2}} \\
5B & $0.30$ & $0.59$ & $0.72$ & $0.73$ & $-0.005$ {\tiny $[-0.012, +0.001]$} & 1/51 & $0.85$ {\tiny $[0.77, 0.91]$} & $0.70$ {\tiny $[0.56, 0.82]$} \\
13B & $0.30$ & $0.59$ & $0.73$ & $0.73$ & $-0.000$ {\tiny $[-0.006, +0.006]$} & 0/51 & $0.93$ {\tiny $[0.88, 0.96]$} & $0.51$ {\tiny $[0.39, 0.67]$} \\
21B & $0.30$ & $0.61$ & $0.73$ & $0.73$ & $-0.001$ {\tiny $[-0.009, +0.007]$} & 0/51 & $0.90$ {\tiny $[0.86, 0.94]$} & $0.60$ {\tiny $[0.35, 0.77]$} \\
30B & $0.30$ & $0.61$ & $0.73$ & $0.74$ & $-0.004$ {\tiny $[-0.010, +0.002]$} & 0/51 & $0.87$ {\tiny $[0.78, 0.93]$} & $0.63$ {\tiny $[0.52, 0.76]$} \\
38B & $0.30$ & $0.61$ & $0.73$ & $0.74$ & $-0.005$ {\tiny $[-0.012, +0.003]$} & 0/51 & $0.88$ {\tiny $[0.79, 0.94]$} & $0.57$ {\tiny $[0.40, 0.72]$} \\
50B & $0.30$ & $0.61$ & $0.73$ & $0.73$ & $-0.001$ {\tiny $[-0.009, +0.007]$} & 0/51 & $0.91$ {\tiny $[0.86, 0.96]$} & $0.51$ {\tiny $[0.29, 0.70]$} \\
\addlinespace
\multicolumn{9}{l}{\emph{released}} \\
base & $0.30$ & $0.62$ & $0.74$ & $0.74$ & $-0.003$ {\tiny $[-0.012, +0.008]$} & 0/51 & $0.85$ {\tiny $[0.72, 0.92]$} & $0.61$ {\tiny $[0.45, 0.74]$} \\
SFT & $0.30$ & $0.63$ & $0.73$ & $0.74$ & $-0.008$ {\tiny $[-0.014, -0.002]$} & 0/51 & $0.81$ {\tiny $[0.73, 0.91]$} & $0.61$ {\tiny $[0.43, 0.75]$} \\
DPO & $0.30$ & $0.63$ & $0.73$ & $0.73$ & $-0.001$ {\tiny $[-0.007, +0.005]$} & 0/51 & $0.88$ {\tiny $[0.81, 0.94]$} & $0.59$ {\tiny $[0.41, 0.76]$} \\
instruct & $0.30$ & $0.63$ & $0.73$ & $0.73$ & $-0.002$ {\tiny $[-0.008, +0.004]$} & 0/51 & $0.87$ {\tiny $[0.81, 0.94]$} & $0.64$ {\tiny $[0.48, 0.78]$} \\
\bottomrule
\end{tabular}
\end{table}

\begin{figure}[t]
\centering
\includegraphics[width=\textwidth]{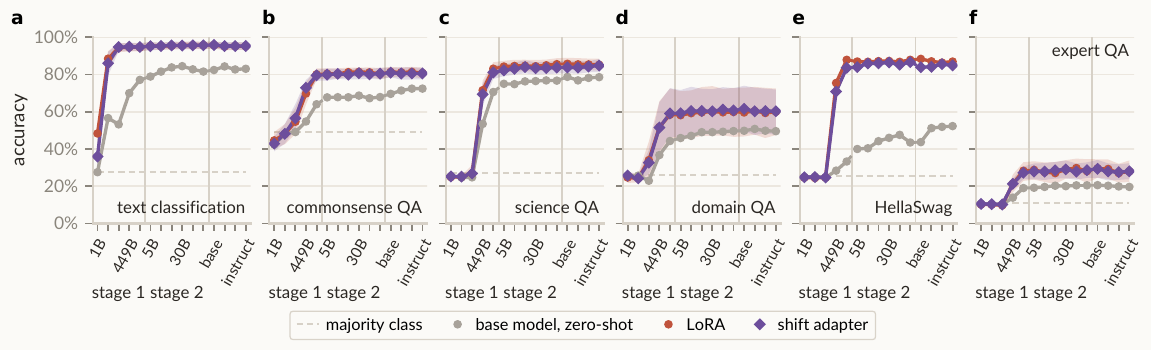}
\caption{\textbf{Accuracy along the training of OLMo-2-7B, by task type.} OLMo-2-7B at $15$ checkpoints: \textbf{(a)}~the three text classifications, \textbf{(b)}~the four commonsense question-answering tasks, \textbf{(c)}~the three science question-answering tasks, \textbf{(d)}~the four domain tasks, \textbf{(e)}~HellaSwag and \textbf{(f)}~the two expert question-answering tasks. Each panel draws the shift adapter (purple diamonds), the LoRA (red), the base model with no adaptation (grey), and the majority class (dashed line), mean over the type's tasks and seeds with $95\%$ intervals over those runs.}
\label{fig:ladder-groups}
\end{figure}

To see when the two parts of the principle appear during training, we repeat the comparison of \cref{app:parity} and the interventions of \cref{app:interventions} along the training of OLMo-2-7B, which AI2 publishes with more than $900$ intermediate checkpoints. We use $15$: five of stage-one pretraining on Dolma, at $1$B, $9$B, $105$B, $449$B and $3{,}896$B tokens, six of the $50$B-token stage two that follows, at $5$B, $13$B, $21$B, $30$B, $38$B and $50$B tokens into it, from the first of the 3 stage-two runs whose average is the released base, and the released base, supervised, preference-tuned and instruct models.
Each checkpoint follows the protocol of \cref{app:training} on the 17 tasks: for each task and method a rate is selected from six on a validation slice of the training split at seed $0$, and every seed is trained on the whole training split at that rate and scored on the full evaluation split. The accuracy of the base model is the zero-shot accuracy of the checkpoint on the same split, read beside the split's majority class. A run where neither method scores more than $0.02$ above the majority class counts as a run in which the potentiality has not yet formed and stays in every average. The rank-8 LoRA of each run is then scored under the two interventions, with every run pooled and intervals clustered by task, since one checkpoint at 3 seeds has $17$ independent units, the tasks, and not $51$. \Cref{tab:ladder} lists every checkpoint, \cref{fig:stages} draws it, and \cref{fig:ladder-groups} draws the task types with the LoRA beside the shift adapter.

By task type (\cref{fig:ladder-groups}), the base model scores $0.57$ on the three text classifications at $9$B tokens and $0.82$ at the released base, where the shift adapter reaches $0.86$ and $0.96$ and the LoRA $0.88$ and $0.96$. On the four commonsense question-answering tasks the base model rises from $0.42$ at $1$B tokens to $0.64$ at the end of stage one and $0.70$ at the released base, where the shift adapter and the LoRA both reach $0.81$. On the three science question-answering tasks it stays at the majority class through $105$B tokens and reaches $0.53$ at $449$B and $0.79$ at the released base, where the shift adapter reaches $0.84$ and the LoRA $0.86$. On the four domain tasks it stays at the majority class through $105$B tokens and reaches $0.37$ at $449$B and $0.50$ at the released base, where the shift adapter reaches $0.52$ and $0.62$ and the LoRA $0.52$ and $0.60$. On HellaSwag it stays at the majority class through $105$B tokens and reaches $0.28$ at $449$B, $0.44$ at the released base and $0.52$ at the instruct model, where the shift adapter reaches $0.71$, $0.84$ and $0.85$ and the LoRA $0.75$, $0.88$ and $0.87$. On the two expert question-answering tasks it stays at the majority class through $105$B tokens and reaches $0.14$ at $449$B and $0.21$ at the released base, where the shift adapter reaches $0.21$ and $0.29$ and the LoRA $0.21$ and $0.30$. From $449$B tokens on, the two adapters are within $0.04$ of each other in every task type, and the LoRA leads clearly only at $1$B tokens on the classifications, $0.48$ against $0.36$.

\begin{wrapfigure}{r}{0.40\linewidth}
\vspace{-1.2em}
\centering
\includegraphics[width=\linewidth]{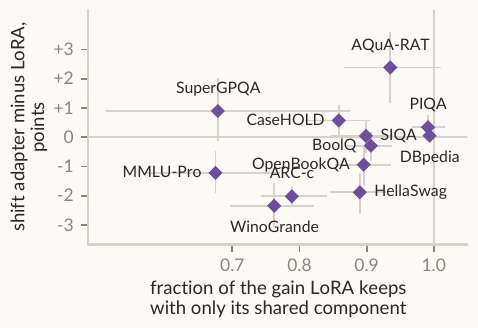}
\vspace{-1.6em}
\caption{\textbf{The gap to the LoRA against the LoRA's reliance on its shared component.} Per task, pooled over the 10 stage-two and released checkpoints of OLMo-2-7B: the difference between the shift adapter and the rank-8 LoRA against the fraction of the LoRA's gain kept with only the shared component. Every run is pooled. The 5 tasks whose mean gain is below $0.05$ over these checkpoints (ARC-Easy, SciQ, SST-2, AG News and MedMCQA) have no gain to divide and are not drawn. Bars are $95\%$ intervals.}
\label{fig:stages-gap}
\vspace{-1em}
\end{wrapfigure}

At the early checkpoints DBpedia, AG News and SST-2 are lifted above the majority class first, and the other tasks join as pretraining continues. At $1$B tokens the LoRA lifts DBpedia to $0.51$ and AG News to $0.41$, where the shift adapter reaches $0.23$ and $0.35$ against majority classes of $0.07$ and $0.25$, and the LoRA alone lifts SST-2 slightly ($0.53$ against a majority class of $0.51$). At $9$B tokens the base model already scores $0.55$ to $0.58$ on these 3 tasks, and both methods lift them, DBpedia to $0.97$, AG News to $0.88$ and $0.87$, and SST-2 to $0.74$ and $0.81$, while the other 14 tasks stay at the majority class. At $105$B tokens BoolQ, SIQA, SciQ and CaseHOLD join them. At $449$B tokens both methods lift every task but AQuA-RAT, and from $3{,}896$B tokens every task. Across all $255$ pairs of task and checkpoint, only three are lifted by one method alone: SST-2 at $1$B and ARC-Easy at $105$B tokens by the LoRA ($0.31$ against $0.26$, majority class $0.27$), and CaseHOLD at $105$B tokens by the shift adapter ($0.28$ against $0.22$, majority class $0.20$).

With only the shared component, the LoRA keeps between $70\%$ and $93\%$ of its gain at every checkpoint, and with that component removed $4\%$ at $1$B tokens, $25\%$ to $47\%$ over the rest of stage one and $51\%$ to $70\%$ from stage two on (\cref{tab:ladder}). At every checkpoint the shared component alone keeps more of the gain than the rest of the update. At $1$B and $9$B tokens the LoRA gains little on most tasks, so both intervals are wide.

\Cref{fig:stages-gap} pools the 10 stage-two and released checkpoints per task. On every task, the two methods differ by at most $0.024$, the shift adapter's lead on AQuA-RAT. The shift adapter trails the LoRA most on WinoGrande, ARC-Challenge and HellaSwag, by about $0.02$ each, and by $0.012$ or less on the other tasks. On those 3 tasks the shared component of the LoRA keeps $76\%$, $79\%$ and $89\%$ of its gain, and $67\%$ to $99\%$ on the other tasks with a gain. Over the 12 tasks with a fraction, the gap between the two methods and the fraction of the LoRA's gain its shared component keeps correlate at Spearman $+0.41$ ($p=0.19$).

\subsection{\update{Behaviors along the training of OLMo-2-7B}}\label{app:ladder-behaviors}

Beside the accuracy on the tasks, we count at each checkpoint how many of a fixed set of behaviors the model can be made to produce, by three routes: two demonstrations in the prompt, a trained shift adapter and a rank-8 LoRA, each adapter trained on $160$ questions and read on $40$ held-out ones.

\textbf{Behaviors.} Three classes of behavior can be scored without a teacher or a judgment. There are $49$ non-Latin writing systems\update{, that is, answering in a language written in its own non-Latin script, such as Thai, Kannada, or Bengali,} scored by Unicode script property, $45$ Latin-script languages, scored by a language identifier, and $42$ output forms, scored by the predicate each instruction states, such as answering only in questions or only in bullets. The output forms follow the verifiable-instruction construction of IFEval \citep{zhou2023ifeval}. Several are its instruction types directly: no commas, a word or sentence count, bullets or a JSON object, capital letters, a forbidden word or letter, and a response language. The others, such as a sonnet or a pros-and-cons table, are built the same way. Every check refuses a degenerate answer, one that is too short, too repetitive or a phrase repeated in a loop, since such an answer can pass a predicate without following the instruction: a repeated character of a script passes a check that reads the share of letters in that script.

\begin{figure}[t]
\centering
\includegraphics[width=\textwidth]{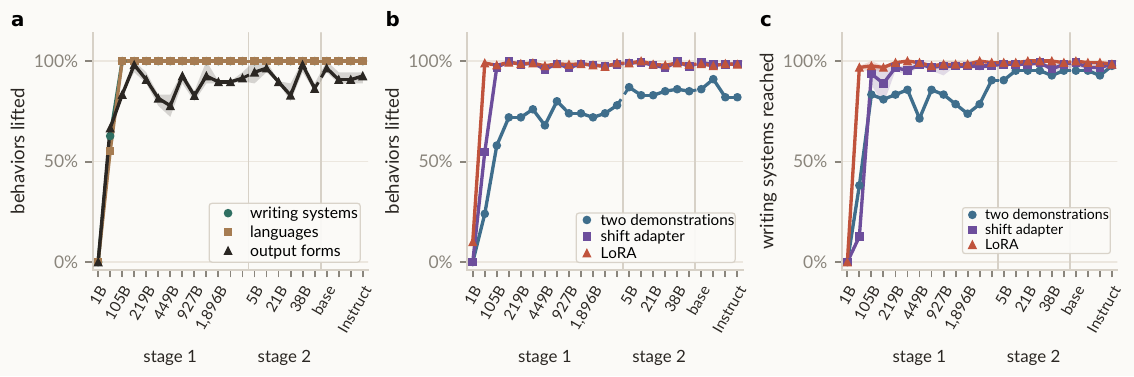}
\caption{\textbf{Three ways of producing a behavior along the training of OLMo-2-7B, at 23 checkpoints.} \textbf{(a)}~The share of each class of behavior a trained shift adapter lifts, where lifted means it raises the model's own rate on the behavior by more than $0.05$ over $40$ held-out answers, of the behaviors the training target itself produces: $42$ writing systems, $38$ languages and $20$ output forms. \textbf{(b)}~The three routes over the $100$ behaviors scored: two demonstrations in the prompt, a trained shift adapter and a rank-8 LoRA. \textbf{(c)}~The $42$ writing systems under one criterion for every route, produced in a majority of the held-out answers. The learning rate is selected per checkpoint, class and method at seed $0$, and the band spans the seeds read at that rate. The horizontal axis changes at each ruled boundary, from stage-one tokens to stage-two tokens to released models, so a segment that crosses one is dashed. \Cref{app:ladder-behaviors,app:ladder-lr} state the criteria and the rate selection.}
\label{fig:development}
\end{figure}

\textbf{Criteria.} A behavior is \emph{lifted} when the adapter raises the model's own rate on it by more than $0.05$ over the $40$ held-out answers (\cref{fig:development}a,~b), and \emph{reached} when the model produces it in a majority of its answers, the criterion the prompt is held to and the one \cref{fig:development}c uses for both adapters.

\textbf{Training target and learning rate.} A pretraining checkpoint's own answers under a system prompt do not contain the behavior: at $9$B tokens they hold the requested script in $0$ of $49$ writing systems. We therefore train every checkpoint toward the instruct model's answers to the same $160$ questions, one target for every checkpoint, and score every answer against the instruction itself. That target produces $42$ of the $49$ writing systems, $38$ of the $45$ languages and $20$ of the output forms, and those are the counts below. The learning rate is selected per checkpoint, class and method, from 4 rates for the shift adapter and 3 for the LoRA, with a higher rate added wherever the highest one won, as the rate that reaches the most behaviors at seed $0$, and every seed is read at it (\cref{app:ladder-lr}).

\textbf{Results.} At $1$B tokens OLMo-2-7B produces none of the $42$ writing systems by any route. At $9$B tokens two demonstrations reach $16$ of them, a rank-8 LoRA $40$ to $41$ across the seeds, and a trained shift adapter $5$ to $6$. From $105$B tokens on, the three routes stay close to the whole set: the prompt between $30$ and $41$, the shift adapter between $33$ and $42$, and the LoRA between $40$ and $42$ at every later checkpoint (\cref{fig:development}c). At 7 further checkpoints between $9$B and $105$B tokens, a trained shift adapter reaches $6$, $10$, $22$, $21$, $29$, $30$, $32$, $39$ and $41$ of the $42$ writing systems at $9$B, $13$B, $21$B, $30$B, $42$B, $55$B, $68$B, $84$B and $105$B tokens, while a rank-8 LoRA stays between $36$ and $42$. \update{A rank-8 LoRA reaches these writing systems at $9$B tokens only with both of its components, while the shared component that a shift adapter actualizes forms between $9$B and $105$B tokens, as the shift adapter reaches more of them (\cref{app:existence}).}
The other two classes follow the same order. On the $38$ languages the prompt reaches $3$ at $9$B tokens, the shift adapter $0$ to $1$ and the LoRA $31$ to $34$, and at $105$B tokens the three reach $11$, $16$ to $26$ and $35$ to $37$. On the $20$ output forms the LoRA reaches $12$ to $13$ at $9$B tokens, against $3$ to $4$ for the shift adapter and $5$ for the prompt, and from $105$B tokens on all three sit between $8$ and $15$. Over all $100$ behaviors the LoRA reaches $84$ at $9$B tokens, the shift adapter $9$ to $11$ and the prompt $24$, and at the instruct model the three reach $90$ to $91$, $88$ to $93$ and $82$ (\cref{fig:development}b). The distance between the LoRA and the shift adapter is widest at the earliest checkpoints and closes as pretraining continues.

\textbf{The prompt along the training.} Over all $100$ behaviors the prompt falls from $91$ at the supervised model to $82$ after preference tuning and at the instruct model. Open-Assistant's supervised chain from Pythia-12B shows the same fall, from $37$ of the $40$ writing systems at the base model to $29$. In both post-training chains the instruction-tuned model follows a bare sequence of demonstrations less closely than the base model it came from, while both adapters reach the same behaviors. From $105$B tokens on, the share of the prompt's behaviors that a shift adapter reaches does not depend on how strongly the prompt reaches them: it is within $13$ points whether the prompt reaches a behavior in all eight of its answers or in fewer (AUC $0.30$ to $0.56$, permutation $p\geq0.352$). At $9$B tokens, the few behaviors a shift adapter reaches are those the prompt reaches best, $5$ of $5$ against $16$ of $23$ (AUC $0.76$, $p=0.023$).

\subsection{\update{Two further models: OLMo-2-1B and Pythia-12B}}\label{app:ladder-further}

\begin{figure}[t]
\centering
\includegraphics[width=\textwidth]{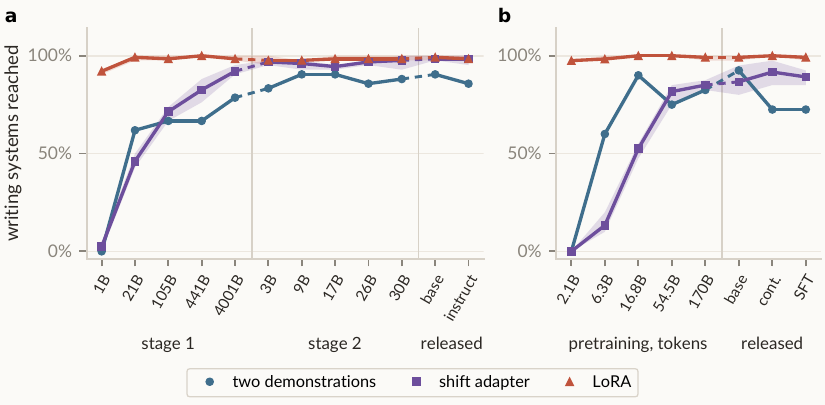}
\caption{\textbf{The same three routes along the training of two further models, on the writing systems each training target produces.} \textbf{(a)}~OLMo-2-1B, a second model with two pretraining stages, at 12 checkpoints over $42$ writing systems. \textbf{(b)}~Pythia-12B, another corpus and another laboratory, with no second pretraining stage, at 8 checkpoints over $40$. A behavior counts when the model produces it in a majority of the held-out answers, the criterion \cref{fig:development}c uses. The learning rate is selected per checkpoint and method at seed $0$, and the band spans the seeds read at that rate. The horizontal axis changes at each ruled boundary, so a segment that crosses one is dashed.}
\label{fig:second-ladders}
\end{figure}

OLMo-2-1B, a second model with two pretraining stages, repeats the measurement at $12$ checkpoints with the same target and the same rate rule (\cref{fig:second-ladders}a). At $1$B tokens a rank-8 LoRA already reaches $38$ to $39$ of the $42$ writing systems, where two demonstrations reach none and a trained shift adapter $0$ to $2$. At $21$B tokens the prompt reaches $26$ and the shift adapter $18$ to $21$, and from $105$B tokens on the three routes stay between $28$ and $42$, reaching $38$, $41$ to $42$ and $41$ to $42$ at the released base model.
Pythia-12B, trained on the Pile by EleutherAI without a second pretraining stage, has no instruct model of its own, so its adapters train toward the answers of Llama-3.1-8B-Instruct to the same questions. A rank-8 LoRA reaches $39$ to $40$ of the $40$ writing systems that target produces at every one of its $8$ checkpoints, including the first at $2.1$B tokens, where the prompt and a trained shift adapter reach none (\cref{fig:second-ladders}b). A trained shift adapter reaches $4$ to $8$ at $6.3$B tokens, $19$ to $22$ at $16.8$B, $30$ to $34$ at $54.5$B and $32$ to $39$ from $170$B tokens on, against $24$ to $37$ for the prompt over the same range.
In all 3 models a rank-8 LoRA reaches these behaviors at the first checkpoint we measure, and a shift adapter reaches them later, in Pythia-12B after about $50$ times more tokens. \update{A constant thus actualizes a behavior once pretraining has formed the capacity for it, as the principle predicts, while a LoRA can reach it earlier by adding what the base model lacks (\cref{app:existence}).}

\subsection{The learning rate along the training}\label{app:ladder-lr}

The selected rate varies along the training and between the classes of behavior. Over the $92$ runs of a checkpoint and a class, the shift adapter takes $3\times10^{-3}$ in $66$, $10^{-3}$ in $13$, $10^{-2}$ in $12$ and $3\times10^{-2}$ in $1$, and the rank-8 LoRA $2\times10^{-4}$ in $51$, $6\times10^{-4}$ in $28$, $6\times10^{-5}$ in $12$ and $2\times10^{-3}$ in $1$. Wherever the highest rate of the range wins, a higher rate is also trained and loses. The separation at $9$B tokens does not depend on the selection: no rate of the shift adapter reaches more than $6$ writing systems there, while the LoRA reaches $39$ to $41$ at three of its rates.

\subsection{What a shift adapter reaches early is new text}\label{app:ladder-recall}

\begin{figure}[t]
\centering
\includegraphics[width=0.65\linewidth]{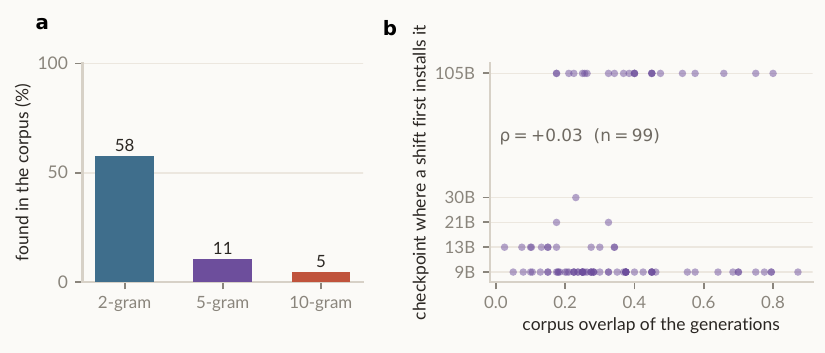}
\caption{\textbf{Corpus overlap of the generations.} \textbf{(a)}~The share of the $n$-grams of a shift adapter's own generations that appear in Dolma, by length, over $99$ behaviors. \textbf{(b)}~The corpus overlap of each behavior against the checkpoint at which its shift adapter first reaches it, at the rate that checkpoint selects.}
\label{fig:memorisation}
\end{figure}
A behavior a shift adapter reaches early could be text the checkpoint has seen and repeats. We count exact $n$-grams of the shift adapter's own generations for the writing systems in the Dolma corpus \citep{soldaini2024dolma} that OLMo-2 was trained on, through the infini-gram index \citep{liu2024infinigram}. The share found falls with length, $58\%$ of bigrams, $11\%$ of $5$-grams and $5\%$ of $10$-grams, which shows that the index finds text where it exists. A behavior's corpus overlap and the checkpoint at which its shift adapter first reaches it correlate at $\rho=0.03$ over $99$ behaviors (\cref{fig:memorisation}). A shift adapter that reaches a behavior early makes the model write new text in it.

\FloatBarrier
\subsection{Width of the region in magnitude and direction}\label{app:basin}

We train a shift adapter for each of the setup's 4 traits on the 18 models, and vary its magnitude and its angle to the trained direction, scoring $256$ held-out probes at every point. Each pair of model and trait trains at its own rate, since the rate that maximizes trained adherence spans a factor of $40$ across the 18 models and varies as widely between the traits of one model. We select it from 5 rates between $5\times10^{-5}$ and $2\times10^{-3}$, as the rate with the highest trained adherence, with ties going to the smaller rate.

\begin{figure}[t]
\centering
\includegraphics[width=\textwidth]{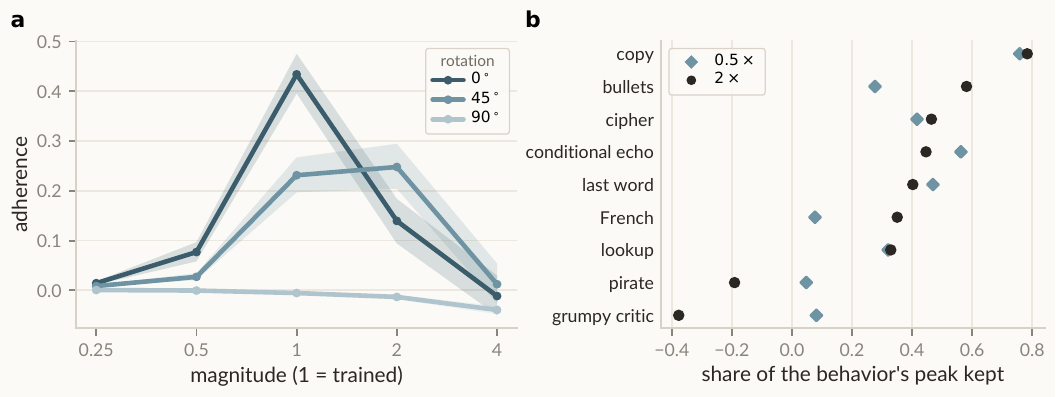}
\caption{\textbf{A region is narrow in magnitude.} \textbf{(a)}~Adherence against the magnitude of the shift adapter (log axis, $1$ is the trained magnitude) with the shift adapter rotated by $0^\circ$, $45^\circ$ or $90^\circ$ toward a random orthogonal direction, mean over the runs of the setup's 4 traits with bootstrap $95\%$ intervals. \textbf{(b)}~Per behavior of the setup, the share of its own peak kept at half and at twice the trained magnitude, without rotation: the 4 traits from these runs and the 5 rule behaviors from the runs of \cref{app:edge}. The rotation toward the three kinds of direction is in \cref{fig:region_geometry}.}
\label{fig:basin}
\end{figure}

Pooled adherence is $0.015$ [$0.012$, $0.019$] at $0.25\times$ the trained magnitude, $0.077$ [$0.058$, $0.097$] at $0.5\times$, $0.434$ [$0.395$, $0.475$] at the trained magnitude, $0.140$ [$0.094$, $0.184$] at $2\times$, and $-0.011$ [$-0.050$, $0.030$] at $4\times$, and a $90^\circ$ rotation leaves $-0.005$ [$-0.009$, $-0.001$] (\cref{fig:basin}a). Rotated by $45^\circ$, the shift adapter keeps more at twice the trained magnitude ($0.248$) than at the trained magnitude ($0.231$): at twice the magnitude its component along the trained direction is $1.41$ times the trained one, the component that sets the edge (\cref{app:edge}). At twice the trained magnitude, bullets keep $58\%$ of their peak and French $35\%$, while pirate and grumpy critic fall below the base model ($-19\%$ and $-38\%$, \cref{fig:basin}b). The peak reproduces across the seeds of a pair: the standard deviation of its adherence over the seeds has a median of $0.011$ and is at most $0.085$ over the pairs.

\textbf{Distance from the base model's states.} We also measure how far the adapted state at the last input token lies from the base model's, on $64$ held-out prompts over the second half of the layers: the median over prompts of the mean distance to the $8$ nearest base states, divided by the same distance among the base states, so that $1$ is as close as the base states lie to each other. The distance rises with magnitude, from $0.91$ at $0.25\times$ and $1.04$ at $0.5\times$ to $1.47$ [$1.42$, $1.51$] at the trained magnitude, $1.79$ at $2\times$ and $2.34$ at $4\times$. Rotated by $90^\circ$ at the trained magnitude, the state lies at $0.98$ [$0.95$, $1.00$]. Below half the trained magnitude, the state stays among the base model's states and the behavior is not produced. Above the trained magnitude, the state moves further from them while adherence falls.

\subsection{Where the edge of the region falls}\label{app:edge}

To locate the edge, we measure the 18 models and the setup's 4 traits at 9 angles and 7 magnitudes near the trained scale. Adherence halves $44.1^\circ$ from the trained direction. An edge is read per run as the point where the straight line between two neighboring grid points crosses the threshold, and we report the median over runs. On the 4 traits of the setup, on the 18 models, the edge at $90\%$ of the trained shift adapter's gain runs from $0.83$ to $1.25$ times the trained magnitude and to $31.4^\circ$, and the edge at half the gain from $0.70$ to $1.66$ times and to $44.1^\circ$. The same grid on the 12 verifiable instruction types of IFEval \citep{zhou2023ifeval}, on the 18 models, puts the edge at $90\%$ of the gain from $0.92$ to $1.05$ times the trained magnitude and at $16.4^\circ$, and the edge at half the gain from $0.59$ to $1.27$ times and at $58.7^\circ$. There a program decides compliance with no teacher, and the shape is the same: a narrow band in magnitude and a wide one in direction. On the 5 rule behaviors of the setup, on the 18 models, the edge at $90\%$ of the gain runs from $0.70$ to $1.40$ times the trained magnitude and to $42.6^\circ$, and the edge at half the gain spans the whole measured range of magnitudes, $0.5$ to $2$ times, and reaches $58.1^\circ$. Per behavior the half-gain angle runs from $52.1^\circ$ (lookup) to $67.5^\circ$ (copy), wider than on any trait. The per-trait medians span $39.8^\circ$ (\emph{grumpy critic}) to $47.2^\circ$ (\emph{bullets}). A magnitude above the trained one costs less than one below it: the median pair keeps $0.070$ more of its peak at $1.5\times$ than at $0.7\times$ (mean $0.061$), and 44 of the 71 pairs of model and trait favor $1.5\times$.

Write a perturbed state in the coordinates the perturbation itself uses, a component $\alpha\cos\theta$ along the trained direction and a component $\alpha\sin\theta$ across it. The second is exact, since the perturbation orthogonalizes the added direction against the trained one before rotating. The first alone sets the edge. One shared curve in $\alpha\cos\theta$ accounts for a median $0.617$ of the variation within one run, against $0.239$ for the distance from the trained state, $0.261$ for the angle and $0.131$ for the magnitude. On the traits, the band of $\alpha\cos\theta$ that holds half the peak runs from $0.70$ to $1.66$ ($0.60$ to $1.84$ with the rule behaviors), and inside that band the fall-off changes by at most $0.16$ of the peak as the across component grows to twice the trained norm ($0.28$ at any fixed $\alpha\cos\theta$). The angle at which adherence halves therefore climbs with magnitude, from $33.9^\circ$ at $0.85\times$ to $44.1^\circ$ at the trained magnitude and $70.0^\circ$ at $2\times$ over the runs that still fall below half there, which is what a constant $\alpha\cos\theta$ predicts. A region is therefore a band in that one component: what a state keeps depends on how much of the trained direction it retains.

\subsection{What the shift adapter is rotated toward}\label{app:rotate}

\begin{figure}[h]
\centering
\includegraphics[width=\textwidth]{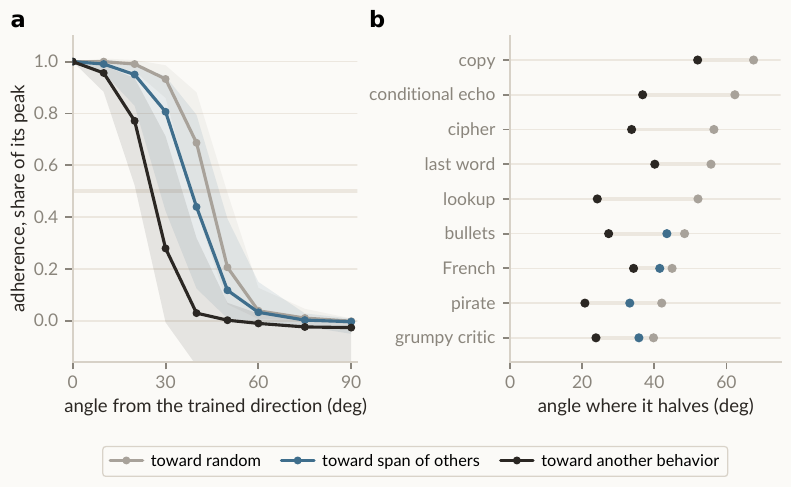}
\caption{\textbf{A region is wide only across the directions the model does not use.} \textbf{(a)}~Adherence as a share of each run's peak against the angle from the trained direction, at the trained magnitude, for each kind of added direction, median over the 18 models and the setup's 4 traits with quartile bands. All three start from the same trained state and are orthogonalized against it, so only the kind of direction differs: turning toward another behavior's own state costs half the peak by $25^\circ$, toward a random direction not until $44^\circ$. \textbf{(b)}~The angle at which adherence halves, per behavior of the setup (4 traits and 5 rule behaviors), median over models, toward each kind of direction. The angle toward another behavior is the smallest in every behavior, and the span lies between the other two in every trait. \Cref{fig:region-inline} draws the region itself over magnitude and angle.}
\label{fig:region_geometry}
\end{figure}

We measure how far the edge depends on the kind of direction the shift adapter is rotated toward. The magnitude is held at the trained one and the added direction is drawn three ways: at random, inside the span of the other traits' shift adapters, and along another trait's own shift adapter. On the 4 traits of the setup and the 18 models, the shift adapter keeps $90\%$ of its gain up to $31.3^\circ$ toward a random direction, $23.8^\circ$ toward a direction inside the span and $12.9^\circ$ toward another trait's shift adapter, and half of it up to $43.9^\circ$, $38.6^\circ$ and $25.6^\circ$. On the 5 rule behaviors, turning toward another behavior's state halves the gain at $36.0^\circ$ and keeps $90\%$ up to $22.0^\circ$, against $58.1^\circ$ and $42.6^\circ$ toward a random direction (\cref{app:edge}). A region is about twice as wide across the directions the model does not use for another trait as across the directions it does, and the random edge exceeds the other-trait edge in $205$ of the $213$ runs and in all 4 traits. The in-span edge sits below the random edge in every trait. On the traits of the setup, the random edge agrees with that of \cref{app:edge}, $43.9^\circ$ against $44.1^\circ$, on separate runs (\cref{fig:region_geometry}).

\subsection{\update{Connectivity of the region}}\label{app:region}

\begin{wrapfigure}{r}{0.40\linewidth}
\vspace{-1.2em}
\centering
\updatebox{\includegraphics[width=0.62\linewidth]{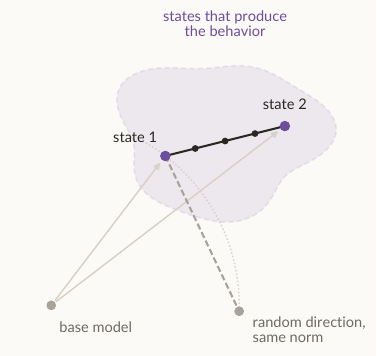}}
\vspace{-0.6em}
\captionsetup{type=\rfigtype}
\caption{\update{\textbf{The connectivity test.} We score adherence at five points on the straight segment between two independently trained states of one behavior, and on a segment from one state toward a random direction of the same norm.}}
\label{fig:mode-paths-schematic}
\end{wrapfigure}
\begin{figure}[h]
\centering
\updatebox{\includegraphics[width=\dimexpr\textwidth-8.4pt\relax]{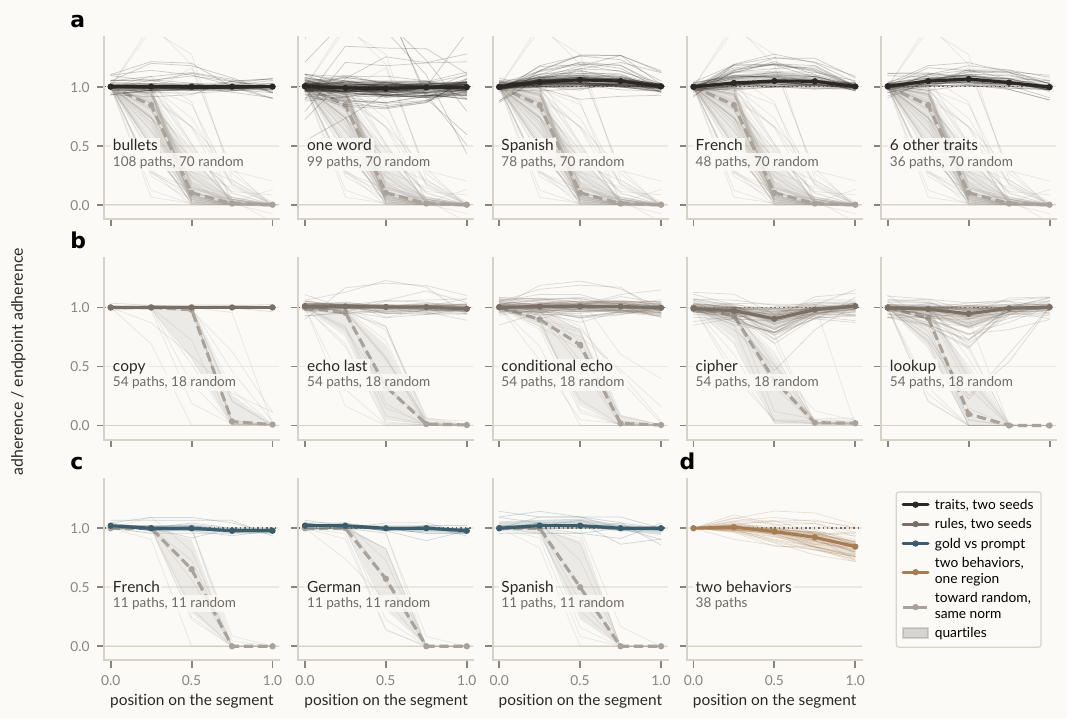}}
\caption{\update{\textbf{Adherence along the straight segment between two trained states of one behavior, and along a segment of the same norm toward a random direction.} Each panel is one behavior. Thin lines are single paths, the bold line is the median, and the band spans the quartiles. Adherence is normalized to the mean of the two endpoints, or to the trained state for the random segment, and the dotted line marks the endpoint level. \textbf{(a)}~Traits, two states trained at different seeds on the 18 models, with the random control pooled over the traits. \textbf{(b)}~The 5 rule behaviors, two seeds, each with its own random control. \textbf{(c)}~A state trained on gold answers against one distilled from a system prompt, 11 models per language. \textbf{(d)}~Two different behaviors that pass the identity call, drawn as the first behavior's adherence.}}
\label{fig:mode_paths}
\end{figure}

We score adherence along the straight segment between two states of one behavior trained at different seeds, on the 18 models. \update{The test is sketched in \cref{fig:mode-paths-schematic}.} We summarize each path by its lowest interior point, as a share of the mean adherence of its two endpoints. The path stays adherent along its whole length: over $369$ segments on 10 traits, the lowest interior point keeps a median $100\%$ of the endpoint gain ($100\%$ on bullets and French, $156$ segments, and $96\%$ on one word, the lowest trait), and the worst per-model mean barrier, over the segments scored at two evaluation seeds, is $-0.019$ (Mistral-7B), at endpoint cosines from $0.13$ on Tulu-3-8B-SFT to $0.91$ on Gemma-2-2B (\cref{fig:mode_paths}). On the 5 rule behaviors of the setup, on the 18 models, the lowest interior point on the line between two trained states keeps a median $98\%$ of their gain, from $90\%$ on the cipher and $94\%$ on the lookup to $100\%$ on copy, while the same line toward a random direction of the same norm keeps $0\%$. Two states that share a sixth of their direction still connect without a trough, so the states that produce one behavior form one connected region. The two states need not share their supervision. A state trained on external gold answers (the base model's own answers translated by a separate machine-translation model) and one distilled from a system prompt stay connected in all 11 models, and this holds in each of 3 languages: the lowest interior point keeps a median $98\%$ of the endpoint gain in French and German and $100\%$ in Spanish, never below $86\%$, with the target language above $0.87$ at every midpoint, $33$ of $33$ model--language pairs. Both routes reach the language they were trained for, so no path is scored on a state that never produced it (lowest rate over the $33$ runs per language $0.92$, $0.88$ and $0.88$, medians $1.00$, $1.00$ and $0.96$). A random direction of the same norm connects in none of the $33$. The two states connect whether or not a nearest-neighbor ranking of their fields places them close: in French the gold state is the prompt state's nearest neighbor in $6$ of the $11$ models and $2$nd to $18$th in $4$, and it connects even on Qwen2.5-0.5B, where the ranking puts it $249$th among the model's trait states ($13$th to $20$th in German).

\subsection{\update{Region size and the strength of the behavior}}\label{app:region-size}

\begin{figure}[t]
\centering
\includegraphics[width=\textwidth]{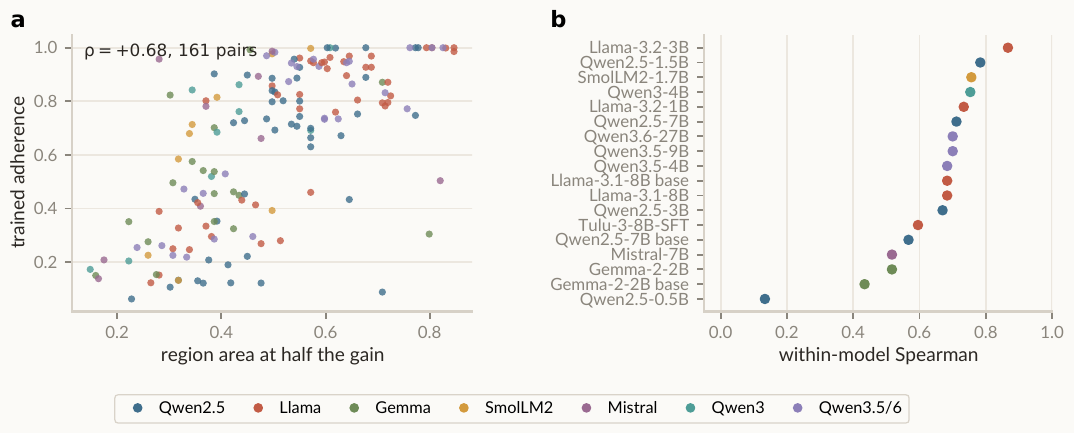}
\caption{\textbf{A larger region gives a stronger behavior, in every model.} \textbf{(a)}~Per pair of model and behavior, the area of the region, the share of the magnitude-by-angle grid of \cref{app:edge} on which a perturbed shift adapter keeps half the trained gain, against the adherence of the trained state, each the mean over runs, one point per pair on the 18 models, colored by family. \textbf{(b)}~Per model, the Spearman correlation of the two over its behaviors.}
\label{fig:region-size}
\end{figure}

\textbf{On the setup.} For each pair of model and behavior, we take the area of the region of actualizing states, the share of the magnitude-by-angle grid of \cref{app:edge} on which a perturbed shift adapter keeps half the trained gain, and the adherence of the trained state, each as the mean over runs. The pairs are the 4 traits and 5 rule behaviors of the setup on the 18 models. A larger region gives a stronger behavior (\cref{fig:region-size}). Within a model, the area correlates with the adherence at Spearman $+0.64$ on average (median $+0.68$) over the 18 models, and the correlation is positive in every one of them, from $+0.13$ on Qwen2.5-0.5B to $+0.87$ on Llama-3.2-3B. Over all pairs it is $+0.68$ ($p<10^{-22}$).

\begin{figure}[h]
\centering
\includegraphics[width=\textwidth]{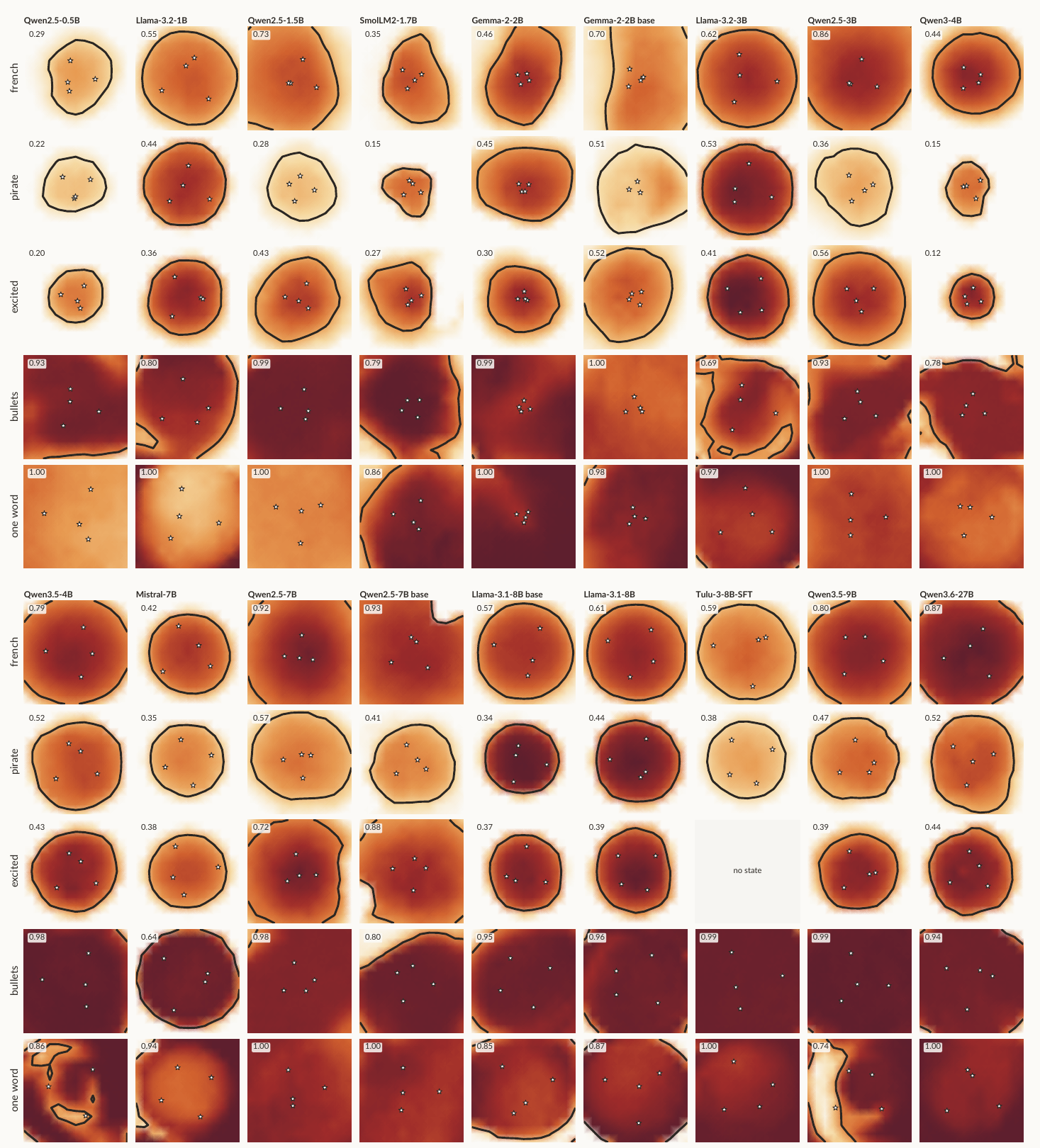}
\caption{\textbf{The region of actualizing states, across behaviors and models.} Each panel is a plane through the shift space of one model, spanned by 4 independently trained states of one behavior (the stars). Every pixel is one shift adapter, added to the model and scored on generated text, so the surface is measured, not fitted. The bold line is the edge of the region at half the peak, and the number in each panel is the area above that edge. Columns are the 18 models from the smallest to the largest, in two blocks of 9, and each row is scaled to its own peak. The top 3 behaviors occupy bounded islands whose size differs across models, while the two format behaviors fill the frame.}
\label{fig:mode_gallery}
\end{figure}

\textbf{What a region looks like.} \Cref{fig:mode_gallery} shows the plane through 4 independently trained states of one behavior, for 5 behaviors on the 18 models. French, pirate and excited child occupy bounded regions whose size differs across models, while bullets and one word fill the whole measured plane, so their regions exceed the window and their reported area is a lower bound. The mean area over the 5 behaviors runs from $0.48$ on SmolLM2-1.7B to $0.84$ on Qwen2.5-7B.

\begin{figure}[t]
\centering
\updatebox{\includegraphics[width=\dimexpr\textwidth-8.4pt\relax]{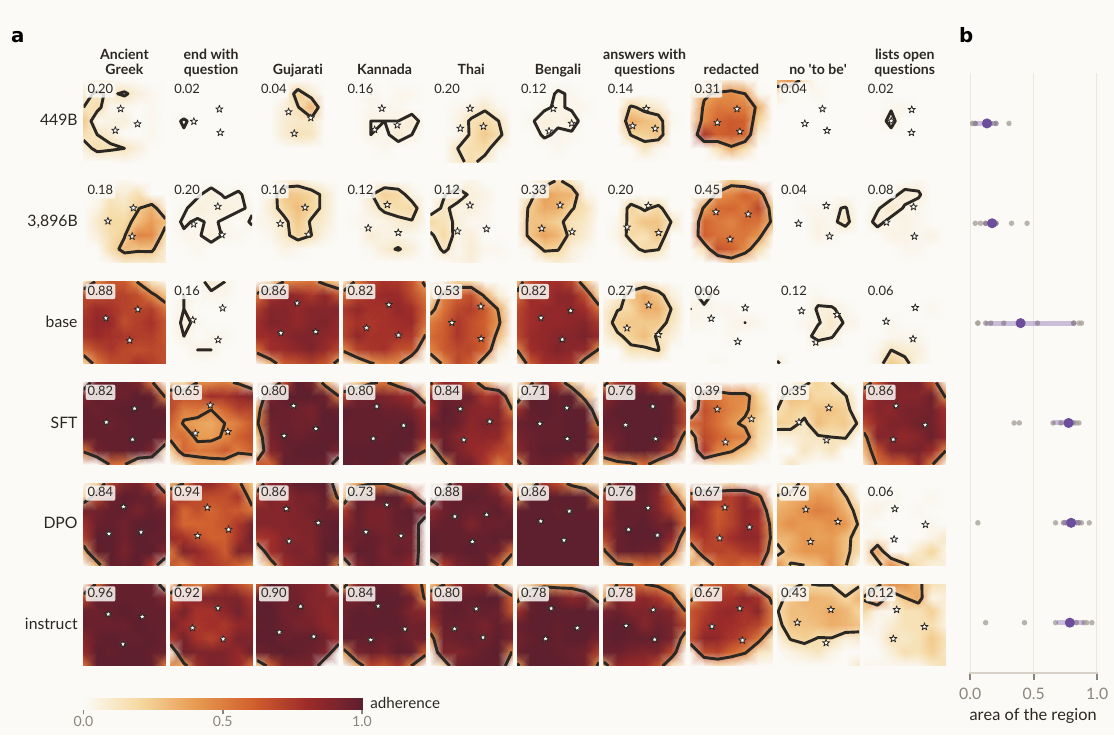}}
\caption{\update{\textbf{Post-training makes the regions larger, behavior by behavior.} \textbf{(a)}~The region of actualizing states for 10 behaviors (columns) at 6 stages of OLMo-2-7B (rows): 2 stage-one checkpoints at $449$B and $3{,}896$B tokens, then the released base, supervised, preference-tuned and instruct models. Each tile is the plane spanned by 3 independently trained states of one behavior (the stars), measured on a $7\times7$ grid. Every grid point is one shift adapter added to the model and scored on generated text, so each plane holds 49 states and the figure 2,940. All tiles share one adherence color scale. The bold line is the edge of the region, where adherence falls to half the plane's peak. The number in each tile is the area, the share of the grid inside that edge. Columns are ordered by area after instruction tuning. \textbf{(b)}~The area at each stage, one gray point per behavior, with the median (dot) and quartiles (bar) over the 10 behaviors.}}
\label{fig:area-stages}
\end{figure}

\textbf{Along training.} At each of 6 stages of OLMo-2-7B, we measure the region on the plane through 3 independently trained states of each of 10 behaviors that a shift adapter reaches by $449$B tokens, as the share of the plane whose adherence stays at or above half its peak. The area rises from $0.13$ at $449$B tokens and $0.17$ at the end of stage one to $0.40$ at the released base and $0.78$, $0.80$ and $0.79$ after supervised, preference and instruction tuning (\cref{fig:area-stages}). Across these $60$ planes, the area predicts \update{the peak adherence on the plane} at Spearman $+0.84$, with a positive correlation within every stage.

\FloatBarrier
\subsection{\update{When the shared component forms during pretraining}}\label{app:existence}

\update{In the sections above, we showed that a rank-8 LoRA produces nearly every writing system (answering in a language written in its own non-Latin script) from the early checkpoints of OLMo-2-7B on, while a shift adapter produces them from $105$B tokens on (\Cref{app:development}). This leaves two possibilities. A constant that produces the writing systems may already exist at $9$B tokens, and training a shift adapter from zero may not find it. Or no such constant exists yet, and the LoRA reaches the writing systems in another way. We can use the interventions defined in \cref{app:interventions} to distinguish the two, since they test whether the gain of the trained LoRA depends on its shared component.}

\update{\textbf{Setup.} We train on OLMo-2-7B at the checkpoints of $9$B and $105$B tokens, with the protocol of \cref{app:development} and 3 seeds. A writing system counts as produced when it appears in the majority of $40$ held-out answers, and we count over the $42$ writing systems of \cref{app:development}. We train every method anew at 3 seeds, so the counts differ by a few from \cref{fig:development}. We compare the following states:}
\update{\begin{itemize}[leftmargin=1.5em,itemsep=2pt,topsep=2pt]
\item \textit{A rank-8 LoRA}, evaluated in full, with only its shared component, and with its shared component removed. These interventions tell whether the gain of the trained LoRA rests on a constant.
\item \textit{A shift adapter trained from zero}, as in \cref{app:development}.
\item \textit{A shift adapter started from the demonstrations.} Its training starts from the shared component of two demonstrations of the target (\cref{app:icl}) instead of from zero. If a suitable constant existed near the one the demonstrations give, training from there would find it more easily.
\item \textit{The demonstrations}, given in the prompt, and their shared component added on its own.
\item \textit{A random vector} with the same norm as the shift adapter at every layer. It shows that a vector of this size does not produce the writing systems by itself.
\end{itemize}}

\update{\textbf{At $9$B tokens.} The full LoRA produces $40$ to $42$ of the $42$ writing systems (\cref{tab:rebuttal-existence}). Its shared component alone produces none of them in any seed, and at most one at any learning rate we tried. Removing the shared component also leaves none. At this checkpoint, the LoRA thus reaches the writing systems only through both of its components together, and no part of its gain comes from a constant. The other constants agree: the shift adapter started from the demonstrations produces at most two, and the shared component of the demonstrations added on its own and the random vector produce none. Trained from zero, the shift adapter reaches $7$ to $8$.}

\begin{rtable}[t]
\centering
\caption{\update{\textbf{Writing systems produced on OLMo-2-7B at $9$B and $105$B tokens.} Number of the $42$ writing systems of \cref{app:development} produced in the majority of $40$ held-out answers, for each of the 3 seeds. The demonstrations are two of the target's own answers (\cref{app:icl}), not those of \cref{fig:development}, so their counts differ from that figure.}}
\label{tab:rebuttal-existence}
\updatebox{\small
\begin{tabular}{lcc}
\toprule
& $9$B tokens & $105$B tokens \\
\midrule
\multicolumn{3}{l}{\textit{A rank-8 LoRA}} \\
\quad full LoRA & $41$\enspace $40$\enspace $42$ & $42$\enspace $42$\enspace $42$ \\
\quad only its shared component & $0$\enspace $0$\enspace $0$ & $37$\enspace $39$\enspace $40$ \\
\quad its shared component removed & $0$\enspace $0$\enspace $0$ & $0$\enspace $0$\enspace $0$ \\
\midrule
\multicolumn{3}{l}{\textit{A shift adapter}} \\
\quad trained from zero & $7$\enspace $7$\enspace $8$ & $41$\enspace $40$\enspace $38$ \\
\quad started from the demonstrations & $2$\enspace $1$\enspace $0$ & $33$\enspace $24$\enspace $29$ \\
\midrule
\multicolumn{3}{l}{\textit{Demonstrations}} \\
\quad two in the prompt & $6$\enspace $13$\enspace $9$ & $17$\enspace $21$\enspace $17$ \\
\quad their constant, added alone & $0$\enspace $0$\enspace $0$ & $0$\enspace $0$\enspace $0$ \\
\midrule
\multicolumn{3}{l}{\textit{Control}} \\
\quad random vector of the same norm & $0$\enspace $0$\enspace $0$ & $0$\enspace $0$\enspace $0$ \\
\bottomrule
\end{tabular}
}
\end{rtable}

\update{\textbf{At $105$B tokens.} The shared component of the LoRA alone now produces $37$ to $40$ writing systems, while removing it leaves none: the gain of the LoRA now rests entirely on its shared component. A shift adapter trained from zero produces $38$ to $41$, nearly as many as the full LoRA ($42$), and one started from the demonstrations $24$ to $33$ (\cref{fig:rebuttal-existence}). The random vector still produces none.}

\begin{rfigure}[t]
\centering
\updatebox{\includegraphics[width=\dimexpr\textwidth-8.4pt\relax]{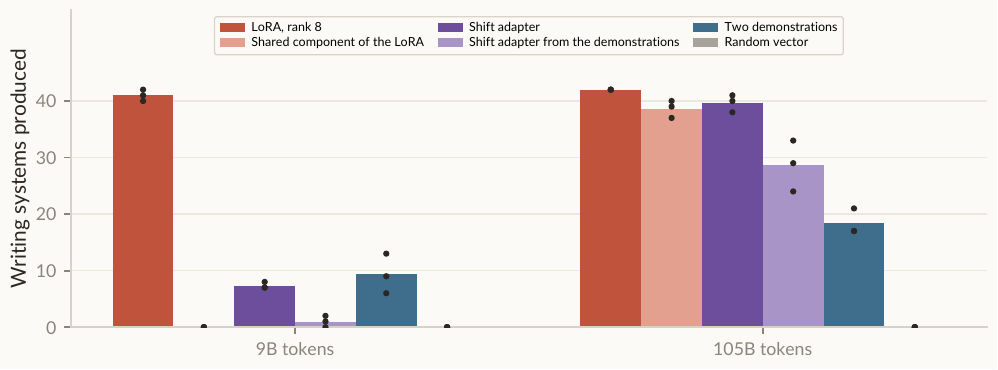}}
\caption{\update{\textbf{Writing systems produced by each method at $9$B and $105$B tokens.} Writing systems produced in a majority of $40$ held-out answers by each method on OLMo-2-7B, bars the mean over 3 seeds and dots the seeds.}}
\label{fig:rebuttal-existence}
\end{rfigure}

\update{The two checkpoints thus show when the capacity forms. At $9$B tokens, a LoRA already produces the writing systems, but not through a constant: neither its own shared component nor any other constant we tested produces them. At $105$B tokens, the LoRA's own shared component produces almost all of them, and a shift adapter trained from zero finds a constant that does the same. The capacity that a constant can actualize therefore forms during pretraining, between $9$B and $105$B tokens. We measure this on the trained LoRA itself, through its interventions, so the result does not depend on whether training a shift adapter succeeds. The potentiality and actuality principle thus states not only how adaptation works, but also from which point in pretraining it applies.}

\subsection{\update{Predicting before adaptation whether an actualizing state exists}}\label{app:conditions}

\update{The potentiality and actuality principle says that a shift adapter actualizes a capacity that the base model already holds. If so, it should be possible to see in the base model, before any adapter is trained, whether an actualizing state exists for a task. We test this with a linear readout of the hidden states of the unadapted model.}

\update{\textbf{Setup.} We use $6$ rules that map a list of $1$ to $12$ pseudo-words to one of $3$ answers: the word count modulo $3$, the word count mapped to the $3$ answers by a fixed random table, the first letter mapped by such a table, the letter count and the vowel count modulo $3$, and random labels, which follow no rule. On 4 architecture families, we train a shift adapter on $10{,}000$ inputs for each rule ($300$ or $1{,}000$ for the random labels) and evaluate it on held-out inputs, whose words never appear in training. Separately, before any adaptation, we fit a linear readout that predicts the answer from the hidden states of the unadapted model. The readout is fit at every input position and at 6 depths, and its position and depth are chosen on validation inputs. A rule can sit at different positions: the first letter, for instance, is read without error at the first token of the input, but at only $0.12$ to $0.30$ at the last token.}

\begin{wrapfigure}{r}{0.42\linewidth}
\vspace{-2.2em}
\centering
\updatebox{\includegraphics[width=\dimexpr\linewidth-8.4pt\relax]{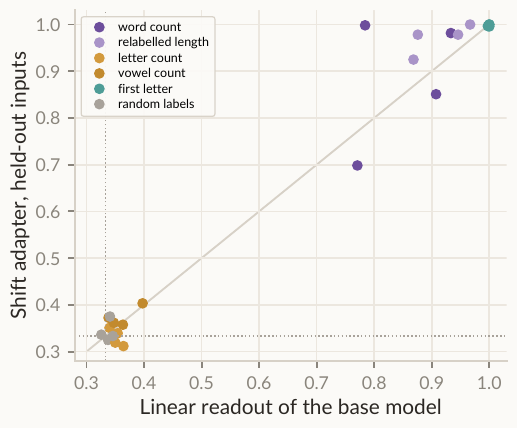}}
\vspace{-0.5em}
\captionsetup{type=\rfigtype}
\caption{\update{\textbf{Held-out accuracy of the shift adapter against a linear readout of the base model.} For $6$ rules on 4 architecture families, means over 3 seeds. Dotted lines mark chance, the solid line equality.}}
\label{fig:rebuttal-conditions}
\vspace{-3.5\baselineskip}
\end{wrapfigure}

\leavevmode\update{\textbf{Results.} For every rule that a shift adapter learns, the readout of the base model already recovers the answer (\cref{tab:rebuttal-conditions}). On the first letter, the word count, and the relabelled length, the shift adapter reaches $0.70$ to $1.00$ on held-out inputs, and the readout $0.77$ to $1.00$. On the letter count, the vowel count, and the random labels, the readout stays near chance, and so does the shift adapter, as the principle predicts ($0.31$ to $0.40$, with chance at $0.33$). Across all $24$ pairs of rule and model, no shift adapter learns a rule that the readout of the base model does not already recover (\cref{fig:rebuttal-conditions}). Across the $24$ pairs, the accuracy of the shift adapter follows that of the readout, at a Pearson correlation of $0.98$ and a Spearman correlation of $0.84$.}

\begin{rtable}[t]
\centering
\caption{\update{\textbf{Accuracy of the shift adapter and of a linear readout of the base model.} Held-out accuracy of the shift adapter (shift) and of a linear readout of the unadapted model (readout), for $6$ rules with $3$ answers each, chance $0.33$. Means over 3 seeds, except the random labels, trained on $300$ or $1{,}000$ inputs with 1 to 3 seeds.}}
\label{tab:rebuttal-conditions}
\updatebox{\scriptsize
\setlength{\tabcolsep}{3.5pt}
\begin{tabular}{lcccccccc}
\toprule
& \multicolumn{2}{c}{Qwen2.5-1.5B} & \multicolumn{2}{c}{Llama-3.2-3B} & \multicolumn{2}{c}{SmolLM2-1.7B} & \multicolumn{2}{c}{Gemma-2-2B} \\
Rule & shift & readout & shift & readout & shift & readout & shift & readout \\
\midrule
First letter & $1.00$ & $1.00$ & $1.00$ & $1.00$ & $1.00$ & $1.00$ & $1.00$ & $1.00$ \\
Word count & $0.98$ & $0.93$ & $1.00$ & $0.78$ & $0.85$ & $0.91$ & $0.70$ & $0.77$ \\
Relabelled length & $1.00$ & $0.97$ & $0.98$ & $0.88$ & $0.98$ & $0.95$ & $0.93$ & $0.87$ \\
Letter count & $0.32$ & $0.35$ & $0.34$ & $0.35$ & $0.31$ & $0.36$ & $0.35$ & $0.34$ \\
Vowel count & $0.36$ & $0.35$ & $0.37$ & $0.34$ & $0.36$ & $0.36$ & $0.40$ & $0.40$ \\
Random labels & $0.33$ & $0.35$ & $0.34$ & $0.33$ & $0.38$ & $0.34$ & $0.33$ & $0.34$ \\
\bottomrule
\end{tabular}
}
\end{rtable}

\leavevmode\update{\textbf{Where adaptation has to add.} The readout also indicates in advance where an actualizing state does not exist, and adaptation has to add to the base model. On the vowel count, which the readout does not recover, a rank-8 LoRA reaches $0.43$ and $0.48$ on Llama-3.2-3B and Qwen2.5-1.5B, and a rank-64 LoRA $0.55$ and $0.52$, while the shift adapter stays near chance, at $0.37$ and $0.36$. A LoRA can add what the base model lacks, while a constant, as the principle predicts, actualizes only what the base model holds.}

\leavevmode\update{A shift adapter thus actualizes what the base model already represents, and whether an actualizing state exists can be estimated on the base model before adaptation. Together with \cref{app:existence}, this suggests a condition for the potentiality and actuality principle: a capacity can be actualized once the base model represents what the task needs, and the base model comes to represent it during pretraining.}

\FloatBarrier
\section{Implications}\label{app:impact}

\subsection{Steering: extraction against training}\label{app:steerstress}

Steering adds a vector to the hidden states of the model. We compare vectors extracted without training with the shift adapter trained on the same behavior, on the 9 behaviors of the setup and the 18 models.

\textbf{Experimental setup.} Three extraction methods give one steering vector per behavior: the mean difference of CAA \citep{rimsky2024caa}, the leading principal component of the same differences \citep{zou2023repe}, and the in-context task vector \citep{hendel2023taskvectors}. Each is added at the single best layer or at every layer, 6 variants in all, each added the way the shift adapter adds its vectors, one vector per layer. An extracted vector gives a direction but not a strength, so the strength of each variant is chosen from 5 values on $200$ held-out instructions, and every method is scored on $800$ further held-out instructions. The shift adapter trains on $400$ instructions for $200$ steps. For each run, the best extracted direction is the variant with the highest adherence on the $800$ instructions, a choice that favors extraction. Scores follow \cref{app:evaluation}: adherence on the traits, where $0$ is the base model, and the exact-match rate on the transformations and maps.

\begin{table}[h]
\centering
\caption{\textbf{Extracted directions against the trained shift adapter on the setup.} Held-out score on $800$ instructions per run (adherence on the traits, exact match on the transformations and maps), mean over runs with a bootstrap $95\%$ interval over models. The best extracted direction is the best of the 6 extracted variants in each run.}
\label{tab:steer-setup}
\footnotesize
\begin{tabular}{l c c r}
\toprule
behavior & shift adapter & best extracted direction & runs \\
\midrule
bullet points & $0.89$ {\tiny $[0.83, 0.94]$} & $0.88$ {\tiny $[0.76, 0.97]$} & $54$ \\
French & $0.43$ {\tiny $[0.39, 0.47]$} & $0.01$ {\tiny $[-0.02, 0.05]$} & $54$ \\
pirate & $0.29$ {\tiny $[0.24, 0.35]$} & $-0.05$ {\tiny $[-0.09, -0.02]$} & $54$ \\
grumpy critic & $0.19$ {\tiny $[0.15, 0.24]$} & $-0.06$ {\tiny $[-0.10, -0.02]$} & $54$ \\
copy & $0.99$ {\tiny $[0.97, 1.00]$} & $0.07$ {\tiny $[0.03, 0.12]$} & $54$ \\
last word & $0.85$ {\tiny $[0.74, 0.94]$} & $0.01$ {\tiny $[0.01, 0.02]$} & $54$ \\
conditional echo & $0.78$ {\tiny $[0.68, 0.86]$} & $0.23$ {\tiny $[0.17, 0.28]$} & $54$ \\
cipher & $0.80$ {\tiny $[0.74, 0.85]$} & $0.43$ {\tiny $[0.34, 0.52]$} & $54$ \\
lookup & $0.86$ {\tiny $[0.80, 0.91]$} & $0.04$ {\tiny $[0.01, 0.06]$} & $54$ \\
\midrule
traits & $0.45$ {\tiny $[0.41, 0.48]$} & $0.20$ {\tiny $[0.15, 0.23]$} & $216$ \\
transformations & $0.87$ {\tiny $[0.80, 0.93]$} & $0.10$ {\tiny $[0.08, 0.14]$} & $162$ \\
maps & $0.83$ {\tiny $[0.77, 0.88]$} & $0.23$ {\tiny $[0.18, 0.28]$} & $108$ \\
all 9 & $0.68$ {\tiny $[0.62, 0.72]$} & $0.17$ {\tiny $[0.14, 0.20]$} & $486$ \\
\bottomrule
\end{tabular}
\end{table}

\textbf{Results.} Pooled over the 9 behaviors, the trained shift adapter reaches a score of $0.68$ [$0.62$, $0.72$], against $0.17$ [$0.14$, $0.20$] for the best extracted direction (\cref{tab:steer-setup}). Per class, the shift adapter reaches $0.45$ on the traits against $0.20$, $0.87$ on the transformations against $0.10$, and $0.83$ on the maps against $0.23$. The extracted directions come close to the trained shift adapter only on bullet points ($0.88$ against $0.89$), and stay at or below the base model on French, pirate and grumpy critic, where the shift adapter reaches $0.19$ to $0.43$. On the transformations and maps, the best extracted direction reaches at most $0.43$, on the cipher, against $0.78$ to $0.99$ for the shift adapter.

\FloatBarrier
\subsection{\update{Steering on AxBench}}\label{app:axbench}
\begin{figure}[t]
\centering
\includegraphics[width=\linewidth]{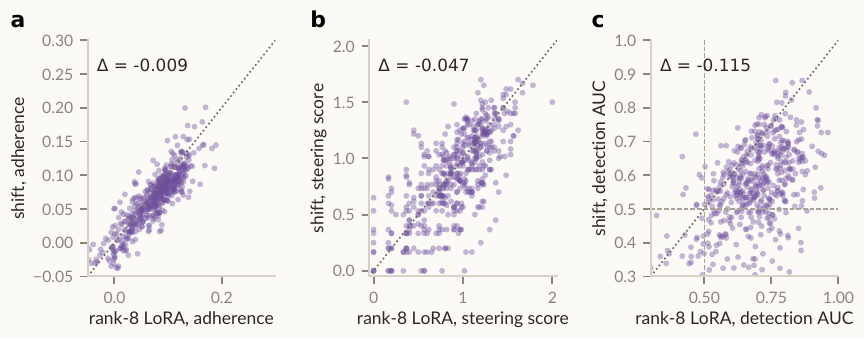}
\caption{\textbf{A shift adapter steers close to the LoRA on AxBench.} The $500$ concepts of AxBench on Gemma-2-2B, a shift adapter against a rank-8 LoRA, paired per concept, at the rates the judged steering score selects. \textbf{(a)}~Adherence, the score of the main text: the overlap of the generation with the reference answer, as a gain over the base model. \textbf{(b)}~Steering score, the benchmark's judge on each generation. \textbf{(c)}~Concept detection, the area under the curve of the trained state read as a probe, with chance as dashed lines. In each panel, $\Delta$ is the mean over concepts of the shift adapter's score minus the LoRA's.}
\label{fig:axbench}
\end{figure}
AxBench \citep{wu2025axbench} steers on $500$ concepts drawn from a sparse autoencoder of Gemma-2, each a text description with positive and negative examples. We train a shift adapter and a rank-8 LoRA on Gemma-2-2B for every concept, 6 rates for the shift adapter and 4 for the LoRA, and score them on the two axes of the benchmark. Steering is judged by GPT-4o-mini, the benchmark's judge model, on each generation with its prompts unchanged: the presence of the concept, the relevance to the instruction and the fluency, each from $0$ to $2$, combined by their harmonic mean. Detection is the concept-against-negative separation the trained state gives when read as a probe. \update{Each method's rate is the one with the best steering score, chosen for every concept on the other $499$ concepts, so that a concept never takes part in choosing its own rate. This gives $3\times10^{-2}$ for the shift adapter and $6\times10^{-4}$ for the LoRA on every concept, and both ranges continue past these rates.} Each concept is judged on its own held-out instructions at $64$ generated tokens, where the benchmark's leaderboard judges $10$ fixed instructions at $128$ tokens, so our levels are not its leaderboard numbers and the comparison is between the two methods under one protocol.
On the benchmark's steering score the shift adapter reaches $0.867$ and the LoRA $0.914$, a paired difference of $-0.047$ $[-0.072, -0.021]$ over the $500$ concepts, $-0.05$, $-0.06$ and $-0.04$ on the seeds (\cref{fig:axbench}b). On adherence to the reference answer, the score of the main text, the shift adapter reaches $0.064$ against $0.073$ for the LoRA, a paired difference of $-0.009$ $[-0.011, -0.007]$ (\cref{fig:axbench}a). At the rate the judge selects, the shift adapter puts the concept into more of its answers and keeps less of the reference wording, and at $3\times10^{-3}$ its adherence is $0.090$, above the LoRA's. On the presence of the concept alone the shift adapter scores $0.965$ and the LoRA $0.983$, and on the other two judged scores it is below the LoRA, by $0.09$ on relevance to the instruction and by $0.02$ on fluency. On detection the shift adapter reaches an area under the curve of $0.571$, above chance on $374$ of the $500$ concepts, and the LoRA $0.686$, above chance on $464$, a difference of $-0.115$ $[-0.126, -0.105]$ (\cref{fig:axbench}c). The rate moves the three scores apart: from $3\times10^{-3}$ to $3\times10^{-2}$ the steering score rises from $0.851$ to $0.867$, while adherence falls from $0.090$ to $0.064$ and the area under the curve from $0.607$ to $0.571$. A shift adapter steers within $6\%$ of the LoRA, and the state of the LoRA separates the concept from its negatives better.

Programming and the web make up $20\%$ of the concepts and science $7\%$, and none of the $500$ is one of our behaviors: none is a writing system, a language or an output form, and the one that names a language we test names it as a topic. The parity of \cref{sec:parameterizations} therefore holds across two concept sets that do not intersect.

\textbf{Extracted directions on the same protocol.}
We compare the trained methods with 3 directions extracted without training, as in \cref{sec:impact}, on the same concepts and under the same protocol.
The mean difference is, at every decoder block, the mean hidden state over the response tokens of the concept's $48$ training examples minus the mean over those of the $64$ shared negatives, AxBench's DiffMean and the contrast of CAA.
The principal component is the leading component of the same positives centered on the negative mean, sign-aligned with the mean difference \citep{zou2023repe}.
The in-context direction is the last-token displacement that a prompt of 4 of the concept's examples induces over the bare prompt, averaged over $16$ of its instructions \citep{hendel2023taskvectors}.
Each direction is added the way the shift adapter adds its vectors, one vector per block, the unit direction at a block scaled by a multiple of the base model's activation norm there, at every block or, for the mean difference, at AxBench's block $20$ alone.
The multiples bracket the best value ($0.005$ to $0.08$ at every block, $0.02$ to $0.8$ at one block), the multiple of each direction is the smallest one within $0.005$ of the best steering score\update{, chosen for every concept on the other $499$ concepts,} the rule that selects the trained methods' rates, and each direction is judged on the same two held-out instructions per concept as the trained methods.
\Cref{tab:axbench-extract} gives the result.
The best extracted direction, the mean difference at every block, reaches a steering score of $0.207$ at multiple $0.04$ and $0.017$ at $0.08$, where the generations degenerate. At block $20$ it reaches $0.180$, the principal component \update{$0.160$} and the in-context direction $0.166$.
The trained shift adapter reaches $0.867$ and the LoRA $0.914$ on the same concepts, a paired difference of $-0.660$ $[-0.697, -0.623]$ between the best extracted direction and the shift adapter.
On the concept sub-score the extracted directions reach $0.16$ to $0.26$ against $0.97$ for the shift adapter, and they raise the steering score above zero on \update{$76$} to $132$ of the $500$ concepts against $435$ for the shift adapter and $446$ for the LoRA. Their relevance sub-scores are above those of the trained methods, and so is their fluency, except for the mean difference at every block ($0.79$).
The level of the mean difference at block $20$, $0.180$, is the level the benchmark's own board reports for it, $0.178$ \citep{wu2025axbench}, under a protocol that differs in the instructions and the length judged.

\begin{table}[h]
\centering
\caption{\textbf{Extracted directions against the trained shift adapter on AxBench.} The $500$ concepts on Gemma-2-2B, each direction at the multiple of the base activation norm (extracted) or the learning rate (trained) its steering score selects \update{on the other concepts (the principal component takes $0.02$ on $5$ concepts, every other method one value on all)}, steering score and its three sub-scores from $0$ to $2$, mean over concepts. The column $>0$ counts the concepts on which the score is above zero, and the last column gives the paired difference to the trained shift adapter with a $95\%$ bootstrap interval over concepts. The extracted directions have one seed, the trained methods three.}
\label{tab:axbench-extract}
\footnotesize
\setlength{\tabcolsep}{2pt}
\begin{tabular}{lccccccl}
\toprule
direction & scale & score & concept & relevance & fluency & $>0$ & minus the shift adapter \\
\midrule
mean difference, every block & $0.04$ & $0.207$ & $0.26$ & $1.73$ & $0.79$ & $132$ & $-0.660$ {\tiny [$-0.697$, $-0.623$]} \\
mean difference, block 20 & $0.4$ & $0.180$ & $0.18$ & $1.92$ & $1.19$ & $90$ & $-0.687$ {\tiny [$-0.727$, $-0.646$]} \\
principal component, every block & $0.005$ & \update{$0.160$} & $0.16$ & $1.94$ & \update{$1.31$} & \update{$76$} & \update{$-0.707$ {\tiny [$-0.746$, $-0.667$]}} \\
in-context direction, every block & $0.01$ & $0.166$ & $0.17$ & $1.94$ & $1.34$ & $80$ & $-0.701$ {\tiny [$-0.740$, $-0.661$]} \\
shift adapter & $3\times10^{-2}$ & $0.867$ & $0.97$ & $1.49$ & $1.12$ & $435$ & --- \\
rank-8 LoRA & $6\times10^{-4}$ & $0.914$ & $0.98$ & $1.58$ & $1.14$ & $446$ & --- \\
\bottomrule
\end{tabular}
\end{table}

\FloatBarrier
\subsection{Convergence across models}\label{app:crossmodel}\label{app:behavior}\label{app:atlas-full}

\begin{wrapfigure}{r}{0.42\linewidth}
\vspace{-1.2em}
\centering
\includegraphics[width=\linewidth]{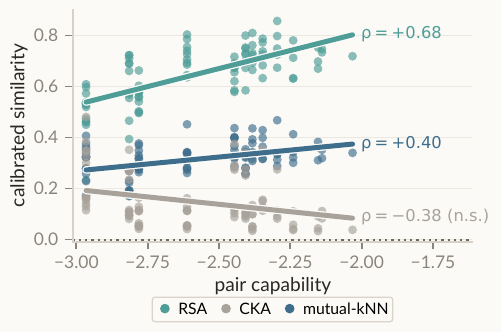}
\vspace{-1.6em}
\caption{\textbf{RSA and mutual-$k$NN rise with capability, CKA does not.} Calibrated similarity of the atlases of two models against the capability of the pair, for the $91$ pairs of the 14 models on the $75$ shared behaviors, with a least-squares line per measure and its Spearman $\rho$ at the line's end. The capability of a pair is the lower of the two models' WikiText log-likelihood per token. \Cref{fig:implications}b draws the same pairs for RSA and mutual-$k$NN.}
\label{fig:convergence-cka}
\end{wrapfigure}
\begin{figure}[ht]
\centering
\includegraphics[width=\linewidth]{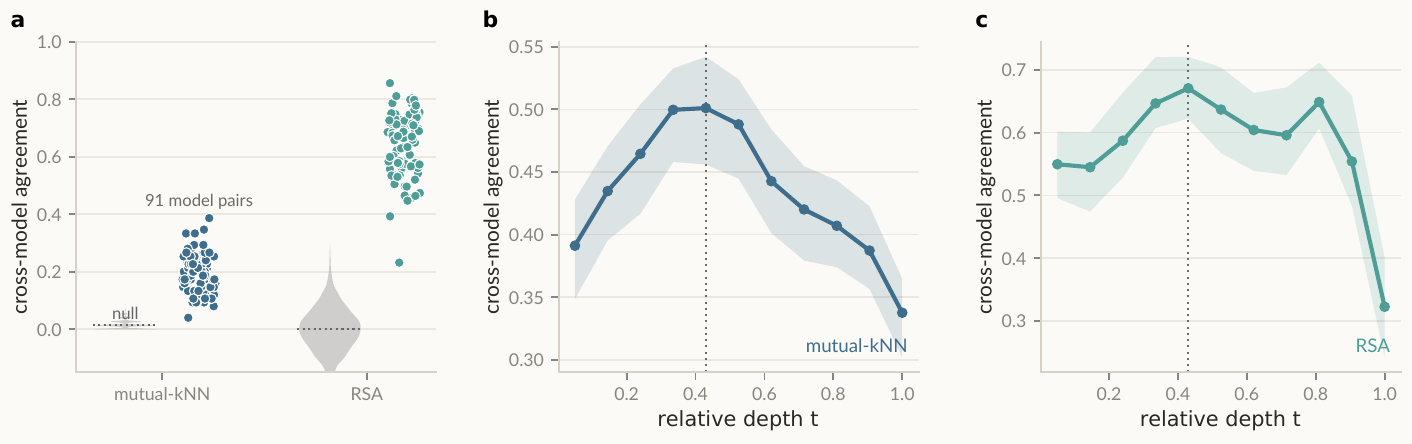}
\caption{\textbf{The arrangement is shared and agrees most in the middle layers} (14 single-seed atlases, $75$ shared behaviors). \textbf{(a)}~Cross-model agreement under mutual-$k$NN and RSA over the $91$ model pairs, against the permutation null ($60$ permutations per pair). \textbf{(b)}~Mutual-$k$NN and \textbf{(c)}~RSA against relative depth, mean over the pairs with quartile bands: mutual-$k$NN peaks at relative depth $0.43$ and RSA at $0.43$, and both fall at the readout.}
\label{fig:arrshared}
\end{figure}

\textbf{Conventions.} The permutation test of each of the $91$ pairs in \cref{sec:impact} is a pairwise comparison and carries no family-wise correction. Every comparison of the 14 atlases uses the $75$ behaviors that every atlas holds (\cref{app:setup-depth}). The association with capability is unchanged ($\rho=0.68$) when capability is measured in bits per byte instead of log-likelihood per token. RSA and mutual-$k$NN rise with the capability of the pair and CKA does not (\cref{fig:convergence-cka}): over the $91$ pairs, RSA at Spearman $\rho=0.68$ ($p=0.003$), mutual-$k$NN at $\rho=0.40$ ($p=0.033$) and CKA at $\rho=-0.38$ ($p=0.93$), each $p$ from a one-sided permutation of the model labels. The 3 measures differ in what they cannot see. RSA ranks the cosines between states, so it does not see one state rescaled or the cosines changed in any order-preserving way. CKA centers the atlas first, so it does not see the atlas moved as a whole or rescaled. Mutual-$k$NN reads neighbor sets, so it does not see any change that leaves each state's neighbors in place. A rotation of the coordinates is invisible to all three: a similarity between two atlases reads the pattern of distances among their states, not their coordinates, which differ from model to model. \Cref{fig:arrshared} draws the similarity over the $91$ pairs and through depth.

\FloatBarrier
\subsection{\update{Neural thickets}}\label{app:thickets}

\citet{gan2026thickets} add Gaussian noise of a fixed size to every weight of a pretrained model, score each perturbed model on a selection split, and keep the best, which they call experts. Under the principle, a perturbation finds an expert only where the region of actualizing states of the behavior is wide enough to be hit by chance (\cref{sec:impact}). We repeat their search on the setup and relate where it finds experts to the size of the region measured in \cref{app:edge}.

\begin{wrapfigure}{R}{0.46\linewidth}
\vspace{-1.2em}
\centering
\updatebox{\includegraphics[width=\dimexpr\linewidth-8.4pt\relax]{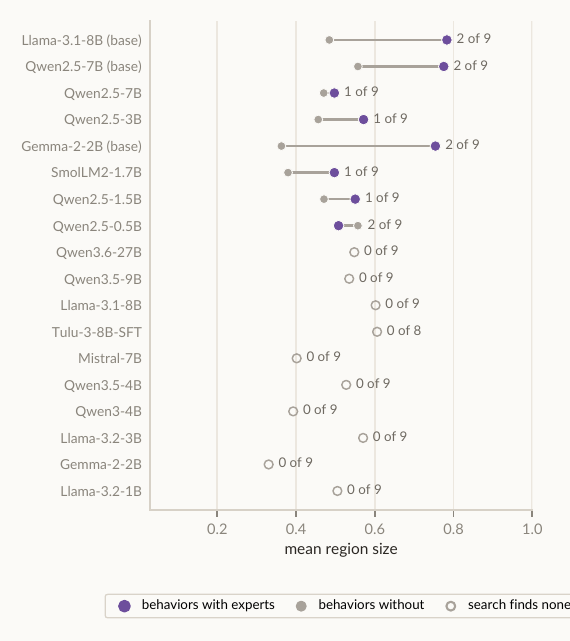}}
\caption{\update{\textbf{Random search finds experts on the widest regions, model by model.} Per model, the mean size of the region of actualizing states (the share of the magnitude-by-angle grid on which a perturbed shift adapter keeps half the trained gain, \cref{app:edge}) over the behaviors of the setup on which the search finds an expert (purple) and over those on which it finds none (grey), with the number of behaviors with an expert. A hollow mark is a model in which the search finds none. All 18 models.}}
\label{fig:thickets-region}
\vspace{-0.5\baselineskip}
\end{wrapfigure}

\textbf{Experimental setup.} For each model we draw $300$ perturbed models at each noise size, 6 sizes for GSM8K, the task of \citet{gan2026thickets}, and 3 for each of the 9 behaviors of the setup (5 GSM8K sizes for Tulu-3-8B-SFT, and 3 and 2 for the Qwen2.5-7B base model), with the sizes placed per model in units of the displacement they cause in the hidden state. Each perturbed model is scored on a selection split of $32$ items, and the $20$ best are scored again on a disjoint held-out split of $64$ items. A perturbed model is an \emph{expert} when it still answers fluently (distinct-2 of at least $0.5$) and gains at least $0.10$ over the base model on the held-out split. For each model, task and seed, the \emph{expert fraction} is the share of the $20$ re-scored models that are experts, \update{at the noise size with the highest mean share over the other two seeds of the same model and task, so that the size is never chosen on the seed it is read on.}

\textbf{Where the search finds experts.} \update{On GSM8K the expert fraction is $0.20$ on average over the 18 models. It is highest on Qwen2.5-3B and Qwen3.5-9B ($0.87$ each) and the Qwen2.5-7B base model ($0.68$), followed by Qwen3.5-4B ($0.38$), Qwen2.5-7B ($0.25$), Llama-3.1-8B ($0.22$), Tulu-3-8B-SFT ($0.18$) and Qwen3.6-27B ($0.12$), and at most $0.03$ on the other 10 models. On the behaviors of the setup it is $0.013$ on average: the search finds experts on 12 pairs of model and behavior in 8 models (copy and the conditional echo on the base models of Llama-3.1-8B, Gemma-2-2B and Qwen2.5-7B, the conditional echo on Qwen2.5-3B, the conditional echo and bullets on Qwen2.5-0.5B, and bullets on Qwen2.5-1.5B, Qwen2.5-7B and SmolLM2-1.7B), and none in the other models.}

\textbf{Experts and the size of the region.} In the 18 models, the behaviors on which the search finds an expert have wider regions than those on which it finds none (\cref{fig:thickets-region}). Over the $72$ pairs of model and behavior in the 8 models where the search finds any expert, the region size separates the two at an AUC of $0.82$ ($p<0.001$) and correlates with the expert fraction at Spearman \update{$+0.43$} ($p<0.001$). Over all pairs in the 18 models, the same values are \update{$0.79$ and $+0.27$} ($p<0.001$ each). Split into quartiles of region area, the search finds an expert in $0\%$ of the pairs of the narrowest quartile (mean area $0.29$), then in \update{$3\%$} and $10\%$, and in \update{$17\%$} of the widest (mean area $0.71$). The \update{12} expert pairs have regions covering $0.39$ to $0.85$ of the grid, and French, pirate and grumpy critic, the three behaviors with the narrowest regions, hold no expert in any model.

\begin{figure}[!htb]
\centering
\includegraphics[width=\textwidth]{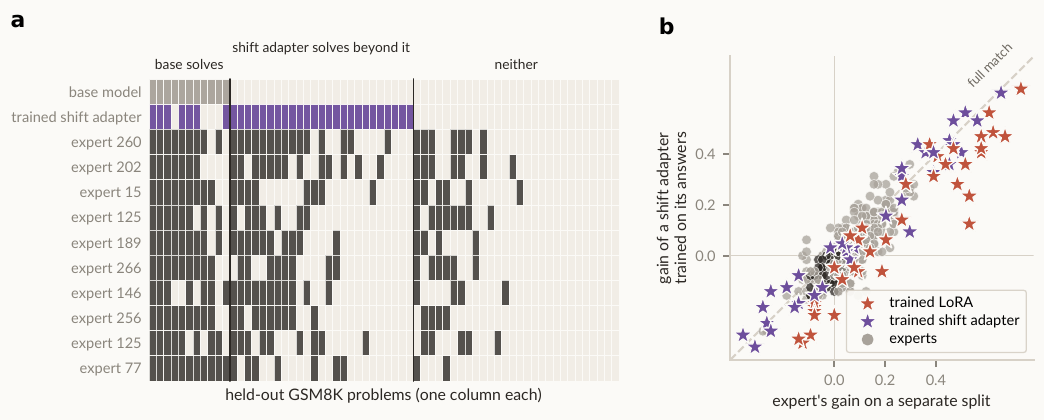}
\caption{\textbf{Experts solve the problems training newly solves, and a shift adapter carries what an expert does.} \textbf{(a)}~Held-out GSM8K problems of Qwen2.5-3B (columns), grouped into those the base model solves, those the trained shift adapter solves beyond it, and neither, against the $10$ experts of highest gain (rows). \textbf{(b)}~The gain of a shift adapter trained on each expert's answers against the expert's own gain, both on a separate split of GSM8K that played no part in selecting the experts, on every model with an expert. The dashed diagonal is a full match. The stars are the trained shift adapter (purple) and LoRA (red) references, whose own answers a shift adapter recovers.}
\label{fig:thicketsid}
\end{figure}
\textbf{Relation to trained states.} We compare the experts on GSM8K with a shift adapter trained on GSM8K, and the two arrive at the same state. Their fields point the same way: over the $328$ experts, the median cosine between an expert's field and the trained adapter's field is $0.23$. They solve the same problems: of the held-out problems an expert solves beyond the base model, the trained adapter solves $1.36$ times as many as chance would give, and more than chance for $91\%$ of the experts (\cref{fig:thicketsid}a). A shift adapter trained on an expert's answers keeps $48\%$ of the experts' improvement, measured on a separate split of GSM8K that played no part in selecting the experts (a ratio of sums over the $328$ experts distilled), where the same training on the trained adapter's own answers keeps $92\%$ (\cref{fig:thicketsid}b).

\end{document}